%% file: main.tex
\documentclass[10pt,journal,compsoc]{IEEEtran}
\ifCLASSOPTIONcompsoc
\usepackage[nocompress]{cite}
\else
\usepackage{cite}
\fi

\usepackage{amsmath,amsfonts}
\usepackage{algorithmicx}
\usepackage{algcompatible}
\usepackage{algorithm}
\usepackage[noend]{algpseudocode}
\usepackage{array}
\usepackage[caption=false,font=normalsize,labelfont=sf,textfont=sf]{subfig}
\usepackage{textcomp}
\usepackage{stfloats}
\usepackage{url}
\usepackage{verbatim}
\usepackage{graphicx}
\usepackage[table]{xcolor}

\usepackage{cite}
\usepackage{hyperref}
\usepackage{xcolor}
\usepackage{hyperref}
\usepackage{booktabs}

\input{notation}

\begin{document}

\definecolor{cvprblue}{rgb}{0.21,0.49,0.74}

\definecolor{softyellow}{RGB}{255, 223, 122}  
\definecolor{softorange}{RGB}{255, 179, 71}   
\definecolor{softred}{RGB}{255, 105, 97}      

\definecolor{softgreen}{RGB}{102, 205, 170} 
\definecolor{softblue}{RGB}{100, 149, 237}  

\title{\ModelName: Gaussian Neural Fields for Multidimensional Signal Representation and Reconstruction}

\author{Abdelaziz Bouzidi, Hamid Laga, Hazem Wannous, Ferdous Sohel
	
    \IEEEcompsocitemizethanks{
        \IEEEcompsocthanksitem Abdelaziz Bouzidi is with the School of Information Technology, Murdoch University (Australia) and with IMT Nord Europe (France). Email: abdelaziz.bouzidi@murdoch.edu.au.
	
        \IEEEcompsocthanksitem  Hamid Laga is with the School of Information Technology, Murdoch University (Australia). Email: H.Laga@murdoch.edu.au
	
        \IEEEcompsocthanksitem Hazem Wannous  is with IMT Nord Europe (France). Email: hazem.wannous@imt-nord-europe.fr.

        \IEEEcompsocthanksitem  Ferdous Sohel is with the School of Information Technology, Murdoch University (Australia). Email: F.Sohel@murdoch.edu.au
    }
    \thanks{Manuscript received August. 8, 2025.}
}

\markboth{\ModelName: Gaussian Neural Fields for Multidimensional Signal Representation and Reconstruction}%
{Bouzidi \MakeLowercase{\etalnospace}: \ModelName}

\IEEEtitleabstractindextext{
\begin{abstract}

Neural fields have emerged as a powerful framework for representing continuous multidimensional signals such as images and videos, 3D and 4D objects and scenes, and radiance fields. While efficient, achieving high-quality representation requires the use of wide and deep neural networks. These, however, are slow to train and evaluate. Although several acceleration techniques have been proposed, they either trade memory for faster training and/or inference, rely on thousands of fitted primitives with considerable optimization time, or compromise the smooth, continuous nature of neural fields. In this paper, we introduce Gaussian Neural Fields (\ModelName), a novel compact neural decoder that maps learned feature grids into continuous non-linear signals, such as RGB images, Signed Distance Functions (SDFs), and radiance fields, using a single compact layer of Gaussian kernels defined in a high-dimensional feature space. Our key observation is that neurons in traditional MLPs perform simple computations, usually a dot product followed by an activation function, necessitating wide and deep MLPs or high-resolution feature grids to model complex functions. In this paper, we show that replacing MLP-based decoders with Gaussian kernels whose centers are learned features yields highly accurate representations of 2D (RGB), 3D (geometry), and 5D (radiance fields) signals with just a single layer of such kernels. This representation is highly parallelizable, operates on low-resolution grids, and trains in under $15$ seconds for 3D geometry and under $11$ minutes for view synthesis. \ModelName {} matches the accuracy of deep MLP-based decoders with far fewer parameters and significantly higher inference throughput. The source code is publicly available at \url{https://grbfnet.github.io/}.

\end{abstract}

\begin{IEEEkeywords}
Radial Basis Functions, Neural Radiance Fields, Deep SDF, 3D reconstruction, novel view synthesis, neural rendering.
\end{IEEEkeywords}
}

\maketitle
\IEEEdisplaynotcompsoctitleabstractindextext
\IEEEpeerreviewmaketitle

\section{Introduction}
\IEEEPARstart{I}{n} recent years, neural fields have gained popularity for their ability to accurately represent, in a resolution-agnostic manner, multidimensional signals such as RGB images and videos~\cite{strumpler2022implicit, chen2021nerv}, 3D geometry~\cite{park2019deepsdf, wang2021neus}, and radiance fields~\cite{mildenhall2020nerf, wang2021neus}. Compared to traditional discrete signal representations, neural fields use Multi-Layer Perceptrons (MLPs) to learn a mapping between a continuous query space, \eg a 2D point $\point \in \rtwo$,  a 3D point $\point \in \rthree$, or a 3D point and a viewing direction $\viewdir \in \stwo$, to its corresponding output value such as an RGB color, a Signed Distance Function (SDF) value, a volume density, or radiance. This flexible formulation has enabled significant breakthroughs in a range of tasks, including 3D and 4D reconstruction, neural rendering, novel view synthesis, and virtual avatar generation.

Despite their effectiveness, early neural fields suffer from substantial training time (\eg vanilla NeRF~\cite{mildenhall2020nerf} requires several hours to train on a single scene) and large memory footprints. To address these issues, recent methods accelerate inference~\cite{sun2022direct, fridovich2022plenoxels, karnewar2022relu} and training~\cite{fridovich2022plenoxels, muller2022instant} by explicitly encoding the signal of interest into features organized into dense grids~\cite{fridovich2022plenoxels,fridovich2023k} or sparse multi-resolution hash grids~\cite{muller2022instant}. These approaches then estimate the signal properties at a given query point either via direct interpolation of the feature grids~\cite{fridovich2022plenoxels} or by using shallow MLPs operating on interpolated features and play the role of feature decoders~\cite{muller2022instant, fridovich2023k}. While effective, these techniques either trade memory for reduced inference time (as the space needed to store the feature grids exceeds that of the network parameters) or introduce computational bottlenecks in the decoding stage. Alternatively, primitive-based methods such as Gaussian Splatting (3DGS)~\cite{kerbl20233d} and splines~\cite{williams2021neural} achieve real-time inference but remain slow to train and struggle to capture the global structure of scene geometry.

In this paper, we introduce Gaussian Neural Fields (\ModelName), a novel lightweight decoder based on Radial Basis Functions (RBFs). Instead of using traditional MLPs, which approximate nonlinear functions via piecewise linear activations such as ReLU, our method uses Gaussian kernels defined directly in the learned feature space. Each kernel is activated by the Euclidean distance between the input feature vector and a learned RBF center, with per-dimension scaling enabling anisotropic and independent modeling of the learned feature space. By aligning the dimensionality of the kernels with the learned input features, \ModelName{} effectively represents highly nonlinear signals with a single decoding layer composed of less than $\nrbf = 64$ units. This results in a representation that is significantly more compact than shallow three-layer MLPs used in state-of-the-art methods~\cite{muller2022instant}. This novel neural architecture with minimal depth shortens gradient paths, thus accelerating training and inference. Additionally, our fully vectorized implementation reduces computational complexity from $\bigO{\batchsize\nrbf^2}$ to $\bigO{\batchsize\nrbf}$, where $\batchsize$ is the batch size and $\nrbf$ is the number of decoding units. As a result, \ModelName{} achieves comparable reconstruction quality using only about half the floating-point operations (\FLOPs) required by a three-layer MLP. We demonstrate the effectiveness of this novel representation in encoding Signed Distance Fields, RGB images, and radiance fields.

The remaining parts of the paper are organized as follows: Section~\ref{sec:related-work} reviews the state of the art. Section~\ref{sec:neural_fields} describes in detail the proposed framework. Section~\ref{sec:results} presents the experimental results and analyzes the performance of the proposed representation. Section~\ref{sec:conclusion} concludes the paper with a summary of the main findings and a discussion on the potential future research directions.

\section{Related Work}
\label{sec:related-work}
We refer the reader to recent surveys on neural fields~\cite{yunus2024recent} and 3D Gaussian Splatting~\cite{chen2024survey}. Here, we focus on the papers that are closely related to our work.

\noi \textbf{Neural fields} use MLPs to represent nD signals, such as 2D images~\cite{muller2022instant,chen2023neurbf,chen2023factor}, SDF values~\cite{piperakis2001affine,piperakis20013d,park2019deepsdf,wang2021neus}, and radiance fields~\cite{mildenhall2020nerf}, as continuous functions. While powerful, these methods require querying deep and wide MLPs, often composed of dozens of layers and thousands of neurons,  millions of times, both during training and at runtime. As such, training scene-specific neural fields can take hours or days per scene. Explicit feature grid-based neural fields~\cite{muller2022instant} significantly improve training and inference times by shifting most of the computation burden to the features and then using shallow MLPs to decode them into geometry and appearance. These methods are expensive in terms of memory requirements, as the space required to store the features exceeds that of the network parameters. Some methods improve the computation time by reducing the number of forward passes through early ray termination~\cite{liu2020neural, reiser2021kilonerf} or efficient surface localization~\cite{neff2021donerf, barron2022mip, piala2021terminerf, kurz2022adanerf,reiser2021kilonerf,liu2020neural, sun2022direct, yu2021plenoctrees, fridovich2022plenoxels}. Others employ baking strategies by precomputing and storing some content in sparse 3D data structures~\cite{hedman2021baking,reiser2023merf,yariv2023bakedsdf,garbin2021fastnerf}.  PlenOctrees~\cite{yu2021plenoctrees}, on the other hand,  represents view-dependent effects as a weighted sum of Spherical Harmonic (SH) basis, parameterized by the viewing direction. The view-dependent color can then be computed in a single forward pass.  Plenoxels~\cite{yu2021plenoctrees} follows the same idea but uses interpolation to decode the view-dependent color, dropping entirely the MLPs. While these techniques can reduce training time to less than  $15$ mins~\cite{muller2022instant,fridovich2022plenoxels}, their memory requirements remain very high.


\noi \textbf{3D Gaussian Splatting}-based methods~\cite{kerbl20233d}, which represent geometry and radiance using a large number of 3D Gaussians, achieve realtime rendering of high-resolution images. They, however,  can be slow at training and expensive in terms of memory requirements since they need to store a large number of Gaussians, usually in the order of millions. Subsequent works reduce the memory requirements, and thus improve scalability,  by pruning insignificant Gaussians~\cite{papantonakis2024reducing,lee2024compact,chen2024hac} or compressing the memory usage of the parameters of the 3D Gaussians~\cite{papantonakis2024reducing,niedermayr2024compressed,lee2024compact}. Although they achieve important reductions in memory post-training, considerable potential remains for reducing memory usage during the training phase.  


\noi \textbf{Radial Basis Functions (RBFs)}   can be interpreted as a neural network that has an input layer and one output neuron. Unlike  MLPs,  a neuron in the input layer performs a slightly more complex transformation of its input by using a kernel function, \eg a Gaussian or a thin-plate spline. To accurately represent complex 3D geometries, Carr \etal~\cite{carr2001reconstruction,ohtake2005multi} use a large number of basis functions.   Kojekine \etal~\cite{kojekine2004surface} optimize the representation by using RBFs with compact support, resulting in a sparse linear system that is fast to solve and efficient in terms of memory storage. 

In the era of deep learning, only a few papers have explored the use of RBF kernels in representing neural fields. For instance,  Neural Splines~\cite{williams2021neural} and NKSR~\cite{huang2023neural} use a large set of spline kernels to represent the signed distance field. Similar to early methods~\cite{carr2001reconstruction,kojekine2004surface,ohtake2005multi}, the RBFs need to be trained on point clouds paired with their corresponding surface normal vectors. They also require a large number (in the order of thousands) of kernels to represent 3D geometry with high accuracy.
Papers such as~\cite{ramasinghe2021learning,ramasinghe2022beyond,chng2022gaussian} proposed a broader class of activation functions, including Gaussians, that enable coordinate MLPs to encode high-frequency signals. They demonstrated that coordinate-based Gaussian MLPs converge better than sinusoid and ReLU-based activations. 

Different from these methods are NeuRBF~\cite{chen2023neurbf} and FactorFields~\cite{chen2023factor,chen2023dictionary}, which use RBFs in the feature encoding stage.  For instance, NeuRBF~\cite{chen2023neurbf} learns the location of the local features, instead of pre-setting them on regular 2D or 3D grids,  and then efficiently interpolates them using RBFs instead of trilinear interpolation. Both NeuRBF and FactorFields use standard MLPs to decode the learned features.  Finally, Jiang \etal~\cite{jiang2024cofie} use a decoding  MLP composed of traditional linear layers followed by one or multiple quadratic layers.  Although these methods have demonstrated a better representation accuracy,  they all require querying, during the decoding stage, deep and wide decoding MLPs.

\begin{figure*}[t]
    \centering

    \begin{tabular}{@{}ccc@{}}
        \includegraphics[trim={3.5cm  2cm 9cm 1cm}, clip,  height=.2\textheight]{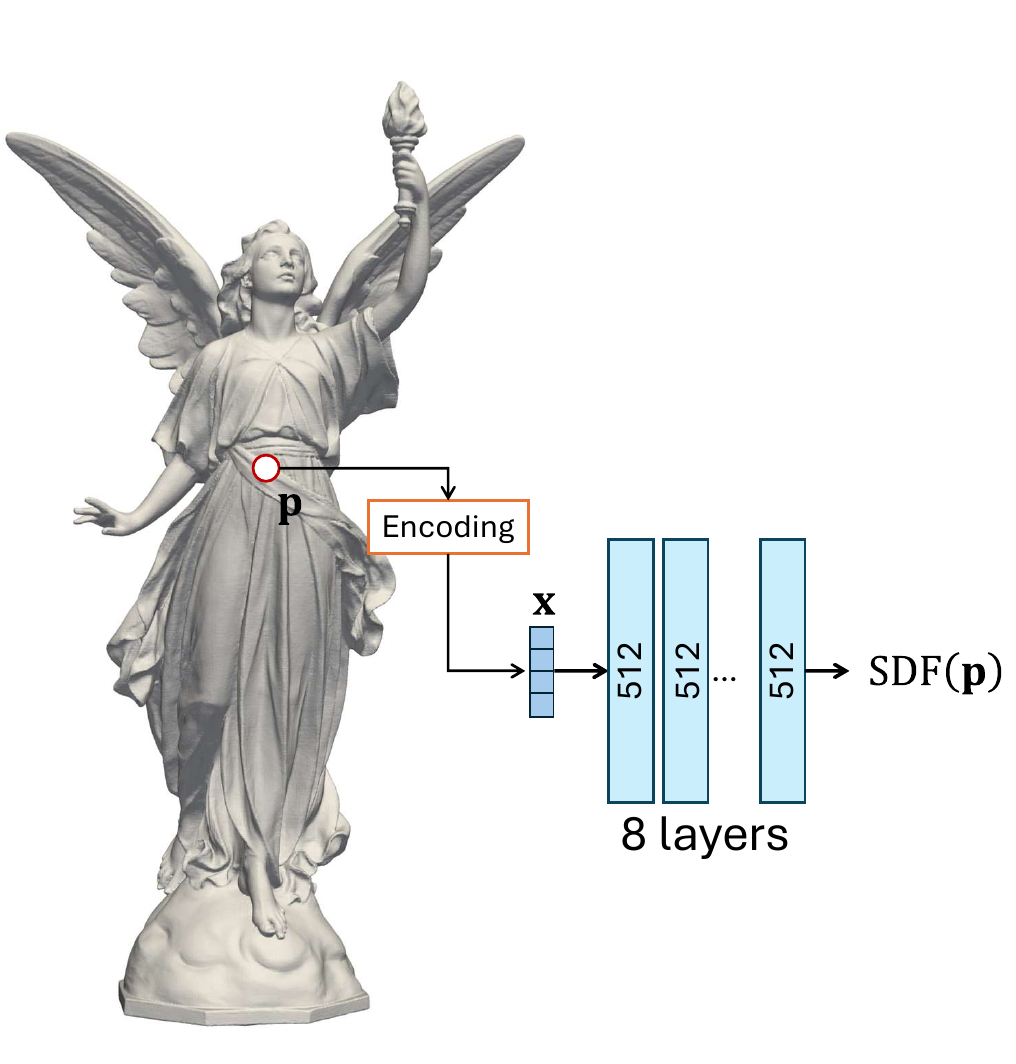} &\includegraphics[trim={3.5cm  2cm 5cm 1cm}, clip,  height=.2\textheight]{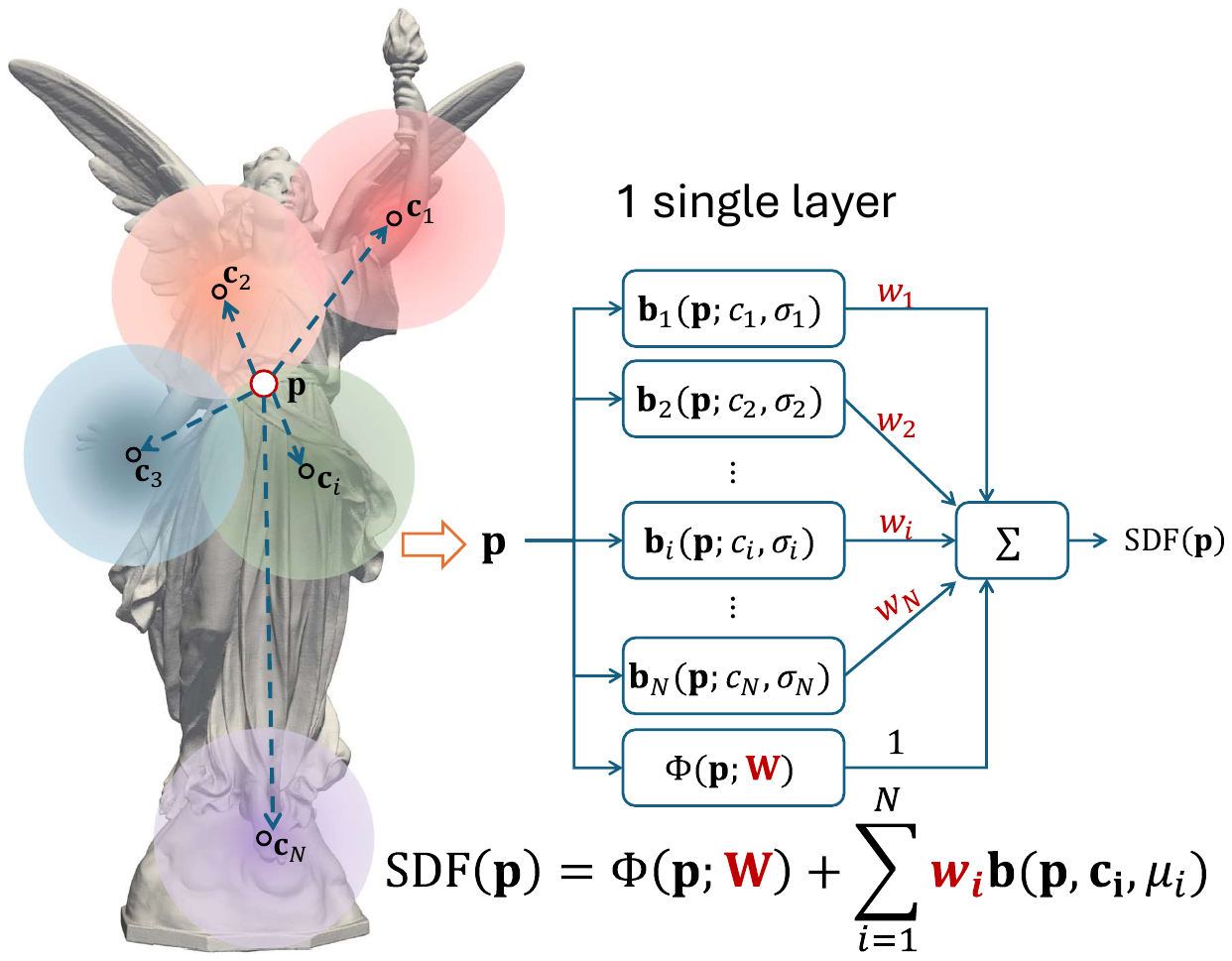} &
        \includegraphics[trim={2.5cm  2cm 1cm 1cm}, clip,  height=.2\textheight]{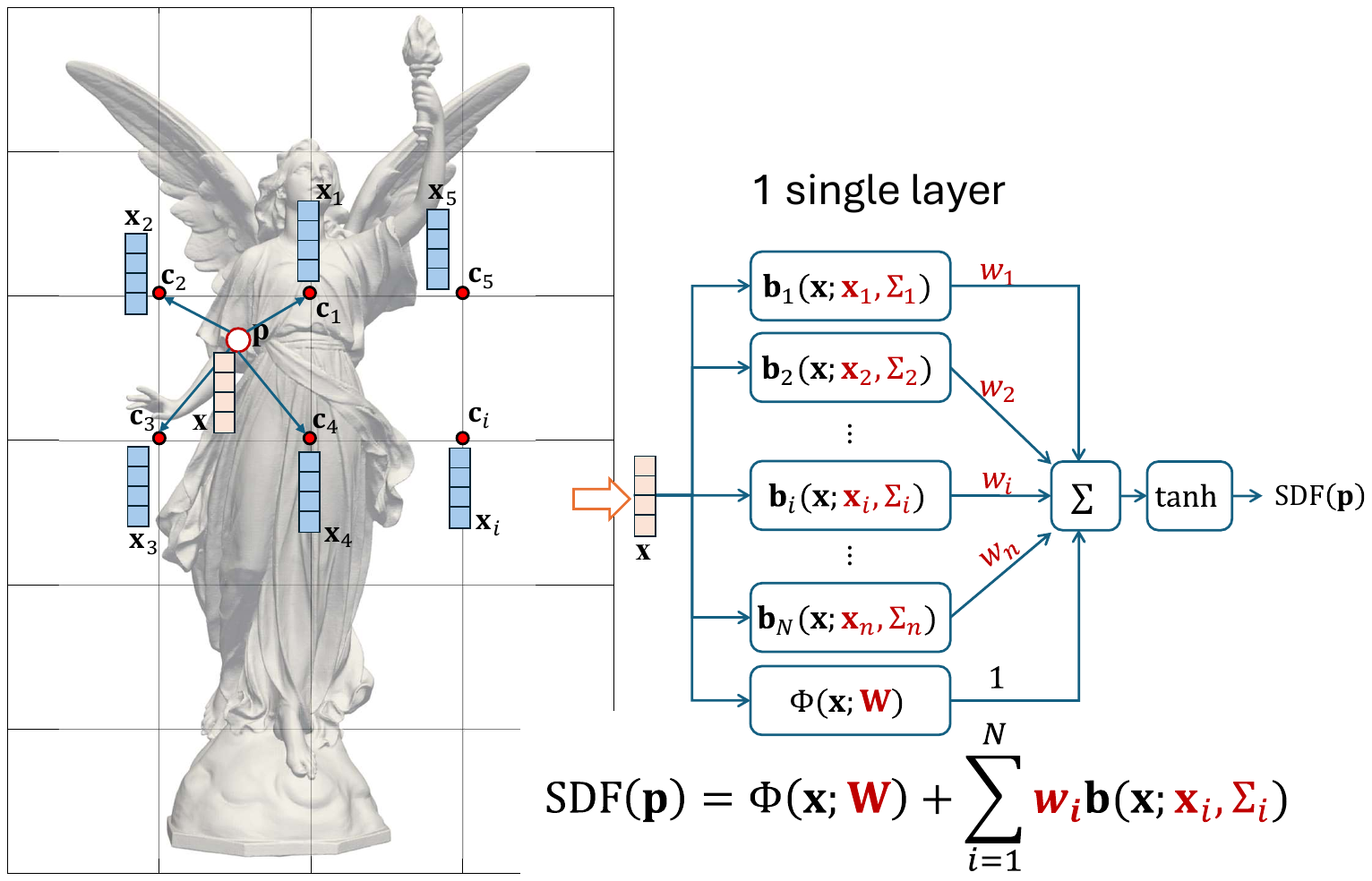}\\
        \textbf{(a)} Traditional Deep SDF. &\textbf{(b)} Traditional RBF networks. & \textbf{(c)} Proposed Gaussian RBF networks.
    \end{tabular}
    \caption{Illustration of the proposed Gaussian RBFNet in comparison to the traditional neural fields such as Deep SDF and the traditional RBF networks for 3D geometry representation.}
    \label{fig:model_architecture}
    
\end{figure*}

\subsubsection*{Contributions}
Compared to the state-of-the-art, our contribution is two-fold; \textbf{First}, we introduce a novel family of neural fields that use RBF kernels, instead of MLPs,  to decode low-resolution feature grids into high-quality nD signals. \textbf{Second}, we demonstrate that a single-layer RBF network composed of a small number of learnable Gaussian kernels that operate in the feature space is sufficient to efficiently represent RGB images, 3D geometry, and radiance fields. Extensive experiments demonstrate that our formulation has several benefits compared to the state-of-the-art:  it is fast to train (less than $15$ secs for SDF representation and around $10$ mins for radiance fields representation) without sacrificing quality. It is very compact and can render images at more than $60$ fps.

\section{Gaussian Neural Fields}
\label{sec:neural_fields}

\subsection{General formulation}
\label{sec:formulation}
Neural fields (Figure~\ref{fig:model_architecture}-(a)) represent a signal $\signal: \real^\inputdim \to \real^\outputdim$, which maps a point $\point\in \real^\inputdim$ to a $\outputdim$-dimension output, with $\inputdim \ge 1, \outputdim\ge 1$, as a continuous function of the form 
$\signal = \dec \circ \enc$. Here, 
    \begin{itemize}
        \item  $\enc: \real^\inputdim \to \real^\featuredim$ is an encoding function that maps a point $\point \in \real^\inputdim$ to a neural feature $\featurevector \in \real^\featuredim$, with $\featuredim\ge 1$, and 
        
        \item $\dec: \real^\featuredim \to  \real^\outputdim$ is a decoding function that extracts the desired output from the neural feature $\featurevector = \enc(\point) \in \real^\featuredim$. 
    \end{itemize}
        
\noi  Point encoding-based neural fields define $\enc$ as a multi-frequency sinusoid function.  Explicit neural fields use learned features organized into grids. The encoding $\enc(\point)$ of a point $\point \in \real^\inputdim$ is then determined by trilinearly interpolating the features around $\point$. 
Different from these methods is NeuRBF~\cite{chen2023neurbf}, which, instead of using trilinear interpolation, it uses Radial Basis Functions (RBFs) as an interpolant to compute the feature at $\point$. Furthermore, instead of 
predefining the location of the features (as the corners of a regular grid in $\rthree$),  NeuRBF~\cite{chen2023neurbf} defines these locations as the centers of the Radial Basis Functions. Both the centers, the shape, and the blending weights are learnable parameters that are optimized during training.

To parameterize the function  $\dec$, which decodes the learned features into the target output, state-of-the-art methods~\cite{chen2023neurbf} use deep MLPs 
composed of multiple layers where each layer is followed by an activation function (usually ReLU). In this paper, we revisit this decoding scheme and demonstrate that a single layer composed of a few RBF kernels can achieve state-of-the-art performance. Thus, our method is orthogonal to existing methods as it focuses on the decoding stage, while NeuRBF~\cite{chen2023neurbf} focuses on the input encoding.

Mathematically, a 1D signal $\signal: \domain \subseteq \rthree \to \real$ can be approximated using a weighted sum of basis functions, \ie:
\begin{equation}
    \forall \point \in \domain, \signal(\point) = \polynomial(\point) + \sum_{i=1}^{\nbases} \weight_i \radialbasis_i\left( \point; \params_i \right).
    \label{eq:rbf1D}
\end{equation}

\noi Here, $\{\radialbasis_i\}_{i=1}^\nbases$ are  real-valued functions defined on $\point\in\domain$ and parameterized by  $\params_i$, $\weight_i \in \real$ are blending weights, and $\polynomial$ is a polynomial of low degree. Commonly used basis functions include Fourier bases, Wavelets, and Radial Basis Functions (RBFs). We are interested in RBFs of the form:
\begin{equation}
    \signal(\point) = \polynomial(\point) + \sum_{i=1}^{\nbases} \weight_i \radialbasis_i\left( \left\| \point - \thecenter_i\right\|; \params_i \right),
    \label{eq:general_rbf1D}
\end{equation}

\noi which have been used to represent Signed Distance Functions~\cite{carr2001reconstruction}.  Here, $\radialbasis_i: \rnonneg \to \real$ is a real-valued basis function on $\rnonneg$, centered at a point $\thecenter_i\in \rthree$ and has $\params_i$ as additional parameters. Popular choices for  $\radialbasis$ include the thin-plate spline, used to represent SDFs~\cite{carr2001reconstruction},  and the Gaussian function of the form $\radialbasis(r) = \exp(-a r^2)$  where   $a$ is the inverse of the variance and controls the shape of the Gaussian.  As shown in Figure~\ref{fig:model_architecture}-(b), an RBF function can be implemented as a feedforward  network composed of:
\begin{itemize}
    \item  an input layer with $\nbases+1$ neurons. The  $i-$th neuron applies the $i-$th basis function $\radialbasis_i$ to its input and the $(\nbases+1)-$th neuron implements the polynomial $\polynomial$, and  
    
    \item an output layer composed of a single neuron that computes the weighted sum of the outputs of the neurons in the previous layer. 
\end{itemize}

\noi It can also be considered as a neural field where the encoding function $\enc$ is set to identity and thus $\signal = \dec$. During training, the centers $\thecenter_i \in \rthree$ of the basis functions are preset in advance, usually selected to be on and around the signal of interest~\cite{carr2001reconstruction,williams2021neural}, while the blending weights $\weight_i$ are learnable. 

While powerful, representing complex signals requires a large number of basis functions. For example, Carr \etal~\cite{carr2001reconstruction} and later Williams \etal~\cite{williams2021neural} and Huang \etal~\cite{huang2023neural}
use  $\nbases \ge 75$K to accurately represent SDFs. Also, Equations~\eqref{eq:rbf1D} and~\eqref{eq:general_rbf1D} are limited to 1D signals with $\inputdim = 3$ and output dimension $\outputdim = 1$. However, neural fields  involve multidimensional signals with $\inputdim \ge 1$ and $\outputdim \ge 1$, \eg $(\inputdim=3, \outputdim=1)$ for SDFs, $(\inputdim=2, \outputdim=3)$ for RGB images, and $(\inputdim=5, \outputdim=4)$ for radiance fields. Additionally, the input is generally encoded using a function $\enc$, yielding a high-dimensional feature vector, which is then decoded using MLPs.  In this paper, we explore the efficiency of RBF kernels as feature decoders. Thus,  we first propose to generalize  RBF networks, hereinafter referred to as RBF-based decoders,  to signals of arbitrary input/output dimensionality. Let $\featurevector = \enc(\point) \in \real^{\featuredim}$,  an RBF-based decoder can be defined as 
\begin{equation}
    \dec(\featurevector) = \polynomial(\featurevector) + \weights^\top \bases(\featurevector; \params_i).    \label{eq:general_multidim_rbf}
\end{equation}

\noi Here, $\bases(\featurevector; \params) = \left(\radialbasis_1(\featurevector; \params_1), \dots, \radialbasis_N(\featurevector; \params_N) \right)^\top$ denotes the output of the $N$ basis functions, each parameterized by $\params_i$. The matrix $\weights$ is defined as $\weights = \left(\Weight^1 | \dots | \Weight^\outputdim\right)$ where each column $\Weight^k = (\weight^k_1, \dots, \weight^k_N)^\top$ defines the weights used for blending the RBF responses to produce the $k$-th output channel.  In our implementation, we omit the polynomial term and use Gaussian kernels  as radial basis, \ie
\begin{equation}
\radialbasis_i(\featurevector) = \exp\left( -(\featurevector - \boldsymbol{\mu}_i)^\top \boldsymbol{\beta}_i (\featurevector - \boldsymbol{\mu}_i) \right).
\end{equation}

\begin{figure*}[!t]
    \centering
    \resizebox{\textwidth}{!}{
    \begin{tabular}{@{}c@{}}
        \includegraphics[width=1\textwidth]{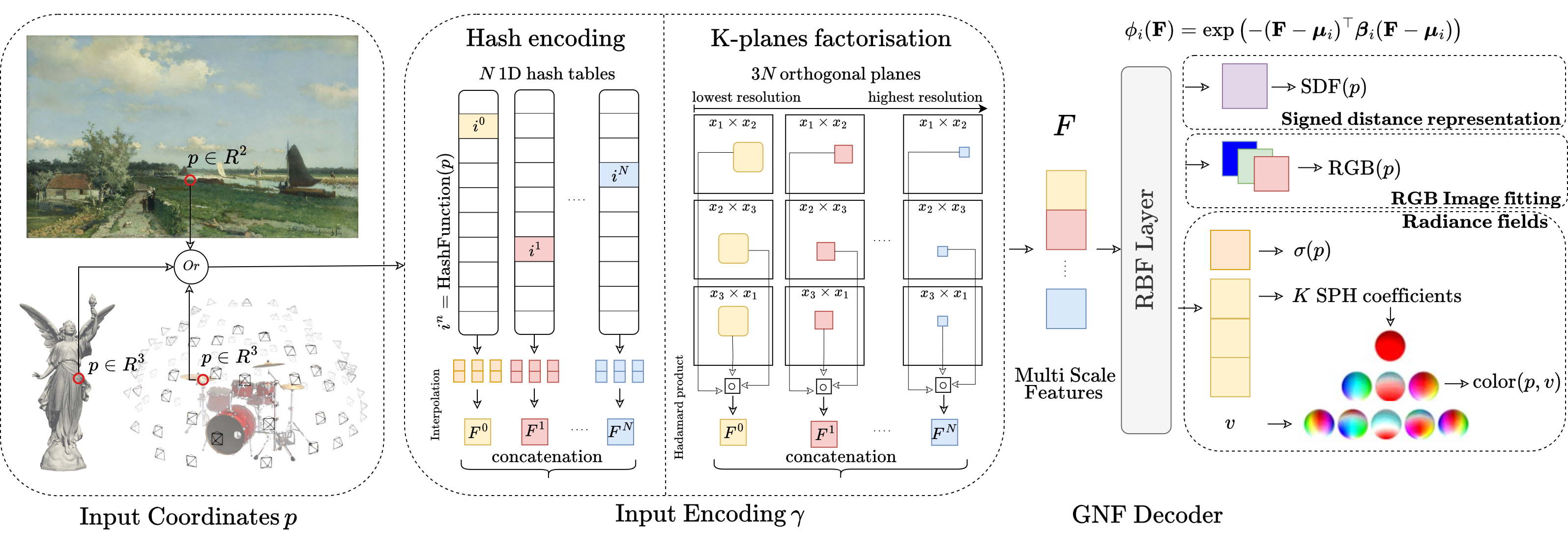} \\
    \end{tabular}
    }
    \caption{Architecture of the proposed \ModelName\; network for representing 3D shapes, 2D images and radiance fields.}
    \label{fig:nerf_architecture}
\end{figure*}

\noi  Here,  \( \featurevector = \enc(\point) \in \real^{\featuredim} \) is the encoded input, \( \boldsymbol{\mu}_i \in \real^{\featuredim} \) is the center of the $i$-th basis function, and \( \boldsymbol{\beta}_i = \mathrm{diag}(\beta_{i_1}, \dots, \beta_{i_\featuredim}) \) is a diagonal covariance matrix. While a full covariance matrix could be used, it requires learning \( \frac{1}{2}\featuredim(\featuredim - 1) + \featuredim \) parameters per basis, which is impractical when using high-dimensional feature encoding. Diagonal covariance matrices provide a scalable and stable parameterization with per-dimension adaptability. We refer to this representation as Gaussian Neural Fields (\ModelName). Figure~\ref{fig:nerf_architecture} summarizes the framework.

\subsection{Complexity Analysis}
\label{sec:complexity}

We analyze the computational complexity of our proposed \ModelName{} decoder and compare it against a standard three-layer MLP. Let $\featdim$ denote the input feature dimension, $\width$ the width of each hidden layer in the MLP, $\nrbf$ the number of radial units in our decoder, analogous to the width $\width$ in a standard three-layer MLP, and $\batchsize$ the batch size.

\vspace{3pt}
\noi\textbf{MLP Decoder.}  An MLP with two hidden layers of width $\width$ requires three matrix-vector multiplications in the forward pass, leading to the following FLOPs:
\begin{equation}
    \FLOPs_\text{MLP} =  2(\featdim\width + \width^2 + \width)\batchsize.
\end{equation}

\noi The backward pass includes computing gradients with respect to the weights, activations, and inputs. This requires:
\begin{equation}
    \text{Total}_\text{MLP} \approx 6(\featdim\width + \width^2 + \width)\batchsize.
\end{equation}

\noi Thus, the total complexity  is $\bigO{\batchsize(\featdim\width + \width^2)}$. This shows that the complexity is quadratic with respect to the width $\width$ of the MLP.

\vspace{3pt}
\noi\textbf{\ModelName{} Decoder.}  
Each radial unit computes a squared distance in the feature space, followed by a Gaussian activation and a weighted sum. Thus, the  cost of a forward pass is:
\begin{equation}
    \FLOPs_\text{RBF} =   (3\featdim + 3) \batchsize \nrbf.
\end{equation}

\noi The backward pass computes gradients with respect to the input, centers, bandwidths, and weights. Thus, the total FLOPs is:
\begin{equation}
\text{Total}_\text{RBF} \approx (7\featdim + 6)\batchsize \nrbf.
\end{equation}

\noi This yields a total complexity of $\bigO{\batchsize\nrbf\featdim}$, which grows linearly with respect to the feature dimension and the number of RBF units. Unlike the MLP, our decoder avoids costly matrix multiplications and eliminates the quadratic dependency on $\width$, offering better efficiency when dealing with high-resolution signals and in resource-constrained scenarios.

\begin{table}[t]
\centering
\begin{tabular}{@{}lc@{ }c@{ }c@{}}
\toprule
\textbf{Model} & \textbf{Forward} & \textbf{Total FLOPs} & \textbf{Complexity} \\
 & \textbf{FLOPs} & \textbf{(incl. backward)} &  \\
\midrule
\textbf{MLP (3 layers)}    & $126.7 \times 10^6$ & $380.2 \times 10^6$ & $\bigO{\batchsize(\featdim\width + \width^2)}$ \\
\textbf{\ModelName{} Decoder}       & $69.1 \times 10^6$  & $192.0 \times 10^6$ & $\bigO{\batchsize\nrbf\featdim}$ \\
\bottomrule
\end{tabular}
\caption{Complexity analysis for $\featdim= 32$, $\width = \nrbf = 64$, and $B = 10$K.}
\label{tab:flops}
\end{table}

\vspace{3pt}
\noi\textbf{Numerical Example.}  
We consider a practical configuration with an input dimension of $\featdim = 32$, decoder width $\width = \nrbf = 64$, and batch size $\batchsize = 10,\!000$. As shown by the total FLOPs (forward + backward) for both models reported in Table~\ref{tab:flops}, our proposed \ModelName{} decoder requires approximately half the FLOPs of a standard 3-layer MLP, highlighting its computational efficiency compared to existing neural field decoders.

\subsection{GNF for Signed Distance Field representation} 
\label{sec:SDF_RBF}

We first show how the proposed \ModelName{} with feature grid-based input encoding can accurately learn and represent, in a compact manner,  the SDF of complex 3D objects.   We focus on bounded scenes, which, when normalized for translation and scale, fit within the volume $\domain = [0, 1]^3$, \ie $\inputdim = 3$ and $\outputdim=1$. 
Algorithm~\ref{alg:rbfdecoder} summarizes the SDF reconstruction process. 

\begin{algorithm}
    \caption{\label{alg:rbfdecoder}\ModelName\, for SDF Estimation}
    \begin{algorithmic}[1]
    \Require Query point $\point \in \domain$
    \Require Multiresolution feature grid $\mathcal{G}$
    \Require Learnable centers $\thecenter_i \in \real^d$, weights $\weight_i \in \real$, and bandwidths $\bandwidth_i \in \mathbb{R}^d$ for $i=1,\dots,n$
    \Ensure Estimated Signed Distance Function (SDF) value for $\point$
    
    \State \textbf{Feature Encoding:}
    \State $\enc(\point) \gets \text{InterpolateGrid}(\point, \mathcal{G})$ \Comment{Retrieve multiscale feature}
    
    \State \textbf{Radial Basis Aggregation:}
    \State $B \gets 0$
    \For{$i = 1$ to $n$}
    \State $d_i^2 \gets \sum_j \bandwidth_{i,j} \cdot \left(\enc_j(\point) - \thecenter_{i,j}\right)^2$ \Comment{Anisotropic squared distance}
    \State $\phi_i \gets \exp(-d_i^2)$ \Comment{Radial activation}
    \State $B \gets B + \weight_i \cdot \phi_i$
    \EndFor
    
    \State \Return $\text{SDF}(\point) \gets B$
    \end{algorithmic}
\end{algorithm}

\noi\textbf{(1) Input encoding $\enc$.} We consider feature grid-based encoding where trainable codes are arranged in a regular 3D grid and optimized, in an auto-decoding fashion, along with the network parameters. For SDF representation, we adopt InstantNGP's  multi-level hash grid. We use $16$ levels with grid resolutions ranging from $4$ to $512$. Each grid cell stores a 1D feature. The feature at any given point $\point\in \domain$ is obtained by trilinearly interpolating the features at its surrounding grid corners,  for each of the $16$ levels of the grid. Thus, the final feature vector is of dimension $\featuredim = 16$. This multi-resolution encoding allows us to capture 3D geometry at different levels of detail. Our formulation, however, is general and supports any grid feature-based encoding of the input, \eg~\cite{chan2022efficient,wang2023pet}.

\begin{figure}[!t]
    \centering
    \resizebox{0.49\textwidth}{!}
    {
    \begin{tabular}{@{}c@{}c@{}c@{}c@{}c@{}}
        \centering
        \includegraphics[trim= {8cm 0cm 8cm 5cm}, clip, height=0.20\textheight]{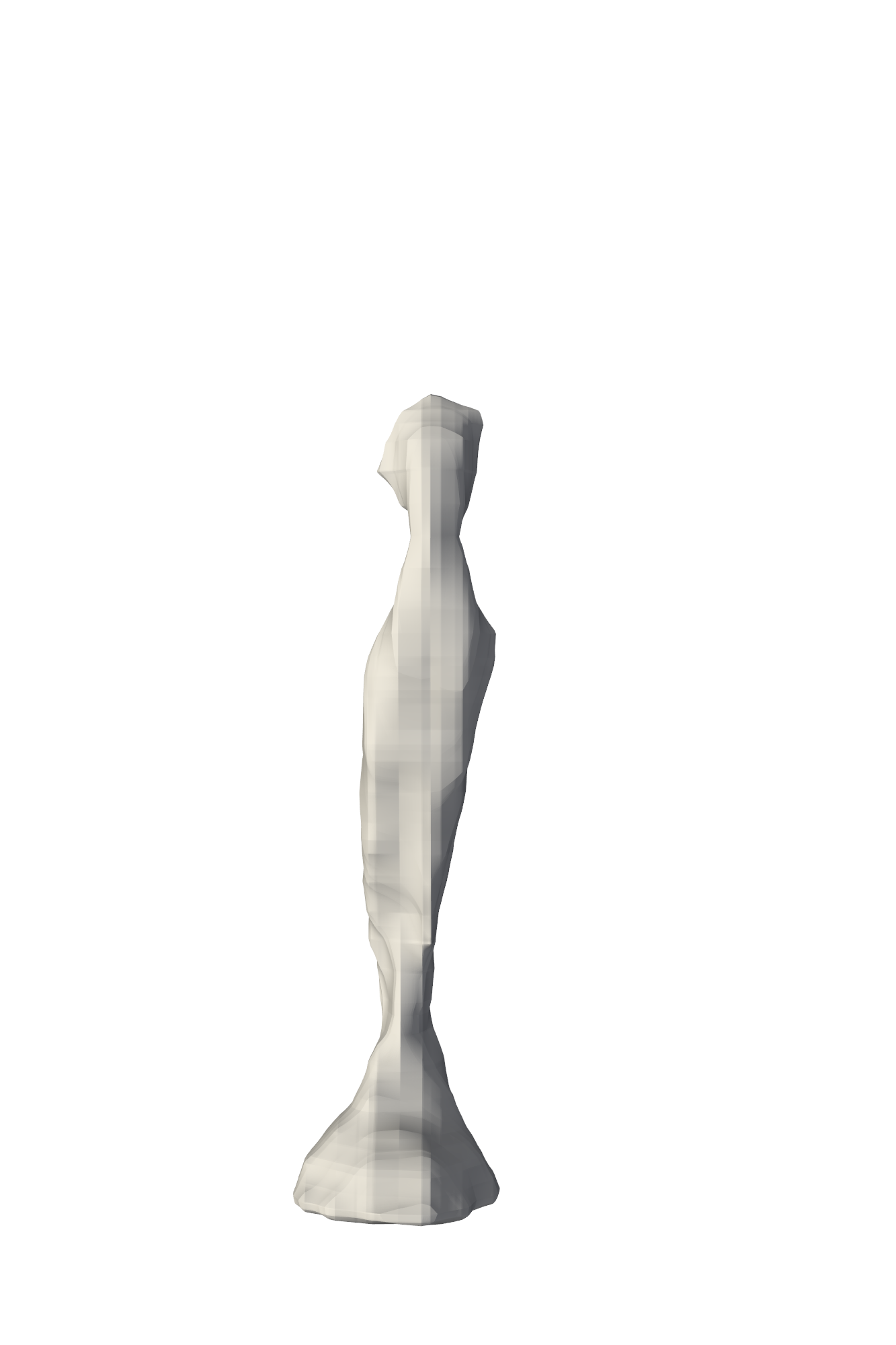} 
        &
        \includegraphics[trim= {7cm 0cm 7cm 5cm}, clip, height=0.20\textheight]{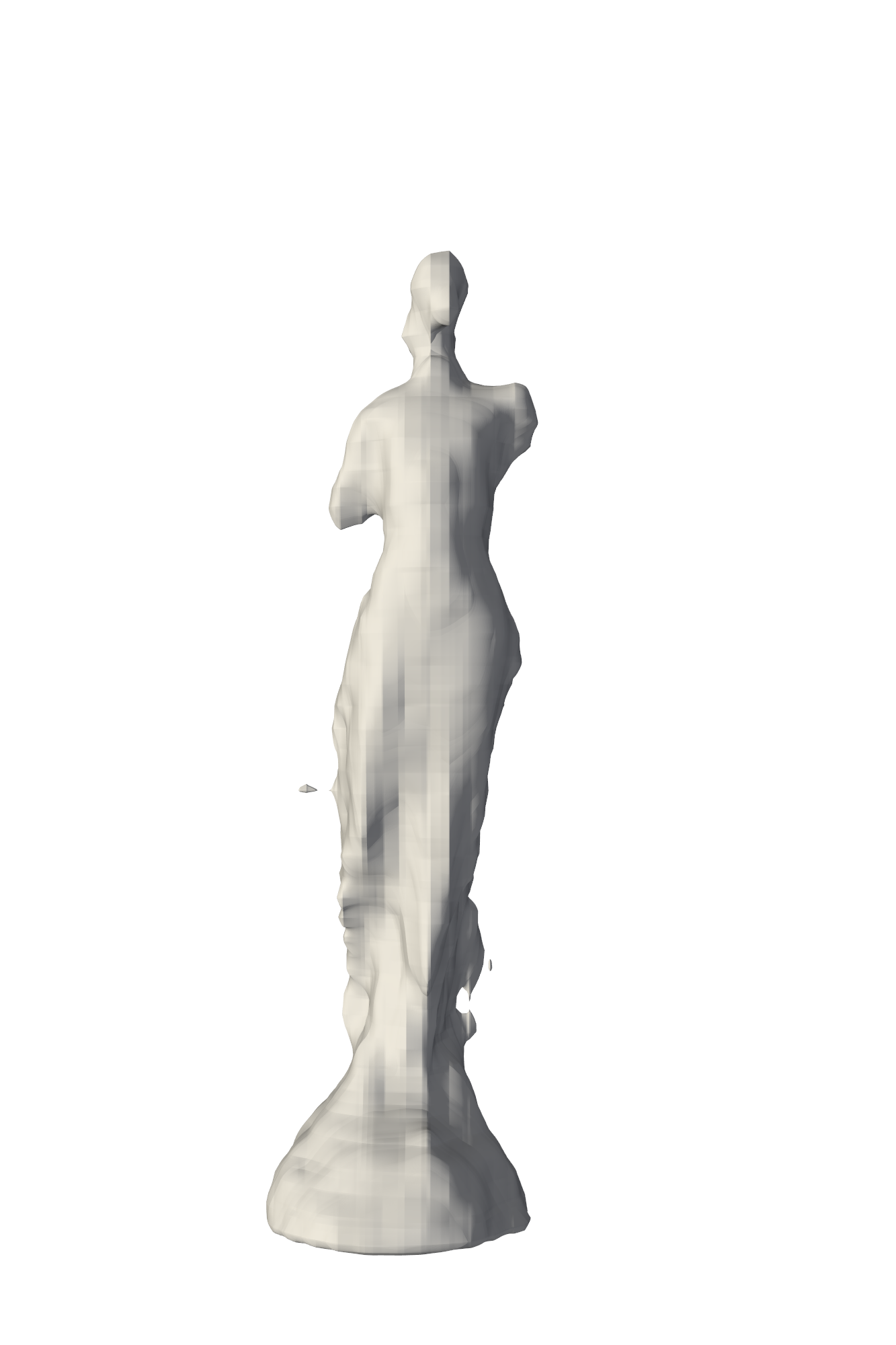} 
        &
        \includegraphics[trim= {5cm 0cm 5cm 5cm}, clip, height=0.20\textheight]{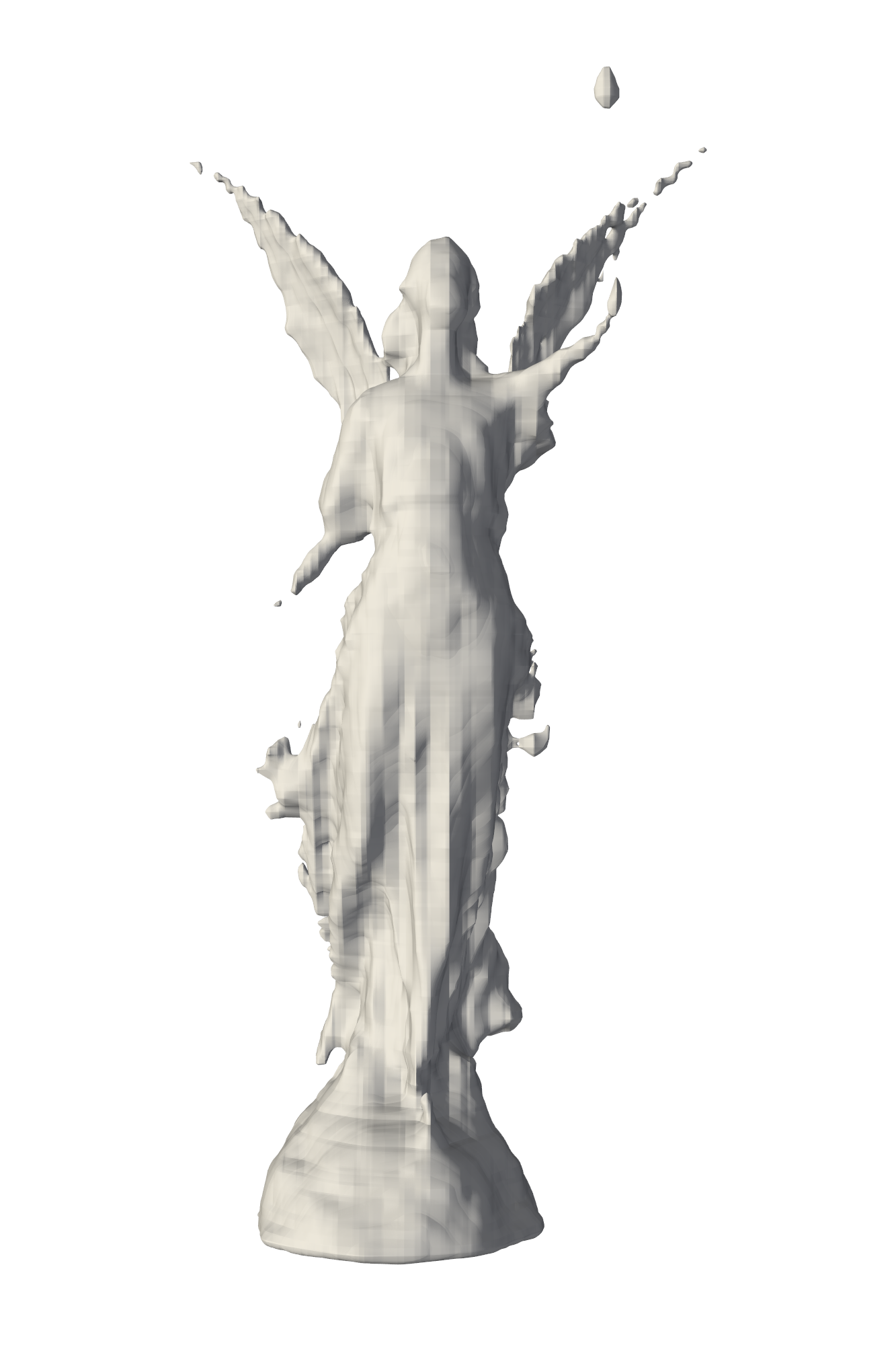} %
         &
        \includegraphics[trim= {2cm 0cm 2cm 2cm},  clip, height=0.20\textheight]{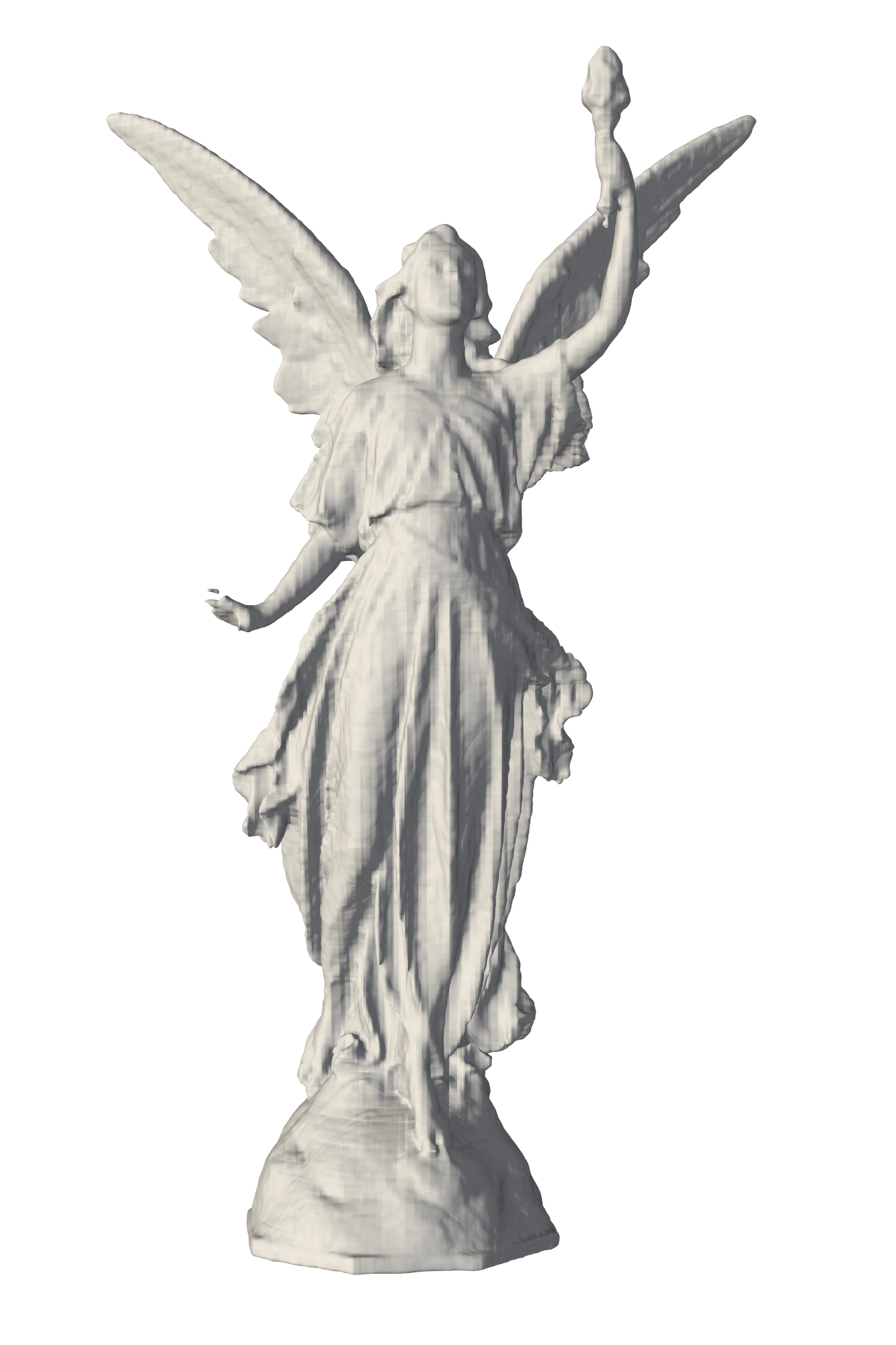} %
         &
        \includegraphics[trim= {1cm 0cm 1cm 2cm},  clip, height=0.20\textheight]{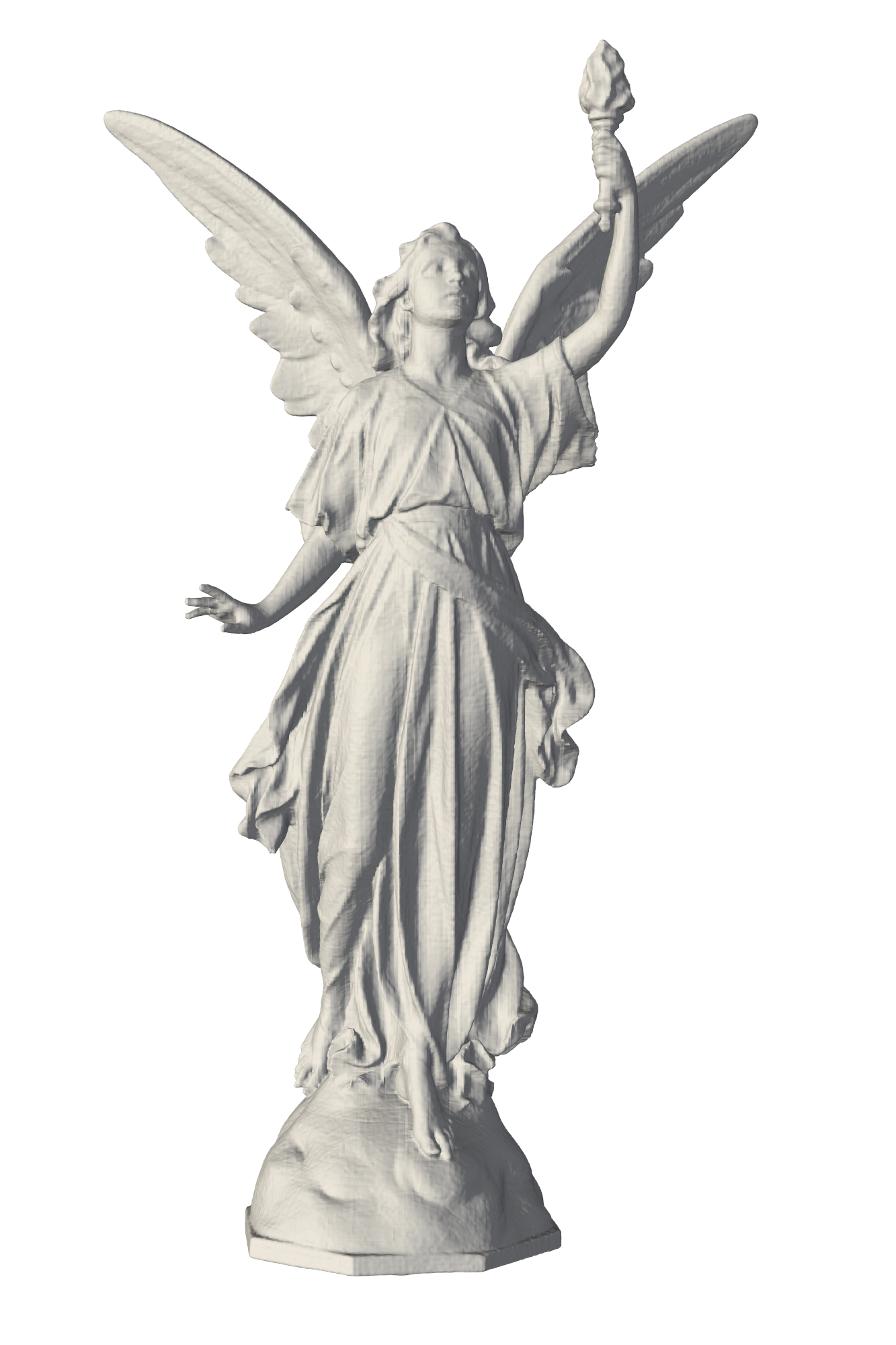} \\
        8/16 levels& 9/16 levels & 10/16 levels & 12/16 levels & Full model\\
     \end{tabular}   
    }
    \caption{Illustration of the multi-scale nature of the proposed  \ModelName {} representation of 3D geometry.}
    \label{fig:multiscale}
\end{figure}

\noi\textbf{(2) Feature decoding using Gaussian Neural Fields.} Figure~\ref{fig:model_architecture}-(c) summarizes the model architecture.  We use Gaussian RBFs defined on the high-dimensional feature space ($16$D for InstantNGP encoding). This enables a localized decoding, making it possible to capture fine geometric details with a significantly smaller number of Gaussians (between $26$ and $128$)  than traditional RBFs~\cite{carr2001reconstruction}, which require more than $75$K Gaussian kernels. 
In our implementation, each Gaussian kernel $\radialbasis_i$ has a spherical shape and thus, it is fully defined by its center $\meanfeaturevector_i \in \real^\featuredim$ and bandwidth $\bandwidth_i \in \rplus$. The latter controls its area of influence in the feature space. All these parameters are learnable and are optimized during training along with the blending weights $\weights$. 
Experimentally, we found that the polynomial term $\polynomial$ does not influence the capacity of the RBF network in representing SDF fields. Thus, we drop this term for 3D geometry representation.

\noi\textbf{(3) Training.} We train the entire pipeline using the $\lone$ loss between the estimated SDF $\signal$ and the ground truth $\text{SDF}_{\text{gt}}$:
\begin{equation}
    \loss_{\text{SDF}} = \frac{1}{\npoints} \sum_{i=1}^{\npoints} \frac{1}{\left| \text{SDF}_{\text{gt}}(\point_i) \right| +  \epsilon } \left| \signal(\point_i) - \text{SDF}_{\text{gt}}(\point_i) \right|. 
    \label{eq:sdf_loss}
\end{equation}

\noi Here, $\npoints$ is the number of training points and $\epsilon$ is a small real number, which allows scaling the loss to focus more on areas near the surface. During training,  the network weights,  the RBF parameters, and the feature vectors are jointly optimized. To efficiently train the model, we sample more training points near the surface than far from the surface. Following DeepSDF~\cite{park2019deepsdf},  $80\%$ of the training points are sampled from the shape surface. Half of these points are slightly perturbed in random directions. The remaining $20\%$   are randomly sampled from the normalized 3D space. We use the Adam optimizer, with gradients estimated with respect to the trainable parameters.  
At the start of the training, we sample $\nbases$ points $\{\thecenter_i \in \rthree\}_{i=1}^{\nbases}$ and initialize the centers $\meanfeaturevector_i \in \real^\featuredim, i=1, \dots, \nbases$ of the RBFs as the feature vectors at these points, \ie $\meanfeaturevector_i  = \enc(\thecenter_i)$. The RBF centers are then optimized during training in an auto-decoding fashion.

\noi\textbf{(4) Hierarchical modeling.} Our formulation naturally supports a multiscale representation of 3D geometry. In particular, the first few dimensions of the feature vector $\featurevector$ encode coarse shape information, while the latter dimensions capture higher-frequency geometric details. This is illustrated in Figure~\ref{fig:multiscale}. Unlike standard grid-based methods that pass the full feature vector directly to the decoder, our approach compares feature vectors to kernel centers via the Euclidean distance in the feature space. Since this distance metric only requires the vectors to be of matching dimensionality, we can selectively slice the feature vector to control the level of detail without modifying the decoder. This enables direct extraction of geometry at different resolutions, making the number of levels a tunable hyperparameter of the encoding.

\subsection{GNF for RGB image representation} 
\label{sec:RGB_image-RBF}

Next, we focus on fitting and representing multi-dimensional signals, \ie   $\inputdim \geq 1$ and $\outputdim \geq 1$,  by leveraging the multi-channel capability of the proposed GNF. An example of such signals is RGB  images, which can be seen as signals of the form $\signal: [0, 1]^2 \times  [0,1]^3 $ where   $\inputdim = 2$ and $\outputdim=3$. They map a pixel $\point$ to its RGB color.  
Similar to the SDF representation, we use a hierarchical feature grid-based input encoding where the trainable codes are arranged in a multiresolution 2D grid. The final feature vector at a given pixel $\point \in [0,1]^2$ is of dimension $\featuredim =16$. We decode the feature $\featurevector \in \real^\featuredim$   using a GNF composed of $64$ Gaussian kernels. Similar to the SDF case, we found that using spherical Gaussian kernels is sufficient to efficiently represent complex images.

We train the model using the $\ltwo$ loss between the estimated colors $\estimated{\signal}$ and the ground truth colors $\signal$, \ie 
$
    \loss_{\text{RGB}} = \frac{1}{\npoints} \sum_{i=1}^{\npoints} \left| \estimated{\signal}(\point_i) - \signal(\point_i) \right| ^ 2 
$. Here,  $\npoints$ is the number of training pixels. During training,  the network weights,  the RBF parameters, and the feature vectors are jointly optimized. At the start of the training, we regularly sample $\nbases$ pixels $\{\thecenter_1, \dots, \thecenter_\nbases\}$ in the image space and initialize the centers $\meanfeaturevector_i \in \real^\featuredim, i=1, \dots, \nbases$ of the RBFs as the feature vectors at these pixels, \ie $\meanfeaturevector_i  = \enc(\thecenter_i)$. The RBF centers are then optimized, jointly with the other trainable parameters, in an auto-decoding fashion.

\subsection{GNF for Radiance Field reconstruction}
\label{sec:radiance_field_RBF}

Finally, we demonstrate how the proposed \ModelName{} can be used to represent radiance fields and synthesize novel views. A radiance field is defined as a function $\signal$ of the form:
\begin{equation}
    \begin{aligned}
    \signal: \rthree \times \stwo \to \real^4, & \quad
    \point \mapsto (\density, \geometrylatentcode) = \geometrydec(\enc(\point)), \\
    &  \quad \geometrylatentcode, \viewdir \mapsto \thecolor = \colordec(\geometrylatentcode, \enc(\viewdir)).
    \end{aligned}
\end{equation}

\noi Here, $\density$ is the volume density at $\point \in \rthree$, $\thecolor$ is the RGB color viewed from direction $\viewdir \in \stwo$,  $\geometrylatentcode$ is a latent geometry code, and $\geometrydec$ and $ \colordec$ are, respectively, the geometry and color decoders. Although $\viewdir \in \stwo$, we represent it as a unit vector in $\rthree$.
We use a hierarchical hash grid encoder with $32$ resolution levels and one 1D feature per level, yielding a $32$D encoding of $\point$. We also experiment with a Tri-Plane encoder with resolutions $128$, $256$, and $512$, each with $32$ features, and report the results in Section~\ref{sec:results_radiance_field}. 

Following~\cite{yu2021plenoctrees}, we model view-dependent color using Spherical Harmonics (SH). Thus, the geometry decoder outputs the volume density $\density$ and $16$ SH coefficients per RGB channel, resulting in $1 + 16 \times 3 = 49$ outputs (see Figure~\ref{fig:nerf_architecture}).
The advantage of this design lies in the decoder: both geometry and appearance are predicted using a single-layer \ModelName, enabling fast inference through a single forward pass after encoding. It also enables fast training since the gradient only needs to be propagated through a single-layer network. The network is trained without 3D supervision using an $\ell_1$ loss between predicted and ground-truth RGB values, averaged over the image. In all our experiments, the RBF centers $\boldsymbol{\mu}$ are initialized randomly from a uniform distribution. The scale parameters $\boldsymbol{\beta}$, which are real-valued, are all initialized to $1.0$.

\begin{table*}[!t] 
    \centering    
    { 
    \begin{tabular}{@{}lcccccc@{}}
    \toprule
    \textbf{Model} & \textbf{CD-L1 (↓)} & \textbf{NC (↑)} & \textbf{NAE (↓)} & \textbf{IoU (↑)} & \textbf{Tr. Time} & \textbf{Parameters} \\
    \midrule
    
    \textbf{\ModelName \ (Ours)}          & \cellcolor{green!20}0.0026 $\pm$ 0.0019 & \cellcolor{green!20}0.9429 $\pm$ 0.0634 & \cellcolor{green!20}11.996 $\pm$ 7.9340 & \cellcolor{green!20}0.9628 $\pm$ 0.0404 & \cellcolor{green!20}15 secs & \cellcolor{orange!20}3,840,108 \\
     \midrule
    
    NeuRBF~\cite{chen2023neurbf}       & 0.0057 $\pm$ 0.0060 & 0.8950 $\pm$ 0.1296 & 16.940 $\pm$ 14.727 & 0.9056 $\pm$ 0.1140 & 54 secs & \cellcolor{green!20}950,409 \\
    
    FactorFields~\cite{chen2023dictionary} & 0.0041 $\pm$ 0.0034 & 0.9162 $\pm$ 0.0904 & 15.637 $\pm$ 10.645 & 0.9318 $\pm$ 0.0684 & 20 secs & 5,342,274 \\
    
    InstantNGP~\cite{muller2022instant}   & \cellcolor{orange!20} 0.0029 $\pm$ 0.0024 & \cellcolor{orange!20}0.9350 $\pm$ 0.0771 & \cellcolor{orange!20}12.574 $\pm$ 9.6156 & \cellcolor{orange!20}0.9598 $\pm$ 0.0515 & 21 secs & 6,125,048 \\
    \bottomrule
    \end{tabular}
    }
    \caption{Comparison of the performance of the proposed method in reconstructing and representing SDFs against three state-of-the-art methods. The best values are highlighted in  \textcolor{green}{green} and the second-best in \textcolor{orange}{orange}. All methods have been trained using  $8$ million query points. Here, we report the average and standard deviation of various performance metrics averaged over four shapes from the Stanford 3D Scanning Repository~\cite{stanford3d}; see Figure~\ref{fig:comparison_grid}.}    
    \label{tab:all_models}
\end{table*}

\section{Results}
\label{sec:results}

We evaluate the performance of the proposed framework in representing nD signals such as 3D geometry, RGB images, and radiance fields. For each case, we compare performance against the state-of-the-art approaches in terms of representation accuracy, training and inference times, and compactness of the representation. All experiments were conducted on a single NVIDIA RTX 4090 GPU. High resolution results as well as the source code are available on the \href{https://grbfnet.github.io/}{project website: \url{https://grbfnet.github.io/}}.

\subsection{3D Reconstruction of Signed Distance Fields} 

We first evaluate the effectiveness of the proposed Gaussian Neural Fields in representing SDFs of 3D objects. We use $12$  3D models of varying complexities, from the Stanford 3D Scanning Repository~\cite{stanford3d} and the DiLiGenT-MV dataset~\cite{li2020multi}. The latter includes polygonal meshes that are partially open, introducing ambiguity in the SDF estimation in certain regions. These are challenging to the model's robustness in generating smooth and accurate SDFs.

\begin{table}[t] 
    \centering    
    \resizebox{\linewidth}{!}
    { 
    \begin{tabular}{@{}l@{ }cccc@{}}
    \toprule
    \textbf{Model} & \textbf{CD-L1 (↓)} & \textbf{NC (↑)} & \textbf{NAE (↓)} & \textbf{IoU (↑)} \\
    \midrule
    \textbf{\ModelName}          & \cellcolor{green!20}0.002 $\pm$ 0.002 & \cellcolor{green!20}0.982 $\pm$ 0.039 & \cellcolor{orange!20}5.469 $\pm$ 5.748 & \cellcolor{green!20}0.986 $\pm$ 0.027  \\
    \textbf{(Ours)}          &  &  & & \\
    \midrule
    
    NeuRBF       & 0.003 $\pm$ 0.004 & 0.969 $\pm$ 0.078 & 6.797 $\pm$ 9.648 & 0.969 $\pm$ 0.070  \\
    \cite{chen2023neurbf}\\
    \midrule
    
    InstantNGP   & \cellcolor{orange!20}0.002 $\pm$ 0.002 & \cellcolor{orange!20} 0.980 $\pm$ 0.047 & \cellcolor{green!20} 5.420 $\pm$ 6.621 & \cellcolor{orange!20}0.985 $\pm$ 0.032 \\
    \cite{muller2022instant} \\
    \bottomrule    
    
    \end{tabular}
    }
    \caption{Quantitative evaluation of the performance of the proposed method in reconstructing and representing Signed Distance Fields. The best values are highlighted in \textcolor{green}{green} and the second-best in \textcolor{orange}{orange}. All methods have been trained using  $8$ million query points. Here, we report the average and standard deviation of different performance metrics averaged over $12$ shapes from the Stanford 3D Scanning Repository~\cite{stanford3d} and the DiLiGenT-MV dataset~\cite{li2020multi}. Here the models share the same number of parameters with Table~\ref{tab:all_models}}
    \label{tab:partial_comparaison}
\end{table}

\begin{figure*}[!ht]
    \centering
   
    \resizebox{\linewidth}{!}{
    \begin{tabular}{@{}c@{ }c@{ }c@{ }c@{ }c@{ }c@{}}
    
        \centering
    
        \includegraphics[width=0.16\textwidth]{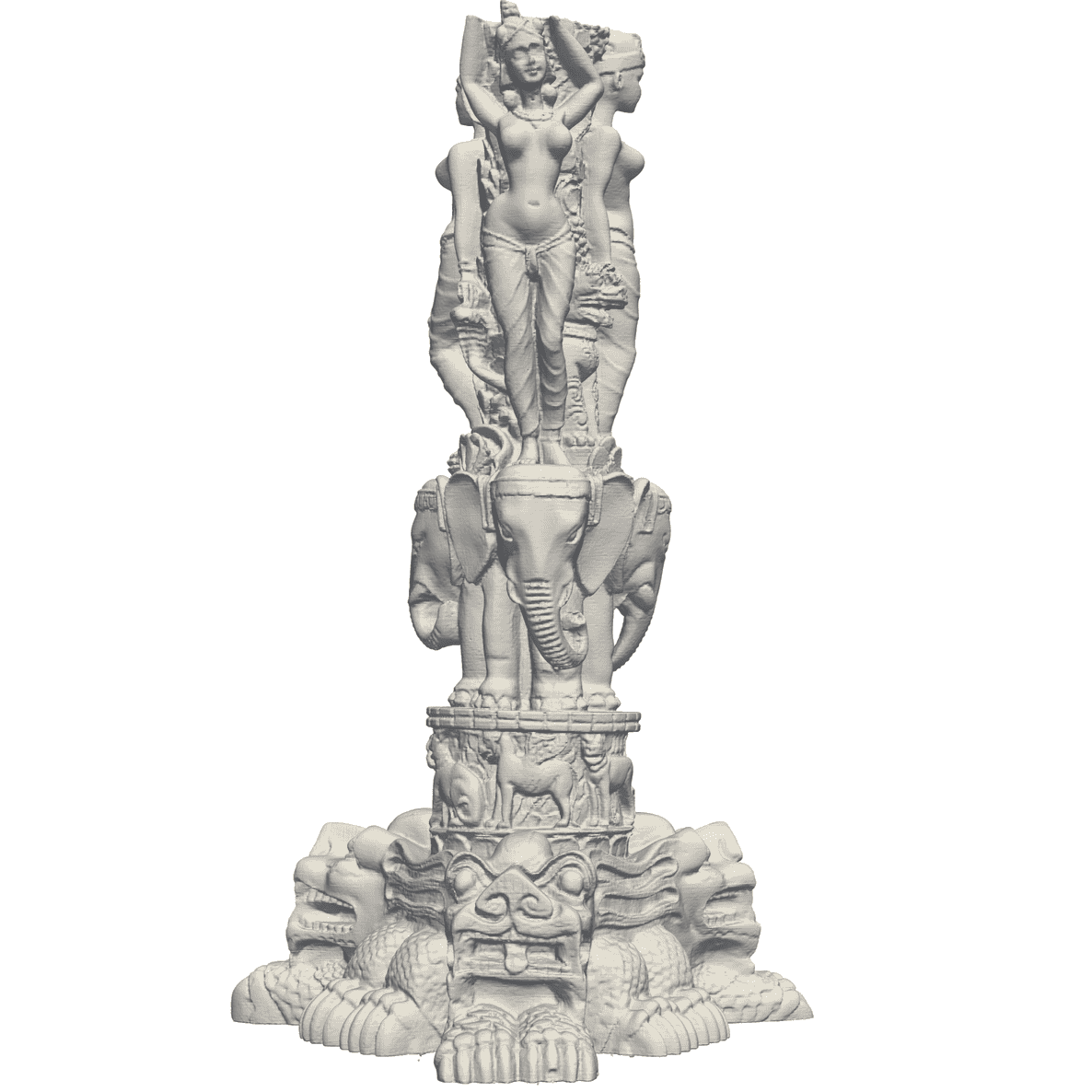} 
        &
        \includegraphics[width=0.16\textwidth]{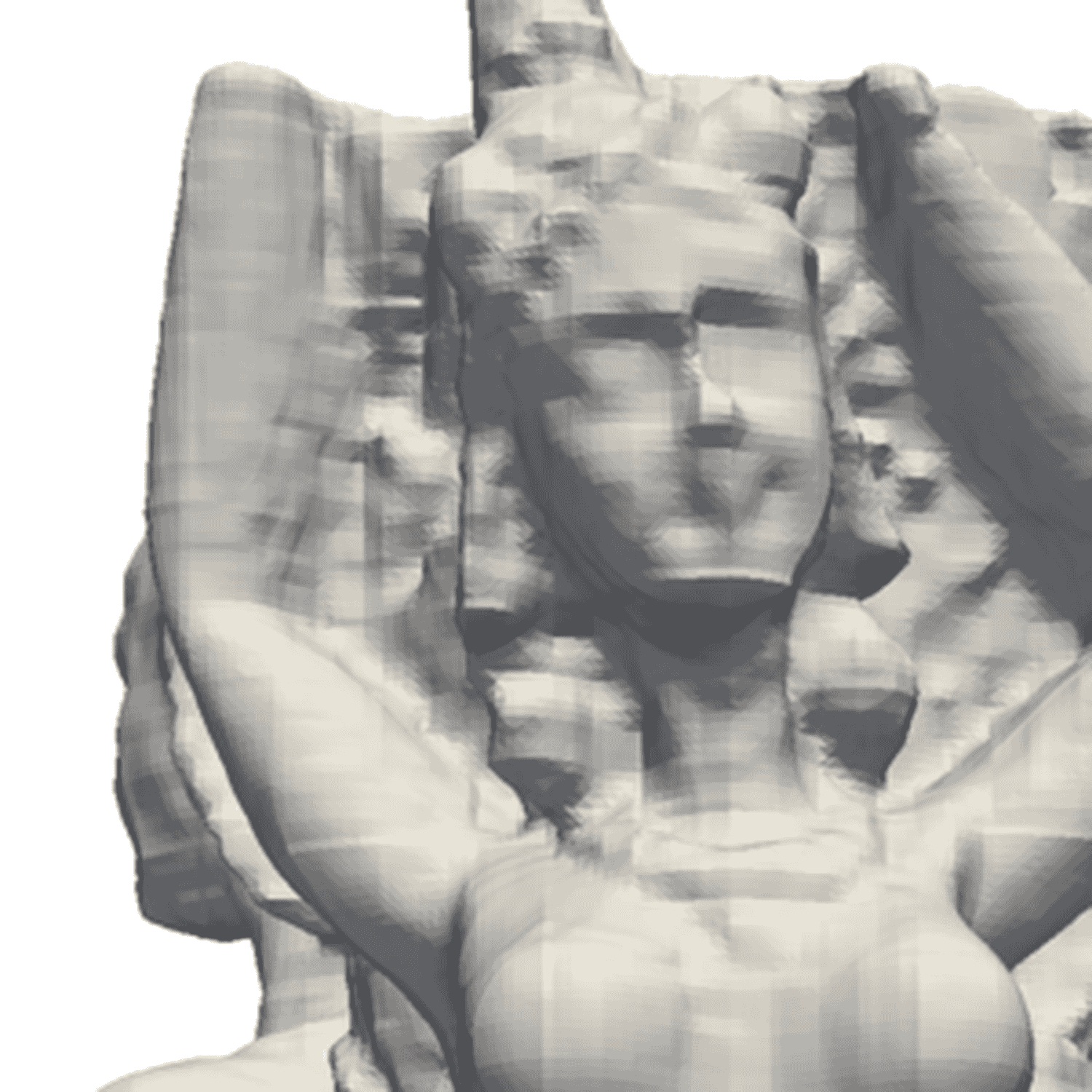} 
        &
        \includegraphics[width=0.16\textwidth]{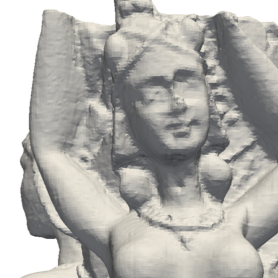} 
        &
        \includegraphics[width=0.16\textwidth]{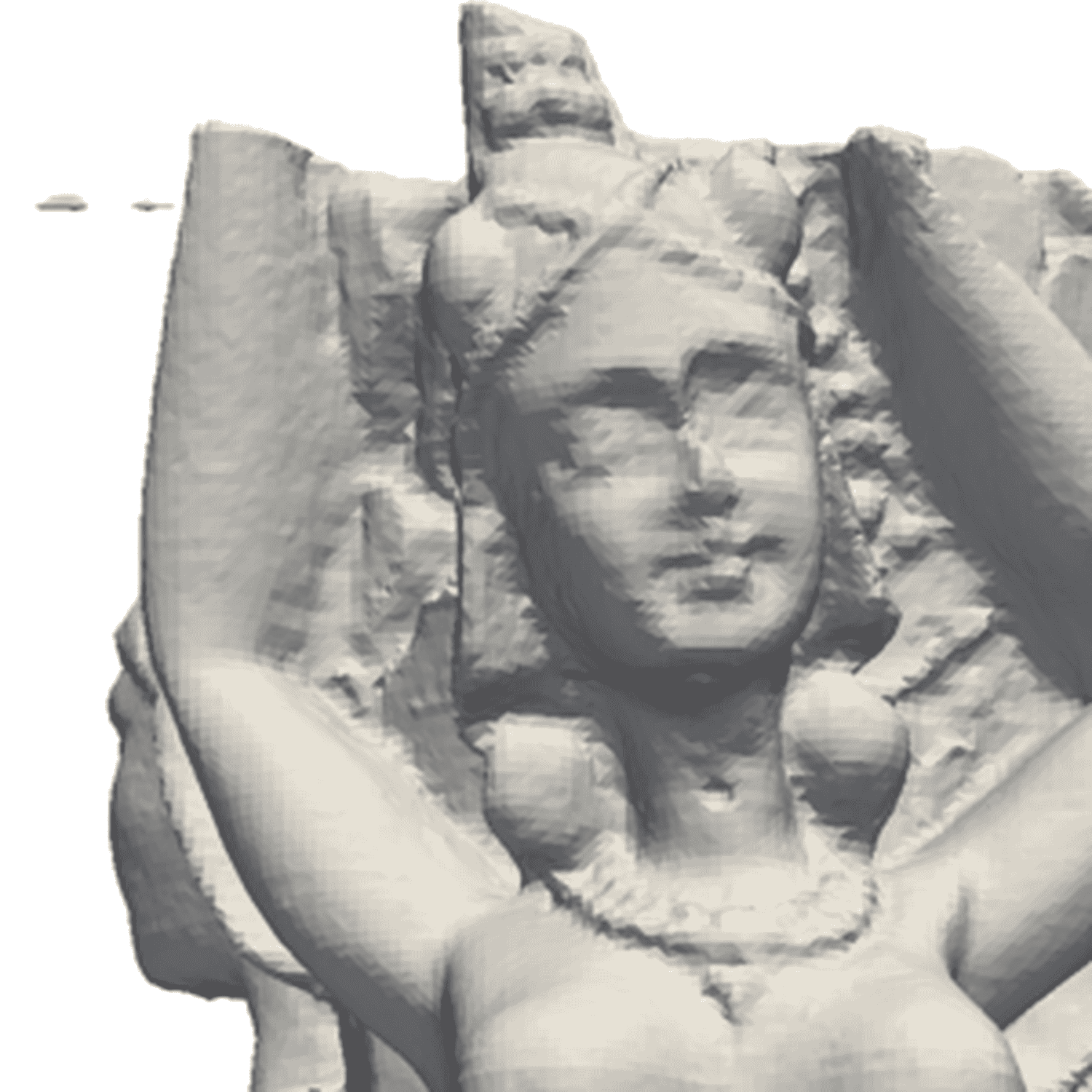} 
        &
        \includegraphics[width=0.16\textwidth]{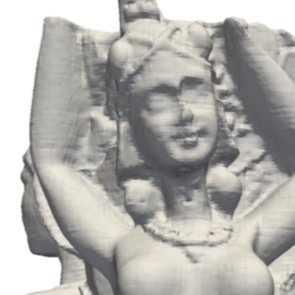} 
        &
        \includegraphics[width=0.16\textwidth]{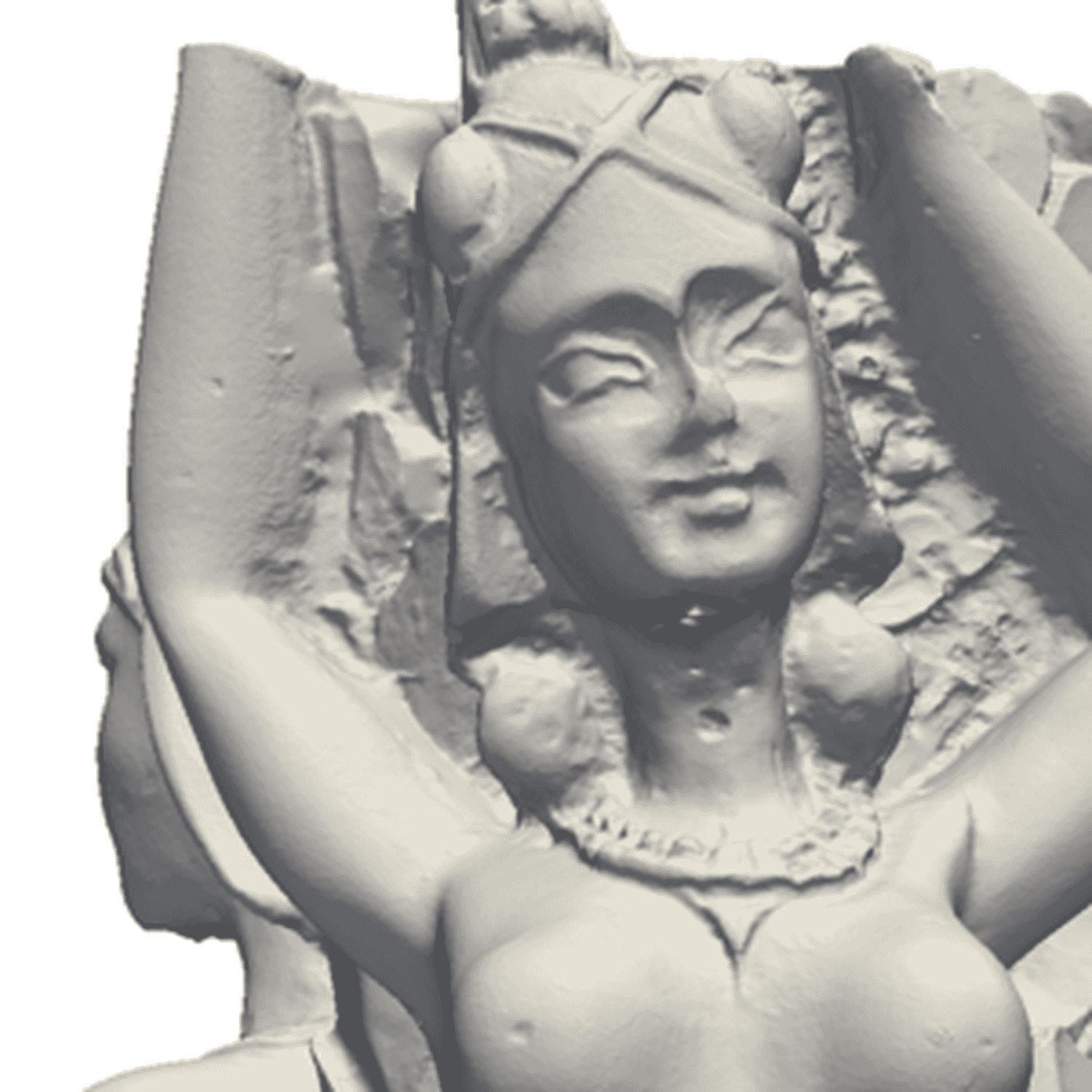}
        
        \\

        \includegraphics[width=0.16\textwidth]{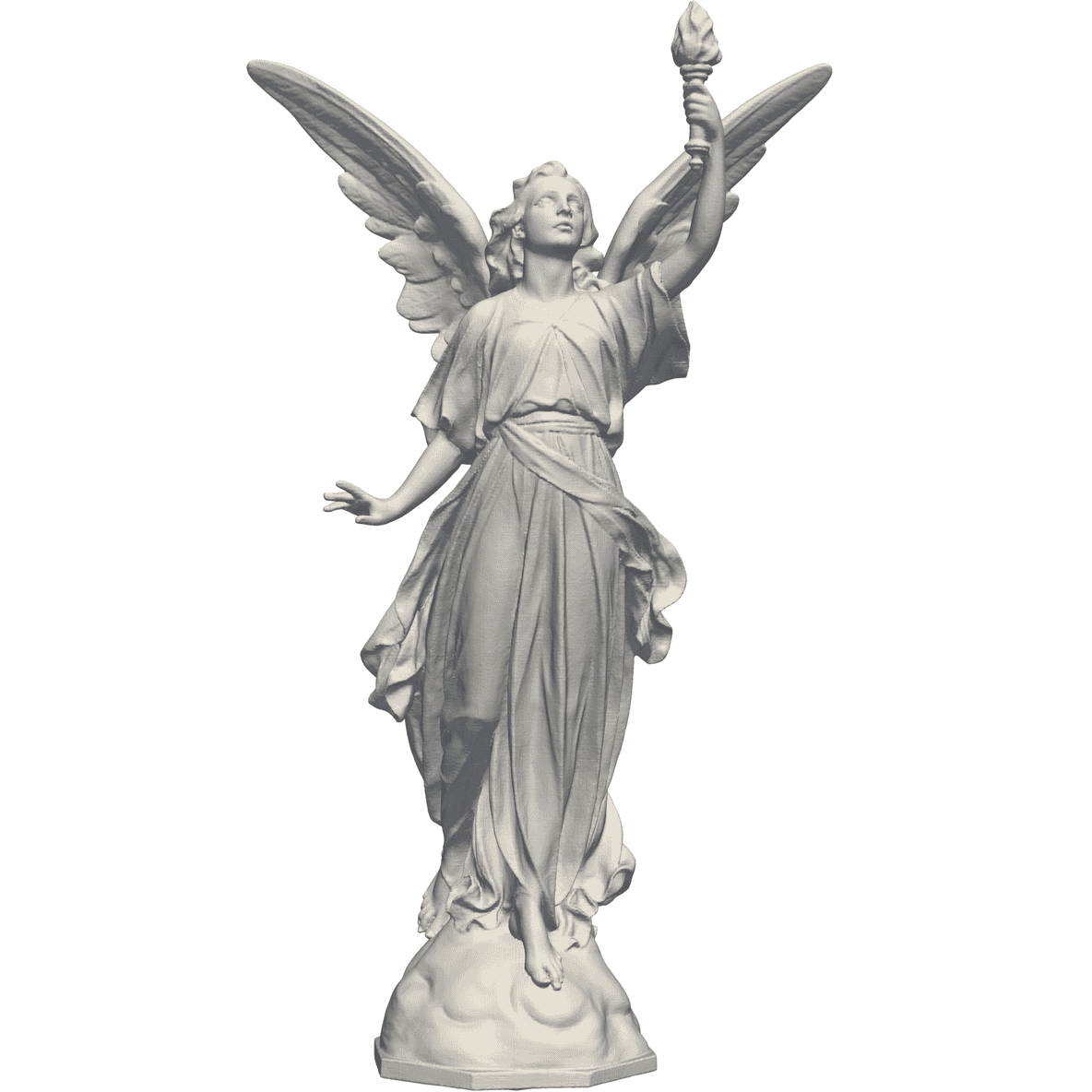} 
        &
        \includegraphics[width=0.16\textwidth]{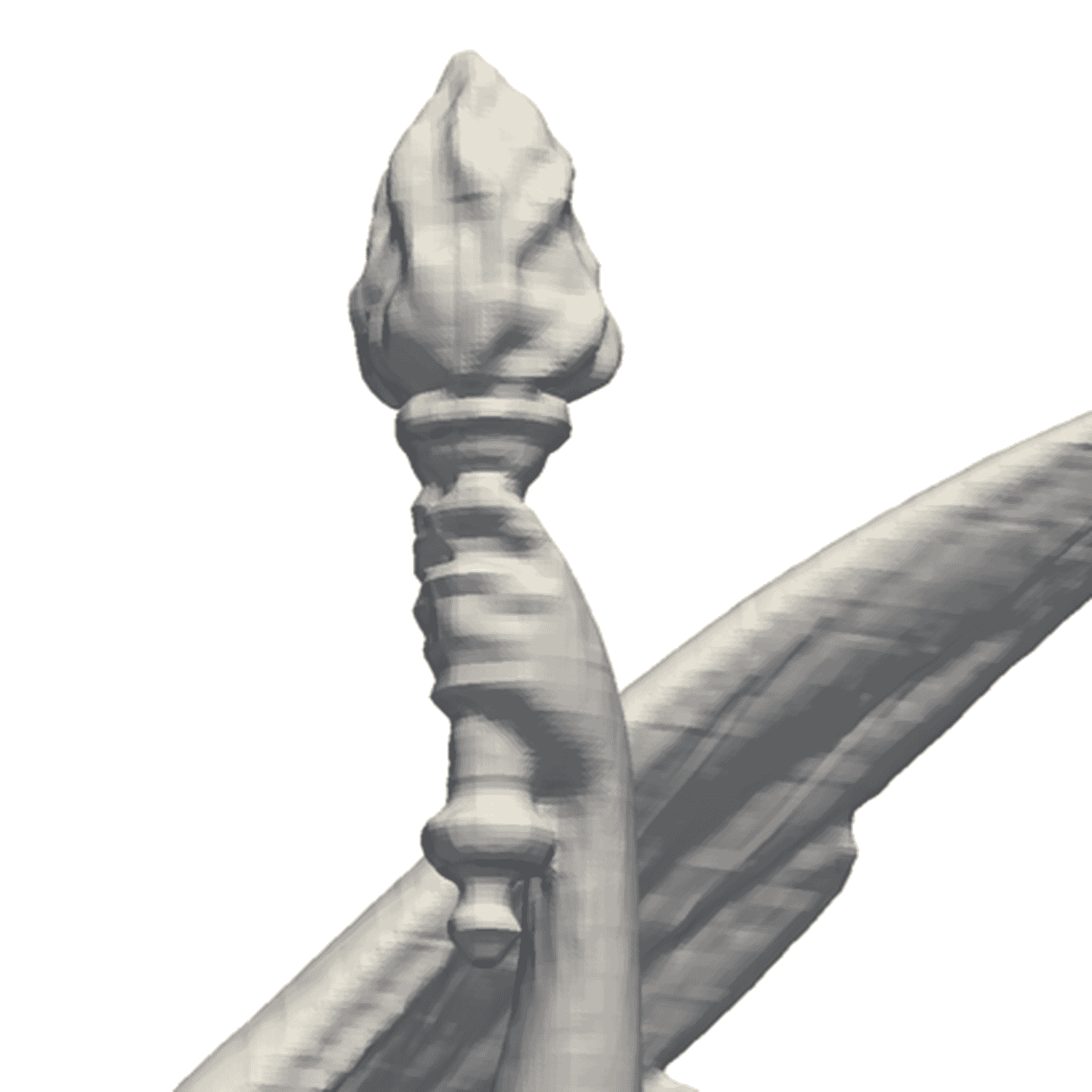} 
        &
        \includegraphics[width=0.16\textwidth]{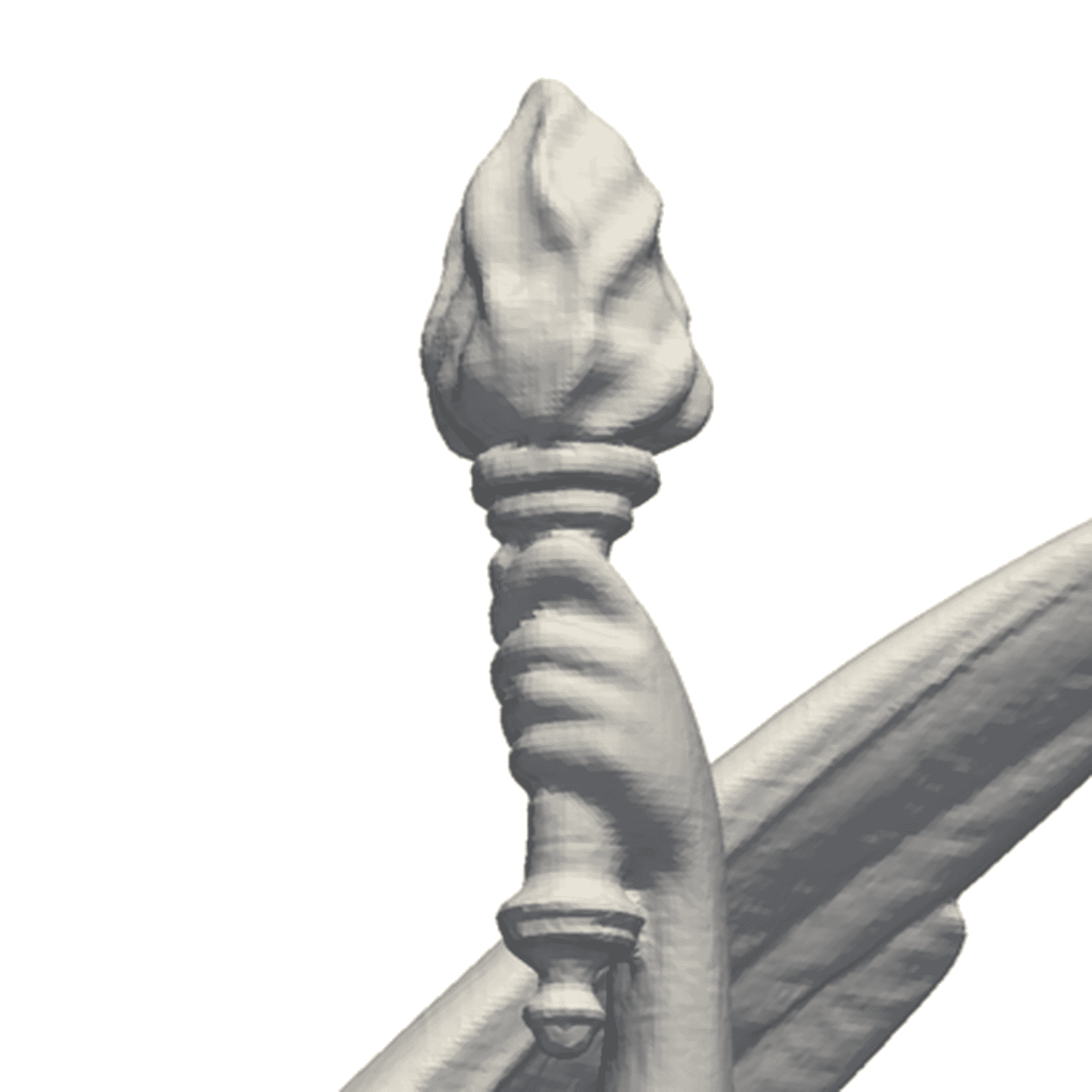} 
        &
        \includegraphics[width=0.16\textwidth]{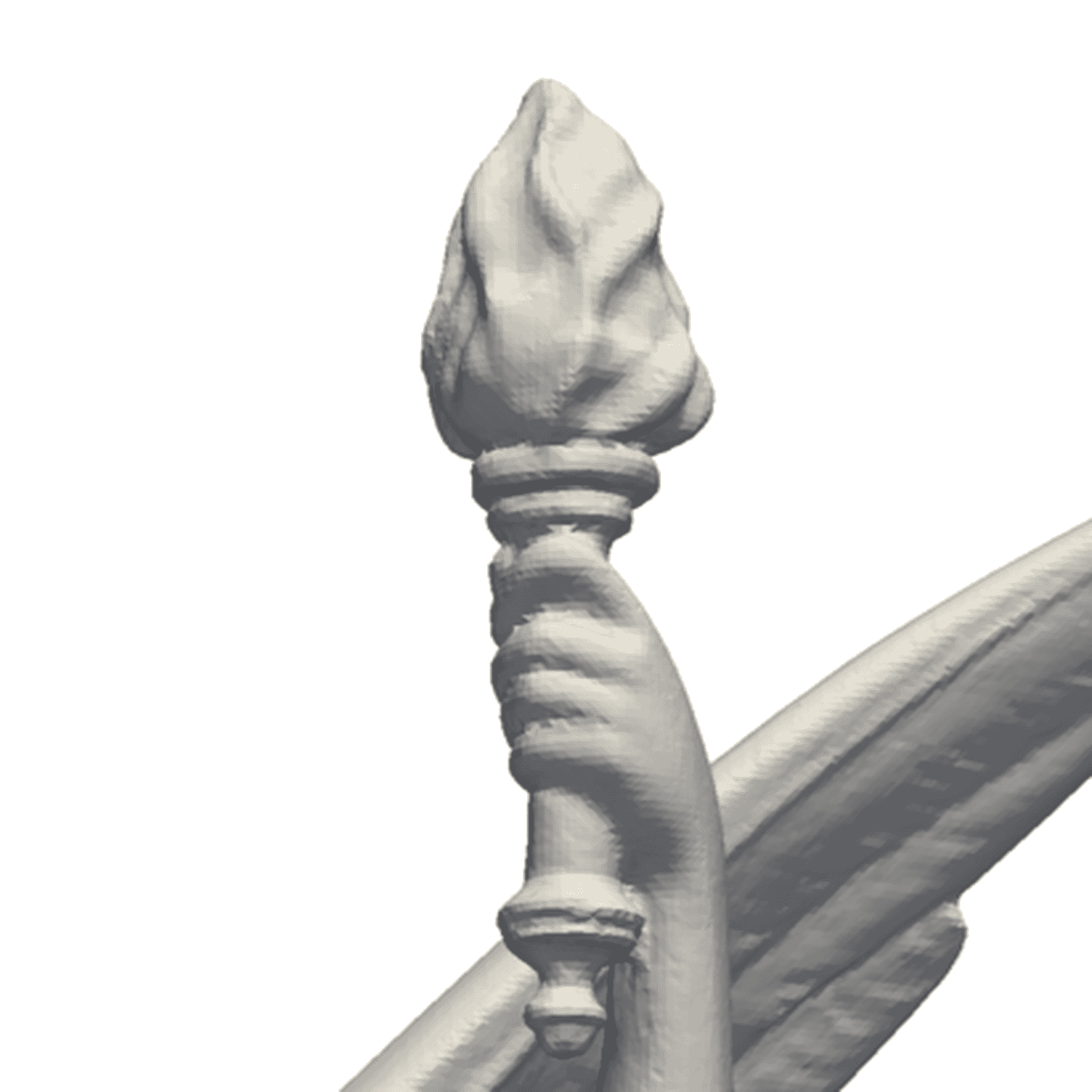} 
        &
        \includegraphics[width=0.16\textwidth]{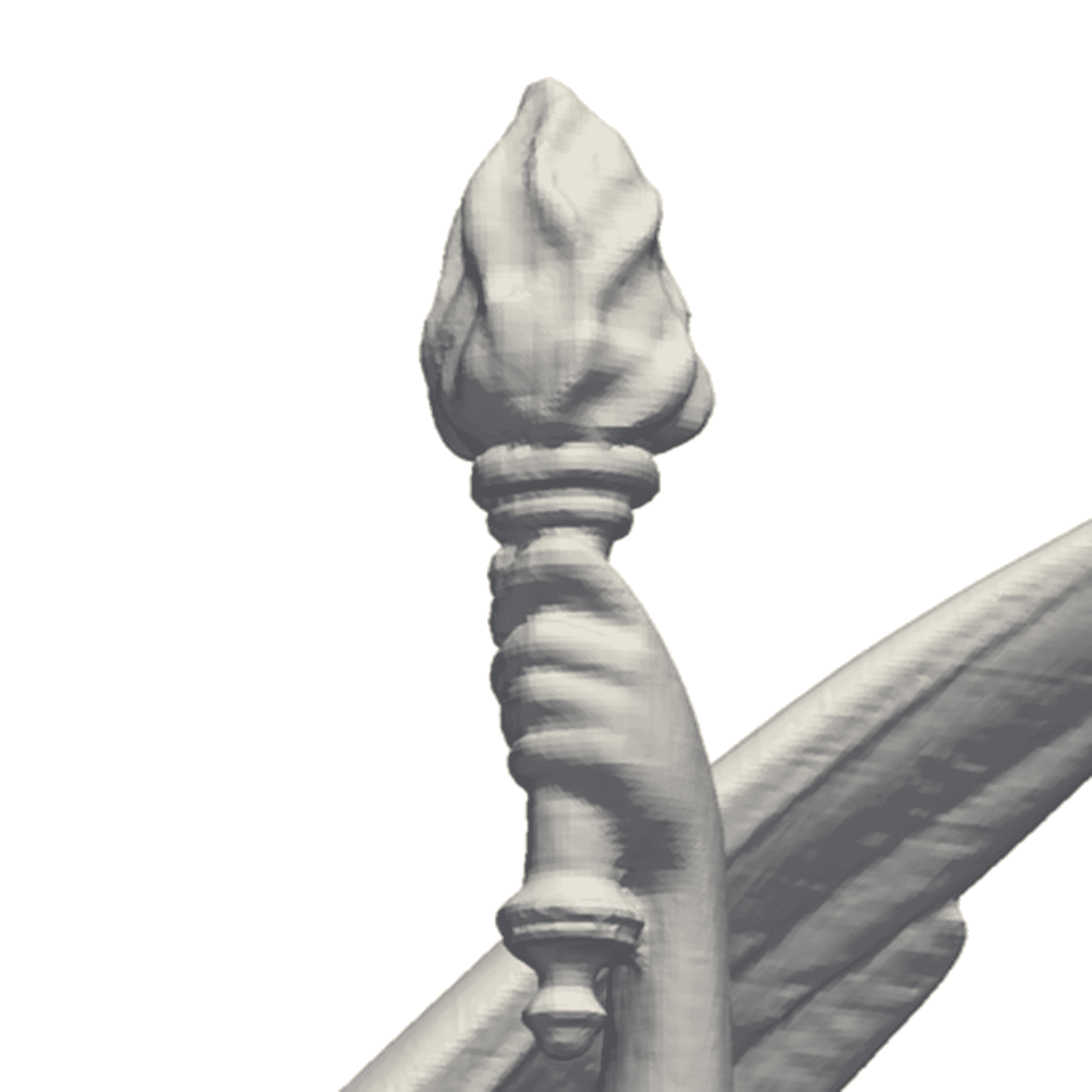} 
        &
        \includegraphics[width=0.16\textwidth]{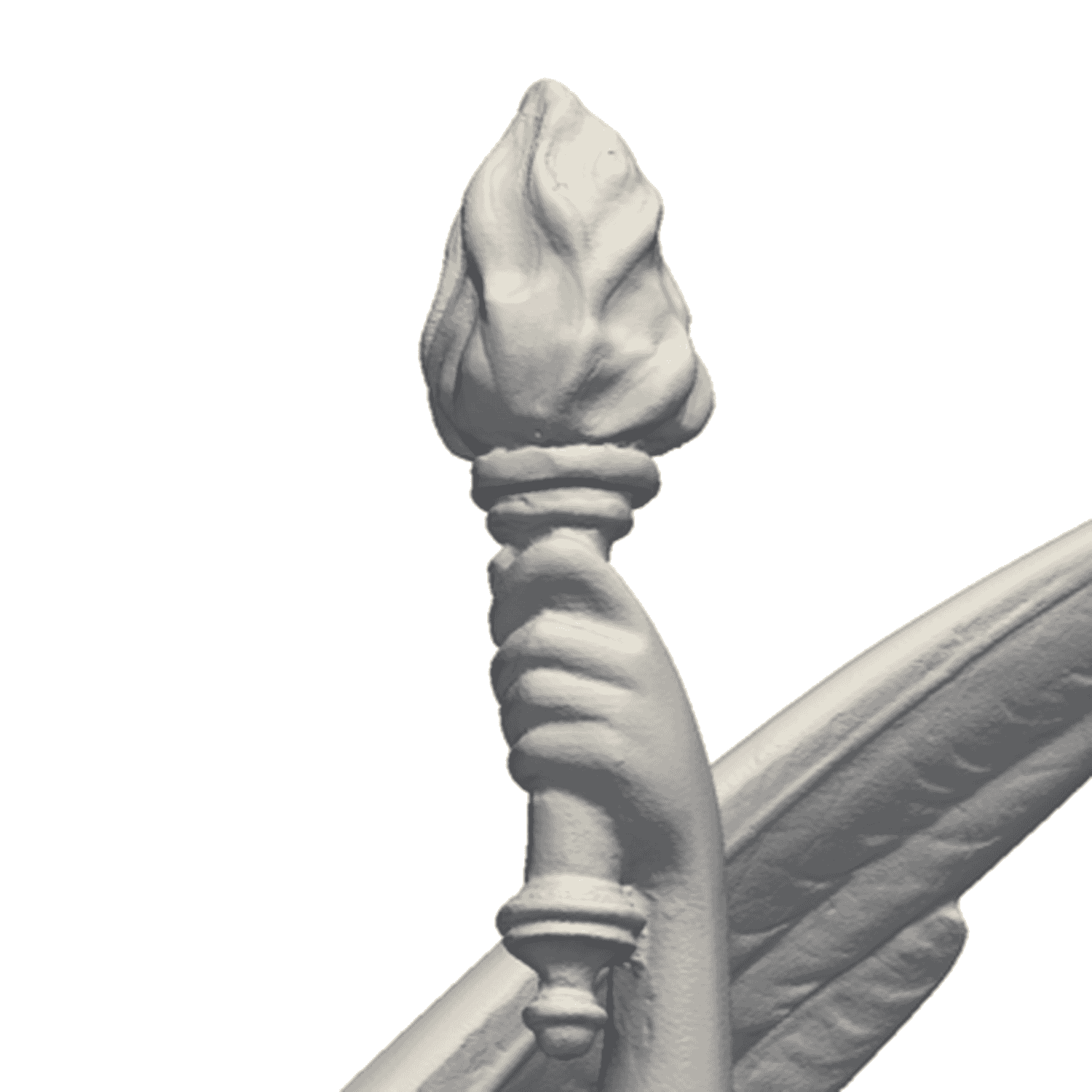}
        
        \\
        \ModelName{} (Ours) & FactorFields & NeuRBF& InstantNGP & \ModelName \ (Ours) & Groundtruth\\
     \end{tabular}   
}

     \resizebox{\linewidth}{!}{
    \begin{tabular}{@{}c@{ }c@{ }c@{}}
        \centering
        \includegraphics[width=0.33\textwidth]{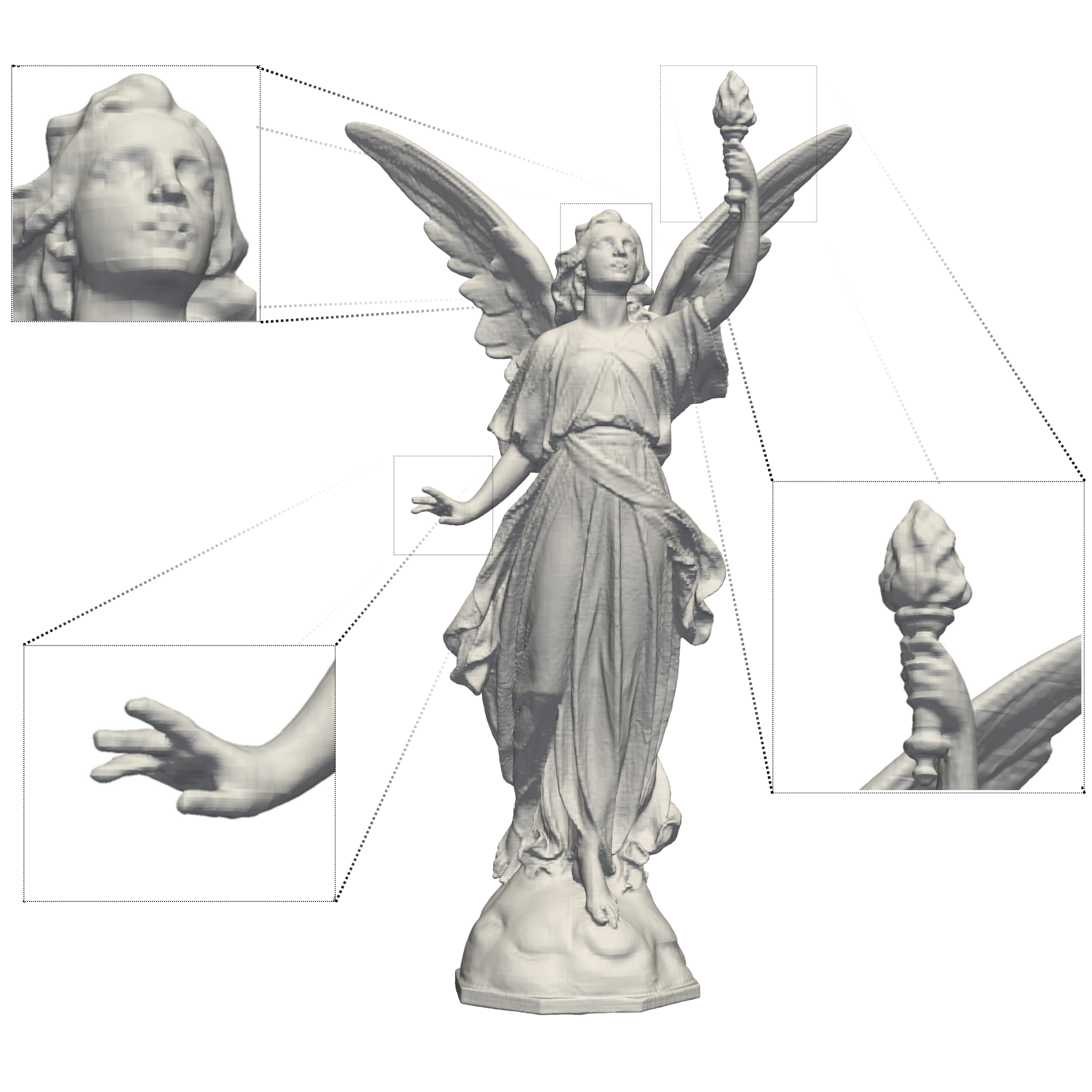} 
        &
        \includegraphics[width=0.33\textwidth]{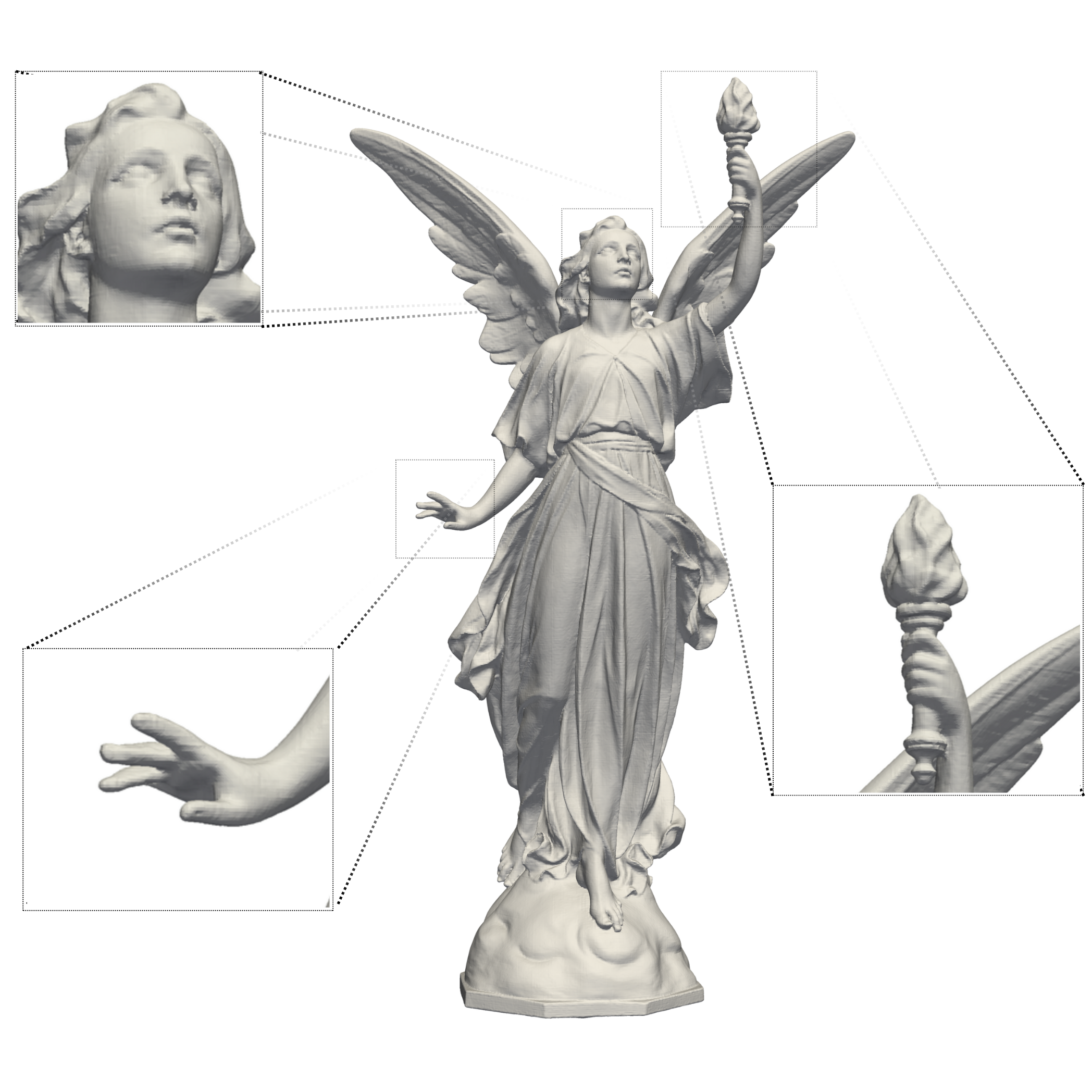} 
        &
        \includegraphics[width=0.33\textwidth]{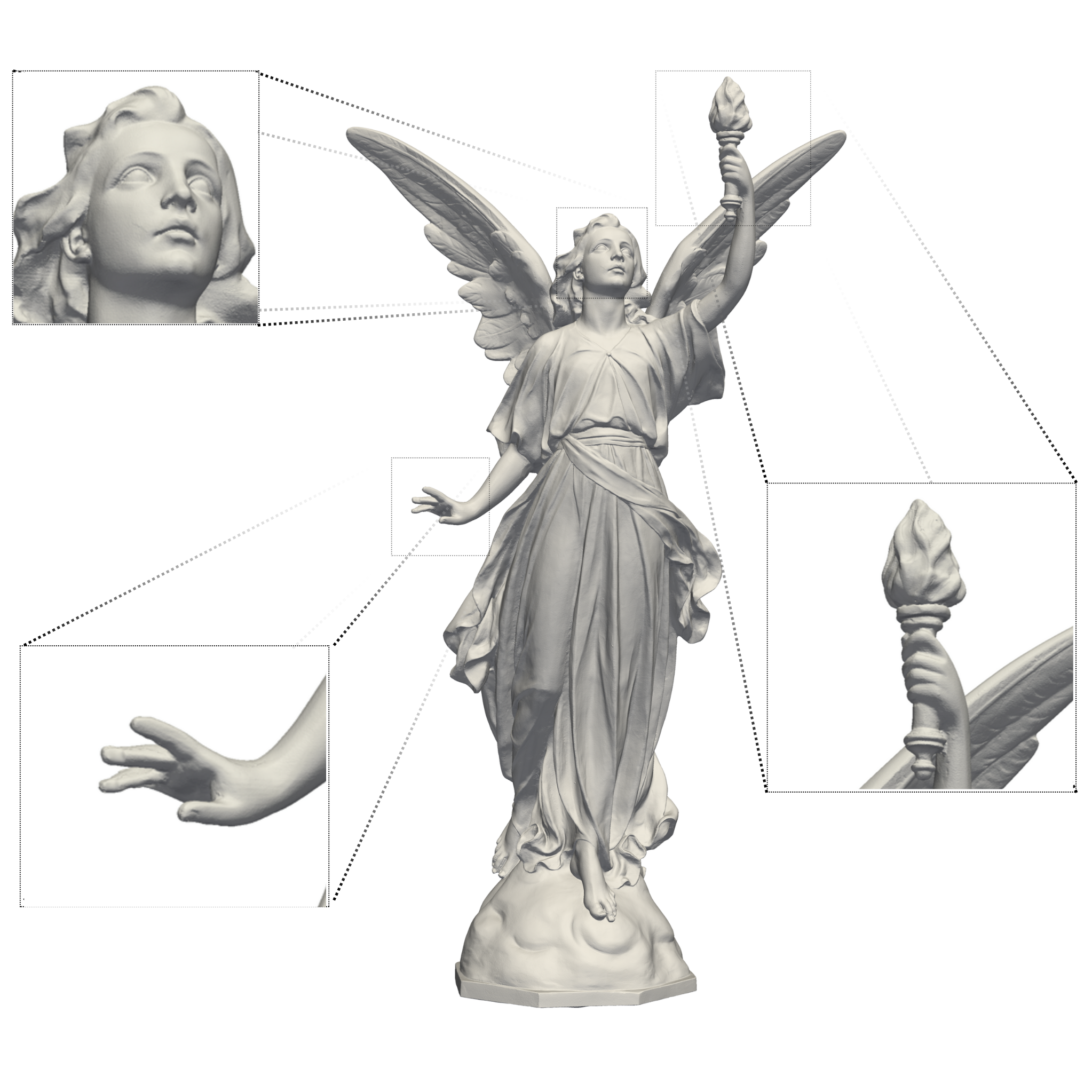} 
         \\
        FactorFields~\cite{chen2023dictionary}  & \ModelName \ (ours) &  Groundtruth
     \end{tabular}   
    }
    \caption{Comparison of the 3D geometry reconstruction quality.  In the first two rows, the leftmost column shows the  3D model reconstructed using our method. The subsequent columns provide zoomed-in views of the reconstructions from the compared methods and our own. The rightmost column shows the ground truth. The third row shows a close-up view of specific surface regions, highlighting our method's capability to capture fine geometric details with high accuracy.}
    \label{fig:comparison_grid}\label{fig:comparison_grid_zoom}
\end{figure*}

Table~\ref{tab:all_models} summarizes the performance of the proposed model and compares it to three state-of-the-art methods. For this quantitative evaluation, we use four meshes from the Stanford Dataset that contain challenging surface structures. For a fair comparison, we use the same number of training points across all methods. 
Our method achieves state-of-the-art performance across multiple evaluation metrics, including Chamfer Distance (CD-L1), Normal Consistency (NC), Normal Angular Error (NAE), and Intersection over Union (IoU). It maintains a compact model size of approximately $3.8$M parameters, compared to $5.3$M for FactorFields and $6.1$M for InstantNGP. Although NeuRBF uses significantly fewer parameters ($0.9$M), it exhibits the slowest training time ($50$ seconds on average). In contrast, our method is the fastest to train ($15$ seconds on average). During inference, our model can query approximately $49$ million 3D points per second, offering an excellent balance between compactness, speed, and accuracy.
As shown in Table~\ref{tab:partial_comparaison}, which summarizes results on $12$ shapes from the Stanford 3D Scanning Repository and DiLiGenT-MV dataset, our method outperforms the state-of-the-art on all metrics except on the NAE where InstantNGP performs slightly better than the proposed method. All the tested methods were trained using the same number of SDF training points, \ie $8$ million points per 3D shape. We also used the same learning rate scheduling, batch size, and number of training steps.
Figure~\ref{fig:comparison_grid}  provides a visual comparison between the extracted meshes from SDF values estimated by different methods, and with the ground truth mesh. We can see that our method can accurately reconstruct fine details; see the eyes of the angel statue (last row). Additional quantitative results and discussions are provided in Section 1 in the Supplementary Material

\begin{figure*}[!ht]
    \centering

    \resizebox{1\textwidth}{!}{
    \begin{tabular}{@{}c@{}c@{}c@{}}
        \centering
        Pluto & Summer Day & Girl with a Pearl Earring
        \\
        \includegraphics[height=0.18\textheight]{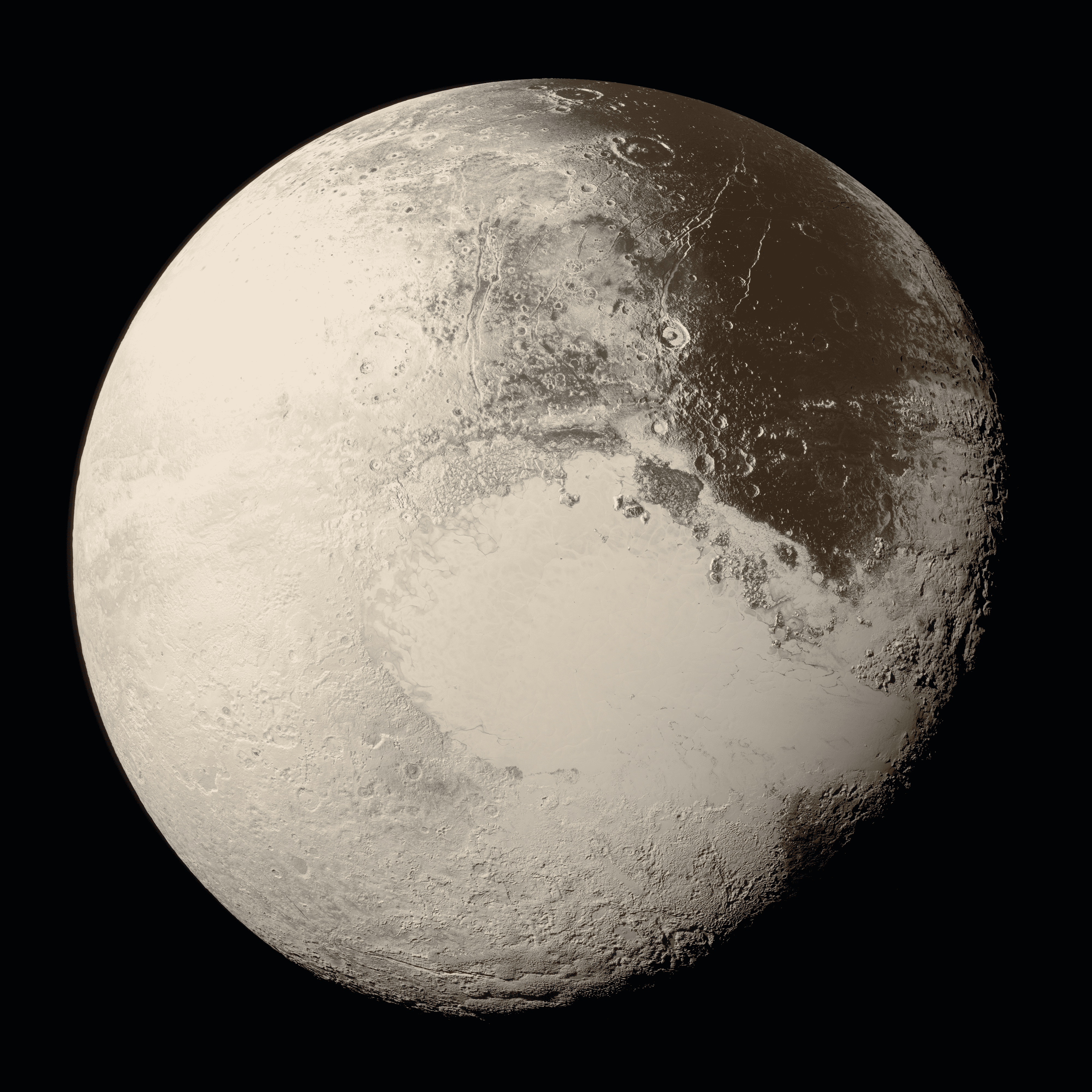} 
        &
        
        \includegraphics[height=0.18\textheight]{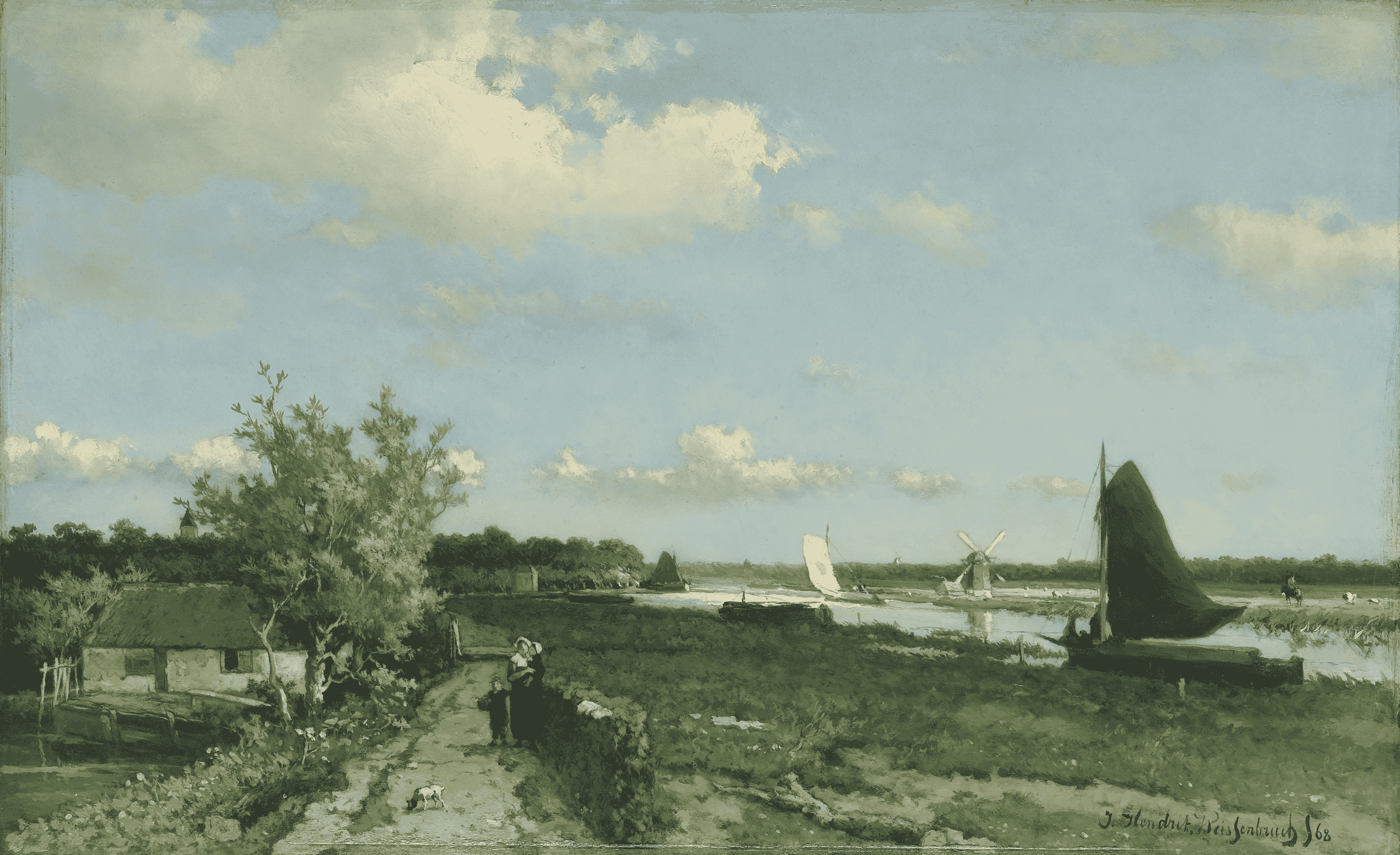} %
        &
        
        \includegraphics[height=0.18\textheight]{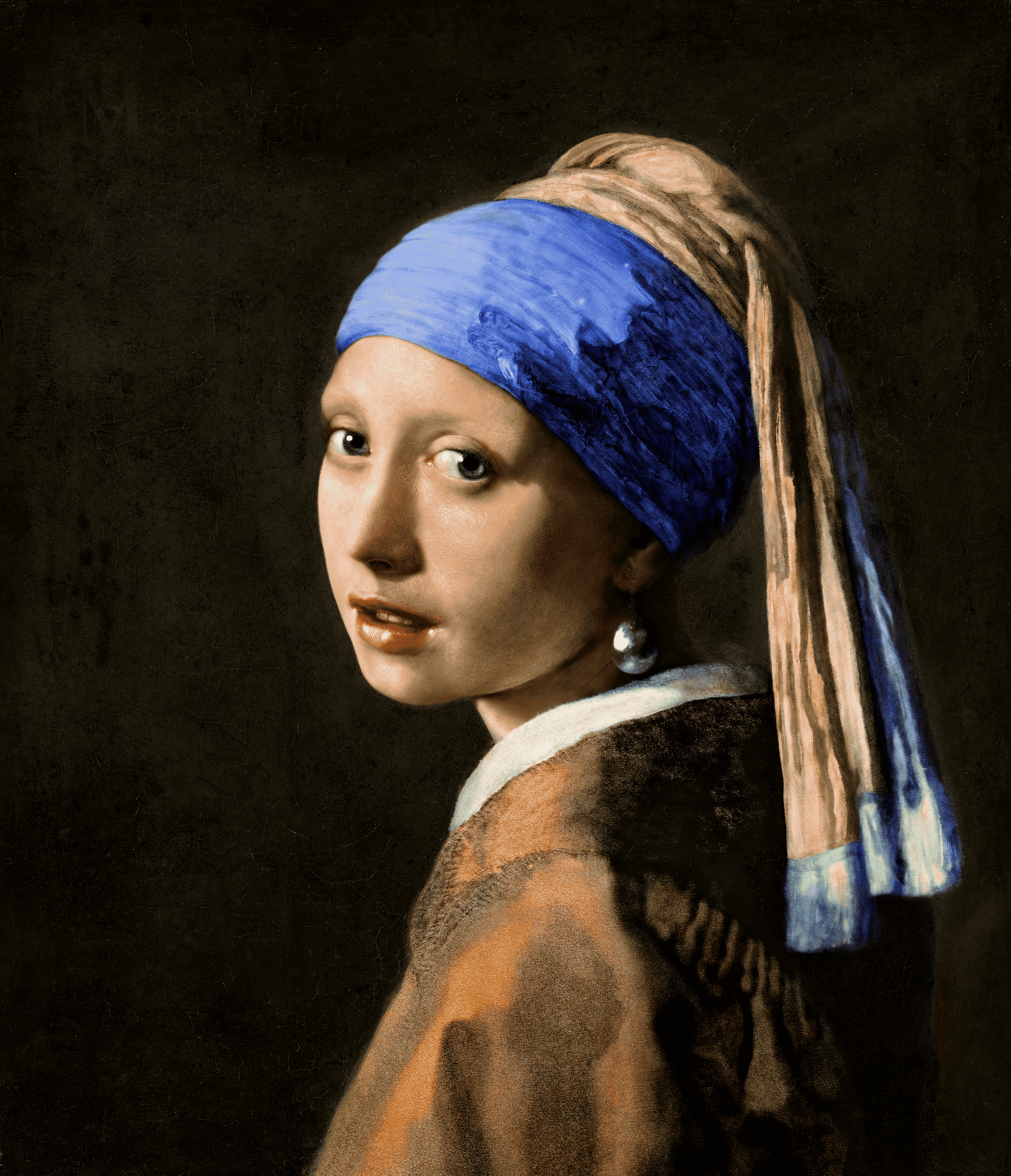} %
        \\
         \textcolor{green}{$8000 \times 8000 \times 3$} / \textcolor{blue}{$40.4$M} / 
         &  \textcolor{green}{$6114 \times 3734 \times 3$} /  \textcolor{blue}{$25.1$M}
         &  \textcolor{green}{$8000 \times 9302 \times 3$} /  \textcolor{blue}{$35.42$M}
         \\
            \textcolor{yellow}{$84$ secs}/ \textcolor{orange}{$44.41$ db} / \textcolor{red}{$18.8$M}& 
           \textcolor{yellow}{ $69$ secs} / \textcolor{orange}{$45.61$ db} / \textcolor{red}{$20.1$M} & 
            \textcolor{yellow}{$75$ secs} / \textcolor{orange}{$33.3$ db} / \textcolor{red}{$12.8$M} \\

       \\
        \begin{tabular}{cc}
        \includegraphics[height=0.15\textheight]{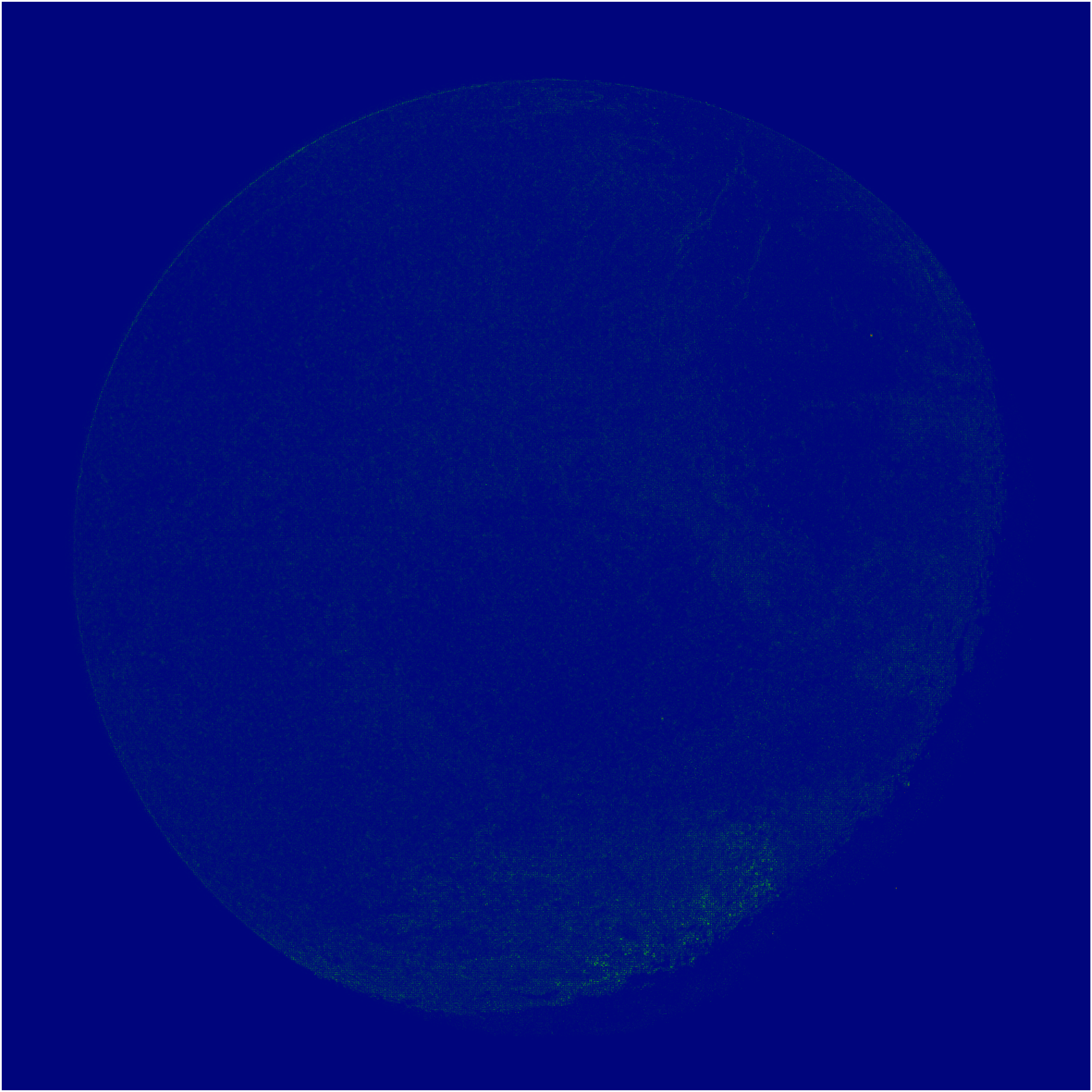} 
        &
        \includegraphics[height=0.2\textwidth]{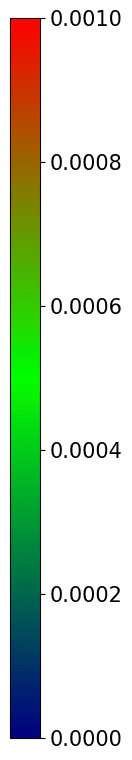}
        \end{tabular}
        &
         \begin{tabular}{cc}
        \includegraphics[height=0.15\textheight]{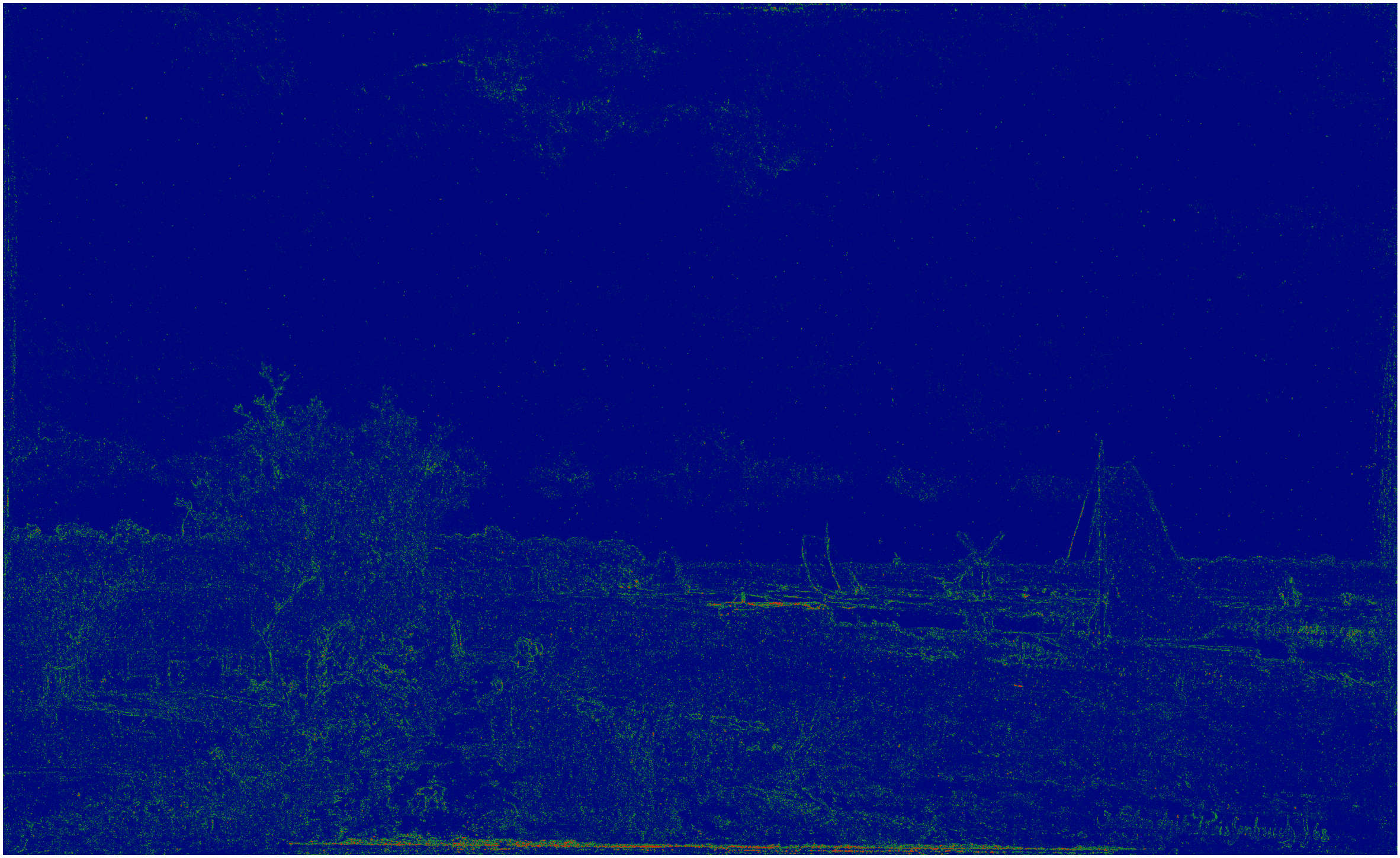} 
        &
        \includegraphics[height=0.2\textwidth]{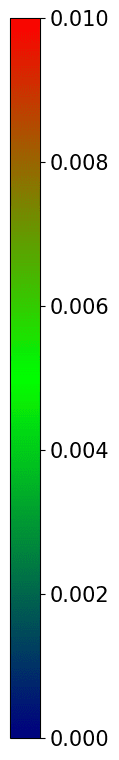}
        \end{tabular}
        &
        \begin{tabular}{cc}
        \includegraphics[height=0.15\textheight]{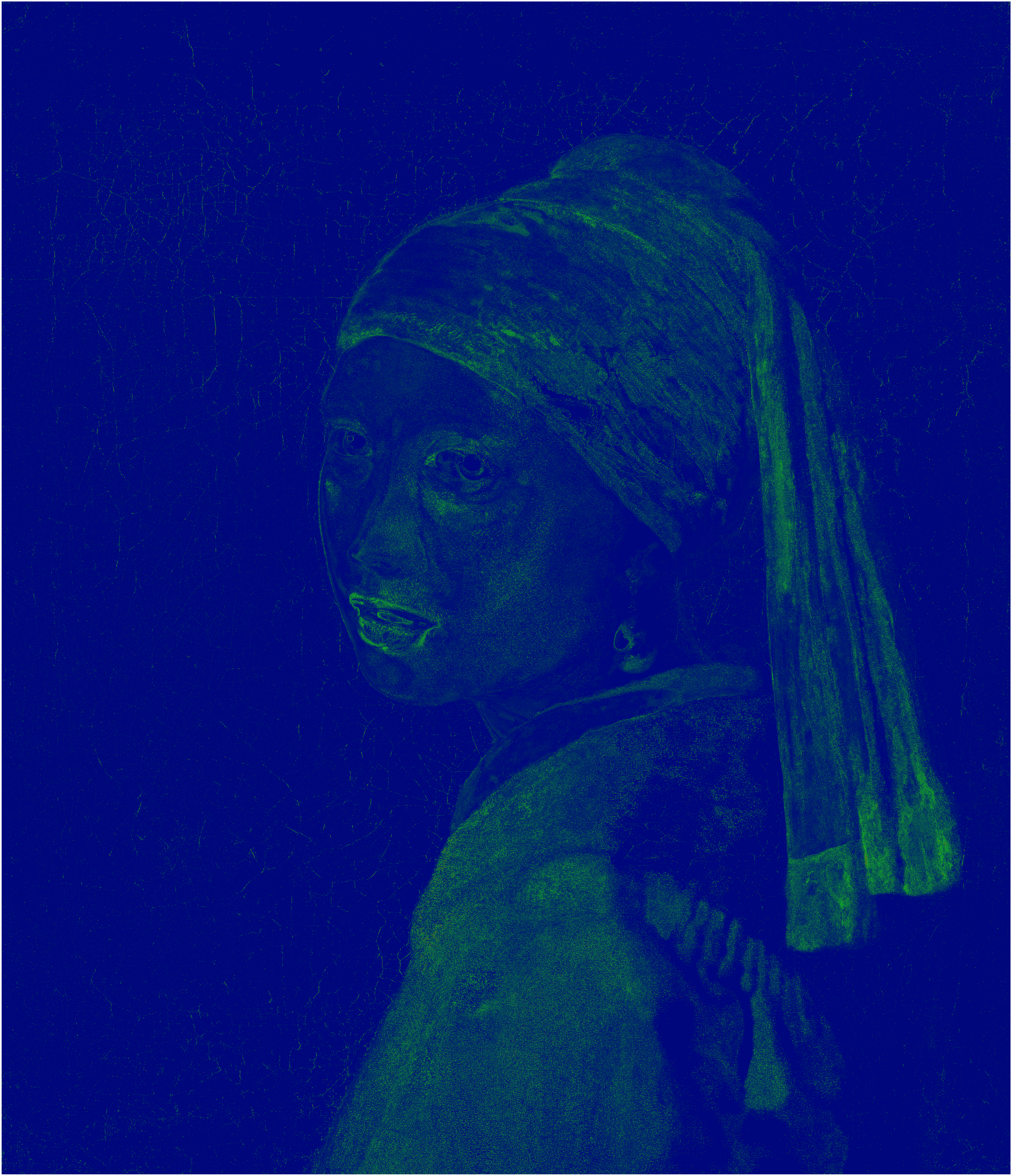} 
        &
        \includegraphics[height=0.2\textwidth]{figures/2d_reconstruction/colorbar.png}
        \end{tabular}

        \\
        
        \includegraphics[height=0.18\textheight]{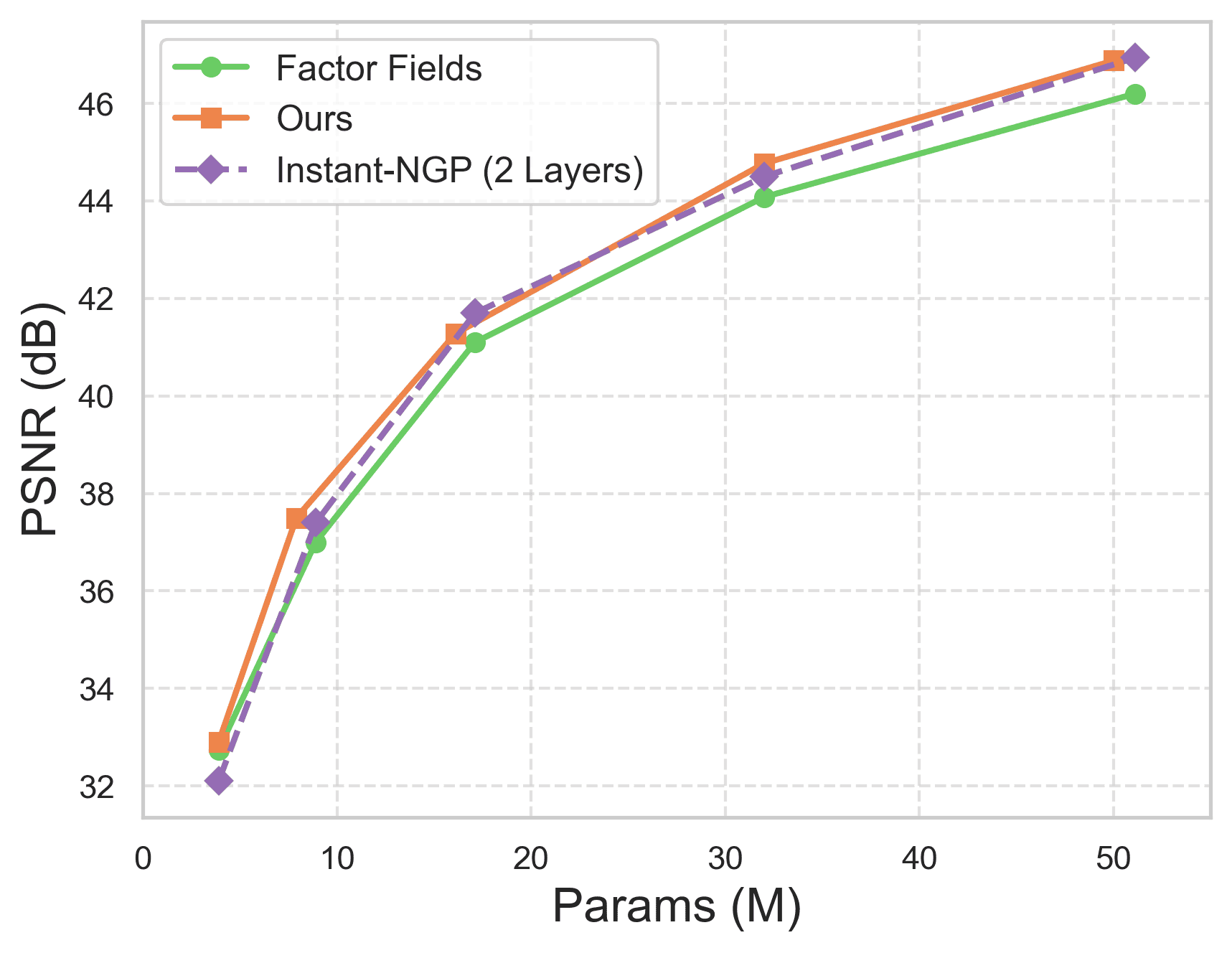} 
        &
        
        \includegraphics[height=0.18\textheight]{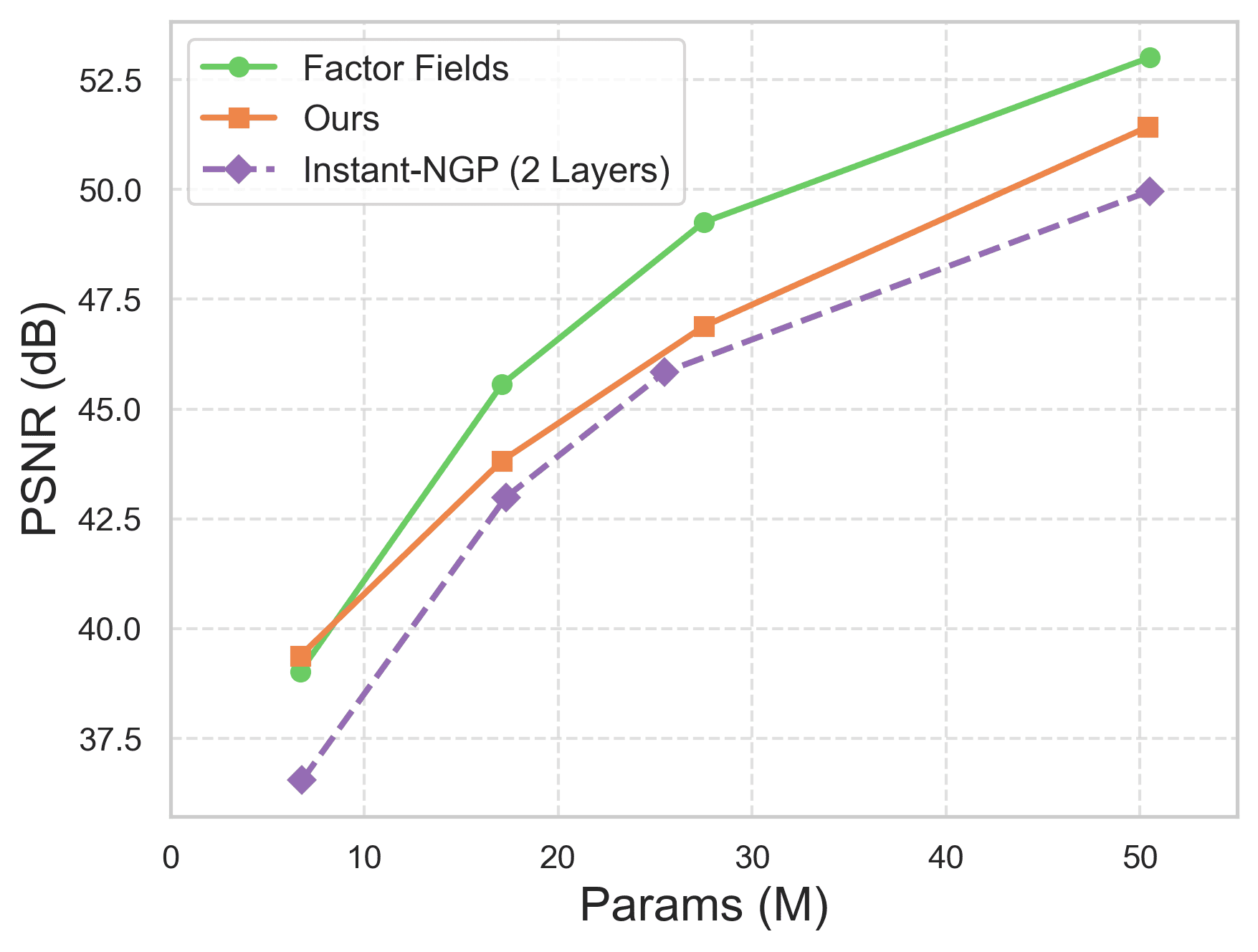} %
        &
        
        \includegraphics[height=0.18\textheight]{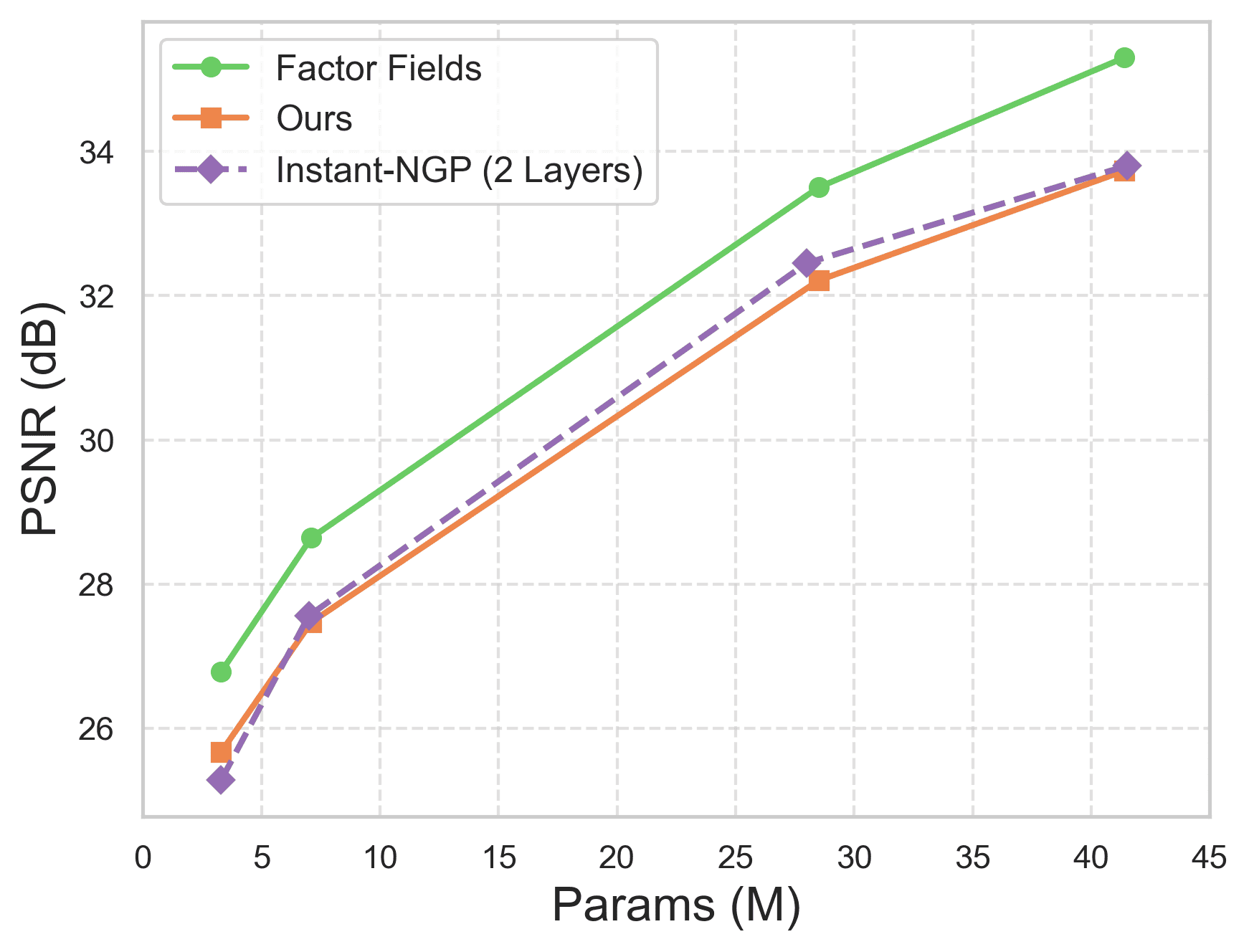} %
        \\

          \includegraphics[height=0.18\textheight]{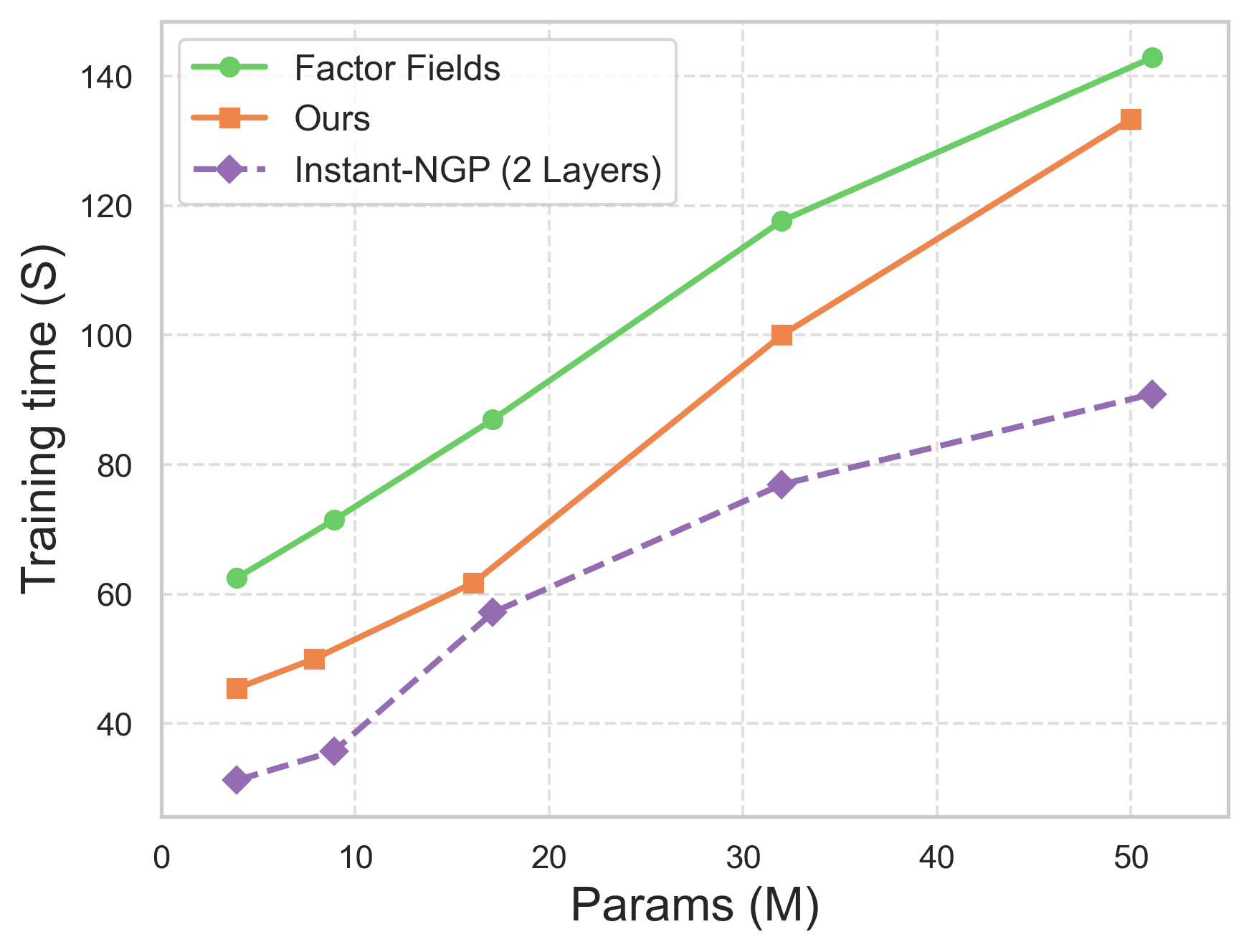} 
        &
        
        \includegraphics[height=0.18\textheight]{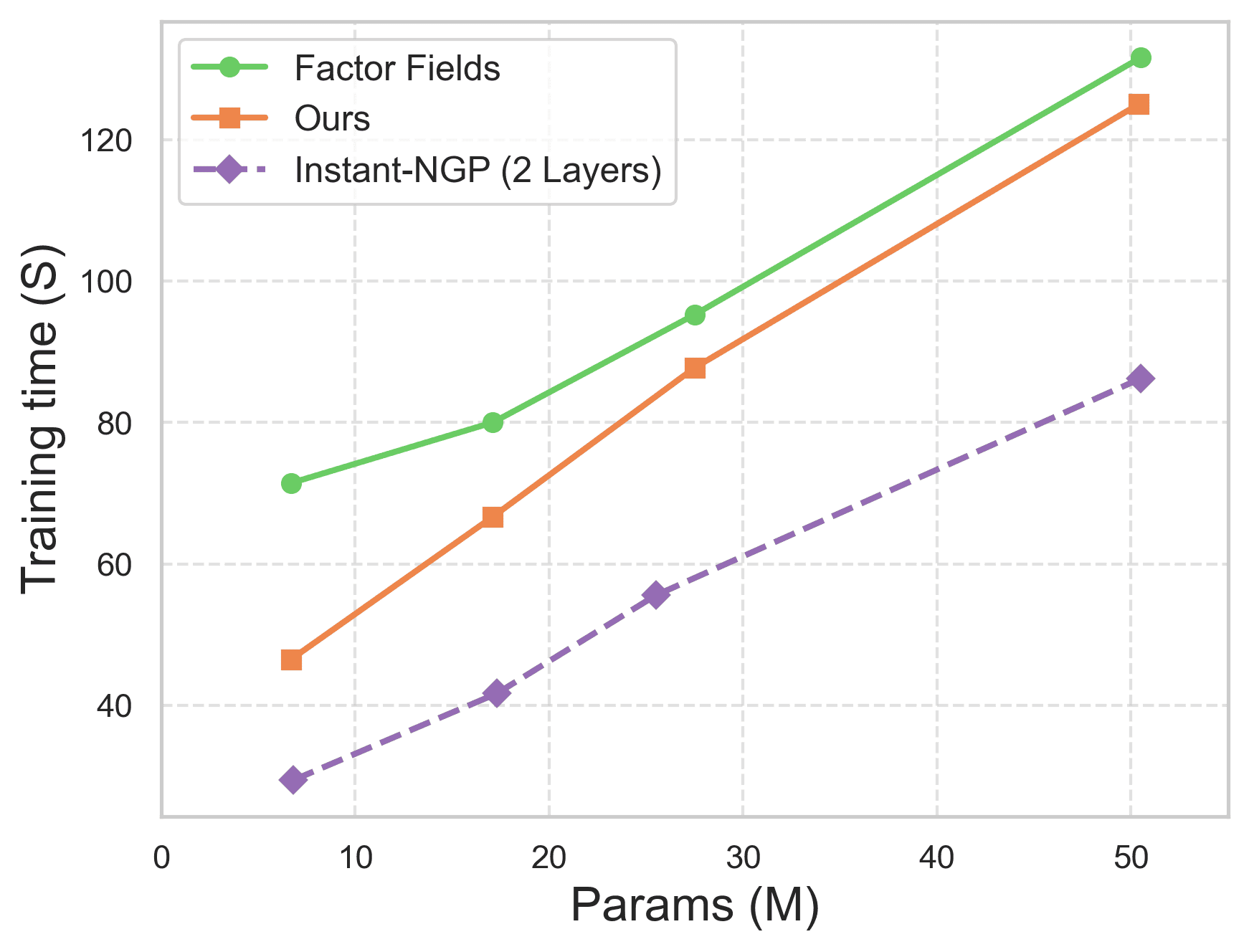} %
        &
        
        \includegraphics[height=0.18\textheight]{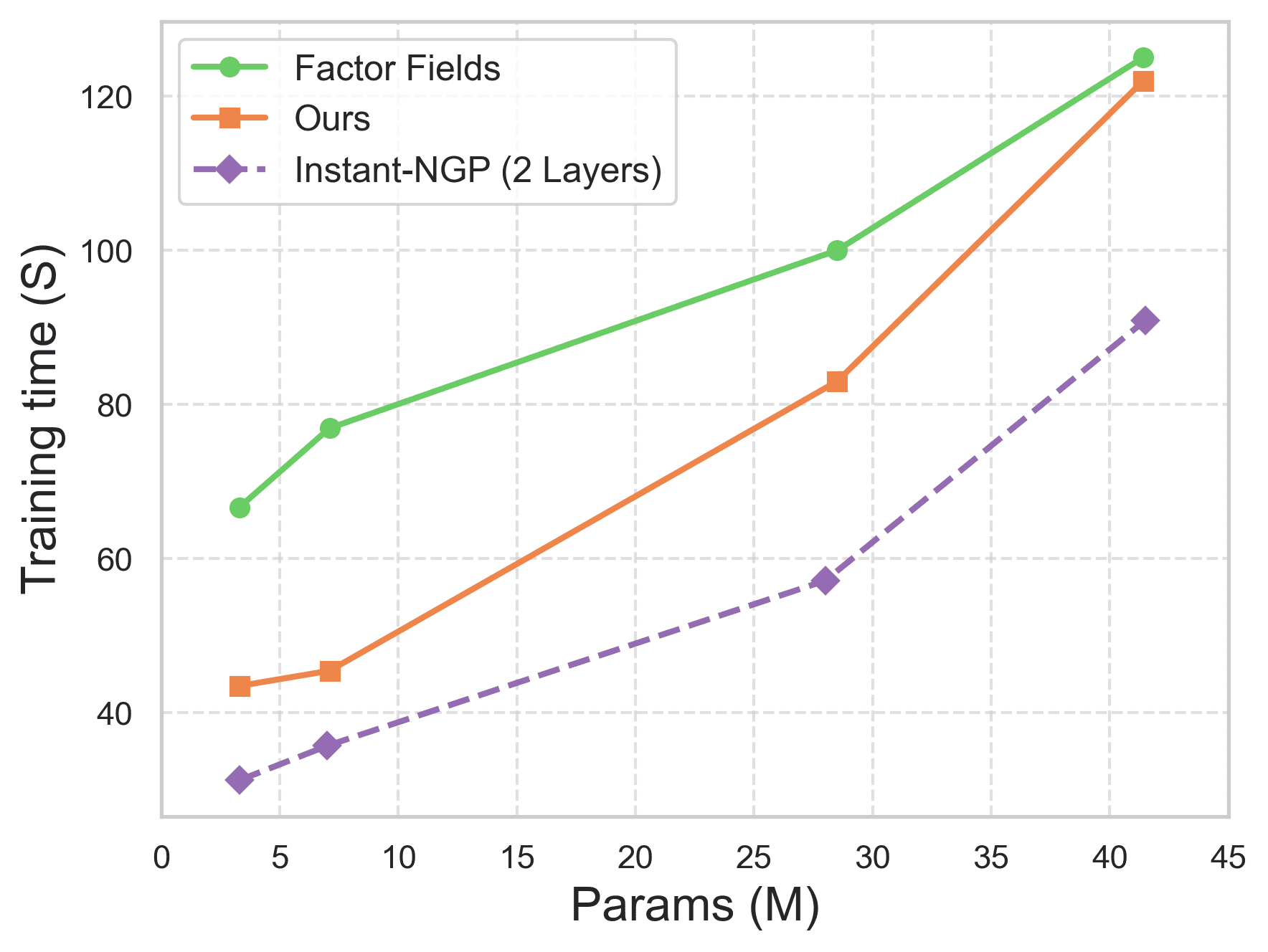} %
             
     \end{tabular}   
     }
     \caption{Comparison of the capacity of the proposed GNF to represent RGB images with InstantNGP~\cite{muller2022instant} and  FactorFields~\cite{chen2023dictionary}. The top row shows the images reconstructed using our method. The captions under the top row refer to  \textcolor{green}{the resolution of the image}, \textcolor{blue}{the total no. of parameters of the network},  \textcolor{yellow}{the training time},  \textcolor{orange}{the reconstruction error in terms of PSNR}, and \textcolor{red}{the number of pixels per second} the model can render at runtime. The second row shows pixel-wise error maps between the images regressed with our model and their corresponding ground truth images. The third row illustrates how the PSNR varies as a function of the number of parameters, while the last row shows how the training time varies as a function of the number of parameters.    
     } 
   
    \label{fig:2d_psnr_training_time} 
\end{figure*}

\subsection{Representation of 2D signals} 

We evaluate the accuracy and efficiency of the proposed \ModelName{} framework in representing multichannel 2D signals, specifically RGB images. We use  three complex, high-resolution images: \textbf{(1)} Summer Day\footnote{Credit to Johan Hendrik Weissenbruch and \href{https://www.rijksmuseum.nl/en/collection/SK-A-3005}{Rijksmuseum}},  \textbf{(2)} Pluto\footnote{Credit to \href{https://solarsystem.nasa.gov/resources/933/true-colors-of-pluto/}{NASA}}, and \textbf{(3)} Girl With a Pearl Earring\footnote{By \href{http://profoundism.com/free_licenses.html}{Koorosh Orooj (CC BY-SA 4.0)}}. These images vary in size, with pixel counts ranging from $4$M to $213$M. To assess the reconstruction quality, we employ the Peak Signal-to-Noise Ratio (PSNR). Efficiency is measured using the number of model parameters, training time,  and inference time.





 We compare our method against two state-of-the-art neural image regression methods: InstantNGP~\cite{muller2022instant} and FactorFields~\cite{chen2023dictionary}.  All the models were trained for $10$K epochs using a batch size of $2^{17}$ pixels randomly sampled from the image.
Figure~\ref{fig:2d_psnr_training_time} (top row) shows images reconstructed using our method. The figure also plots the PSNR  and training time, as functions of the number of trainable parameters, of our method and the state-of-the-art methods. In these plots, higher PSNR values indicate better reconstruction quality, while lower training times indicate greater efficiency. From these results, we observe that:
\begin{itemize}
    \item The reconstruction quality of our method, as measured by the PSNR, outperforms existing methods on the tested images, except FactorFields whose PSNR is slightly higher than our method on two images. 
    
    \item  With the same number of parameters (around $29$M), we observe that on average (from Figure 5 in the Supplementary Material), our method is around $12$ seconds faster to train than FactorFields with only $1$dB drop in reconstruction quality compared to FactorFields.  
\end{itemize}

\noi In summary, the proposed model strikes a balance between reconstruction performance and training efficiency as it uses only a single RBF layer operating on the feature space. Thus, it is highly parallelizable and computationally more efficient. Section 2 in the Supplementary Material provides additional analysis, results, and discussion.

    




\begin{table*}[t]
    \centering
    {
    \begin{tabular}{@{}l@{ }c@{ }cc@{ }c@{ }c@{ }cc@{ }c@{}}
        \toprule
         & \textbf{Batch} &   &    \multicolumn{4}{@{}c@{}}{\textbf{Synthetic-NeRF}} & \multicolumn{2}{@{}c@{}}{\textbf{Tanks\&Temples}} \\
        
        \cmidrule(lr){4-7} \cmidrule(lr){8-9}
        \textbf{Method} &  \textbf{size} & \textbf{Steps} &   \textbf{Tr. time $\downarrow$} & \textbf{Size (M)$\downarrow$}  & \textbf{PSNR$\uparrow$} & \textbf{SSIM$\uparrow$} & \textbf{PSNR$\uparrow$} & \textbf{SSIM$\uparrow$} \\
        \midrule
        NeRF~\cite{mildenhall2020nerf} & 4096 & 300k & $\sim$35h &\cellcolor{green!20} 1.25 & 31.01 & 0.947 & 25.78 & 0.864 \\
        DVGO~\cite{sun2022direct} & 5000 & 30k & 15.0m & 153.0 & 31.95 & 0.957 &\cellcolor{yellow!20} 28.41 & 0.911 \\

        Plenoxels~\cite{fridovich2022plenoxels} &$5000$ &  $128$k & $ 11.4$m & $194.5$ &  $31.71$ & $0.958$ & $27.43$ &  $0.906$ \\
        
        Instant-NGP~\cite{muller2022instant}  & 10-85k & 30k &\cellcolor{green!20} $03.9$m & 11.64 & 32.59 & \cellcolor{yellow!20} 0.960 & 27.09 & 0.905 \\
        NeurRBF~\cite{chen2023neurbf} & 4096 & 30k & 33.6m & 17.74 &\cellcolor{green!20} 34.62 & \cellcolor{green!20}0.975 & - & - \\

         K-Planes~\cite{fridovich2023k}  & 4096 & $30$K & 38m &33.0  & 32.36 & - &- &- \\
        
         DiF-Grid~\cite{chen2023dictionary}  & 4096 & $30$K &\cellcolor{yellow!20} 12.2m &\cellcolor{orange!20} 05.10  &\cellcolor{orange!20} 33.14 &\cellcolor{orange!20} 0.961 & \cellcolor{green!20}29.00 &\cellcolor{green!20} 0.938 
         \\

         \midrule
        \textbf{HashGrid-GNF-SH (Ours)} & 4096 & $30$K &\cellcolor{orange!20} $10.2$m &\cellcolor{yellow!20} $10.30$  & \cellcolor{yellow!20} $32.69$ &  $0.957$ & \cellcolor{orange!20}$28.73$ &\cellcolor{orange!20} $0.921$\\

        \bottomrule
    \end{tabular}
    }
    \caption{Neural Radiance Field reconstruction. We quantitatively compare the performance of the proposed models against different NeRF methods on Synthetic-NeRF~\cite{mildenhall2020nerf}  and Tanks and Temples~\cite{knapitsch2017tanks} datasets. The first, second, and third best values in each column are highlighted in \textcolor{green}{green}, \textcolor{orange}{orange}, and \textcolor{yellow}{yellow}, respectively. "Tr. time" refers to training time.}
    \label{tab:nerf_1vs1}
\end{table*}

\begin{table}[t]
    \centering
  \resizebox{.5\textwidth}{!}
    {
    \begin{tabular}{@{}l@{ }c@{ }c@{ }c@{ }cc@{ }c@{}}
        \toprule
         &    \multicolumn{4}{@{}c@{}}{\textbf{Synthetic-NeRF}} & \multicolumn{2}{@{}c@{}}{\textbf{Tanks\&Temples}} \\
        
        \cmidrule(lr){2-5} \cmidrule(lr){6-7}
        \textbf{Method}  &   \textbf{Tr. time $\downarrow$} & \textbf{Size (M)$\downarrow$}  & \textbf{PSNR$\uparrow$} & \textbf{SSIM$\uparrow$} & \textbf{PSNR$\uparrow$} & \textbf{SSIM$\uparrow$} \\
        \midrule

         K-Planes-MLP~\cite{fridovich2023k} &  $38.0$m &$33.00$  & 32.36 & - &- &- \\
          \textbf{K-Plane-GNF-SH (Ours)} & \cellcolor{green!20}$18.0$m &\cellcolor{green!20} $31.00$  & \cellcolor{green!20}$32.43$ & $0.956$ & $28.40$ &$0.919$\\    
         \midrule

         HashGrid-MLP-SH~\cite{muller2022instant} &$10.80$m & $16.40$  & $32.60$ &  $0.955$ & $28.60$ &\cellcolor{green!20}$0.922$\\
         
        \textbf{HashGrid-GNF-SH (Ours)}   &\cellcolor{green!20} $10.20$m &\cellcolor{green!20} $16.30$  & \cellcolor{green!20} $32.69$ & \cellcolor{green!20}  $0.957$ & \cellcolor{green!20}$28.73$ & $0.921$\\

        \midrule
        
         Dif-Grid-MLP-SH~\cite{chen2023dictionary}  & $10.98$m & $5.34$  &\cellcolor{green!20} $32.63$ &\cellcolor{green!20} $0.958$ & $27.07$ &$0.907$\\
         \textbf{Dif-Grid-GNF-SH (Ours)}   &\cellcolor{green!20} $10.09$m &\cellcolor{green!20} $5.32$  & $32.60$ &\cellcolor{green!20} $0.958$ & $27.00$ &\cellcolor{green!20} $0.909$\\
        \bottomrule
    \end{tabular}
    }
    \caption{Ablation study that demonstrates the importance of the decoder. Each encoder (K-Planes, HashGrid, Dif-Grid) is paired with both MLP and our GNF decoder while keeping all other components identical. All methods were trained for 30K steps with a batch size of $4096$.}
    \label{tab:side_side_comparison}
\end{table}

\begin{figure*}[t]
    \centering
      \begin{tabular}{@{}c@{}c@{}c@{}c@{}}
        \centering
        \includegraphics[width=0.25\textwidth]{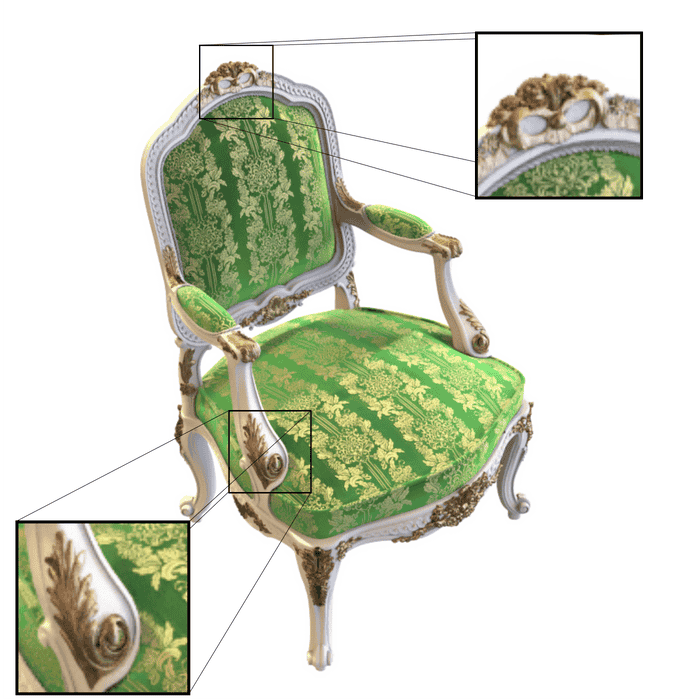}  &
        \includegraphics[width=0.25\textwidth]{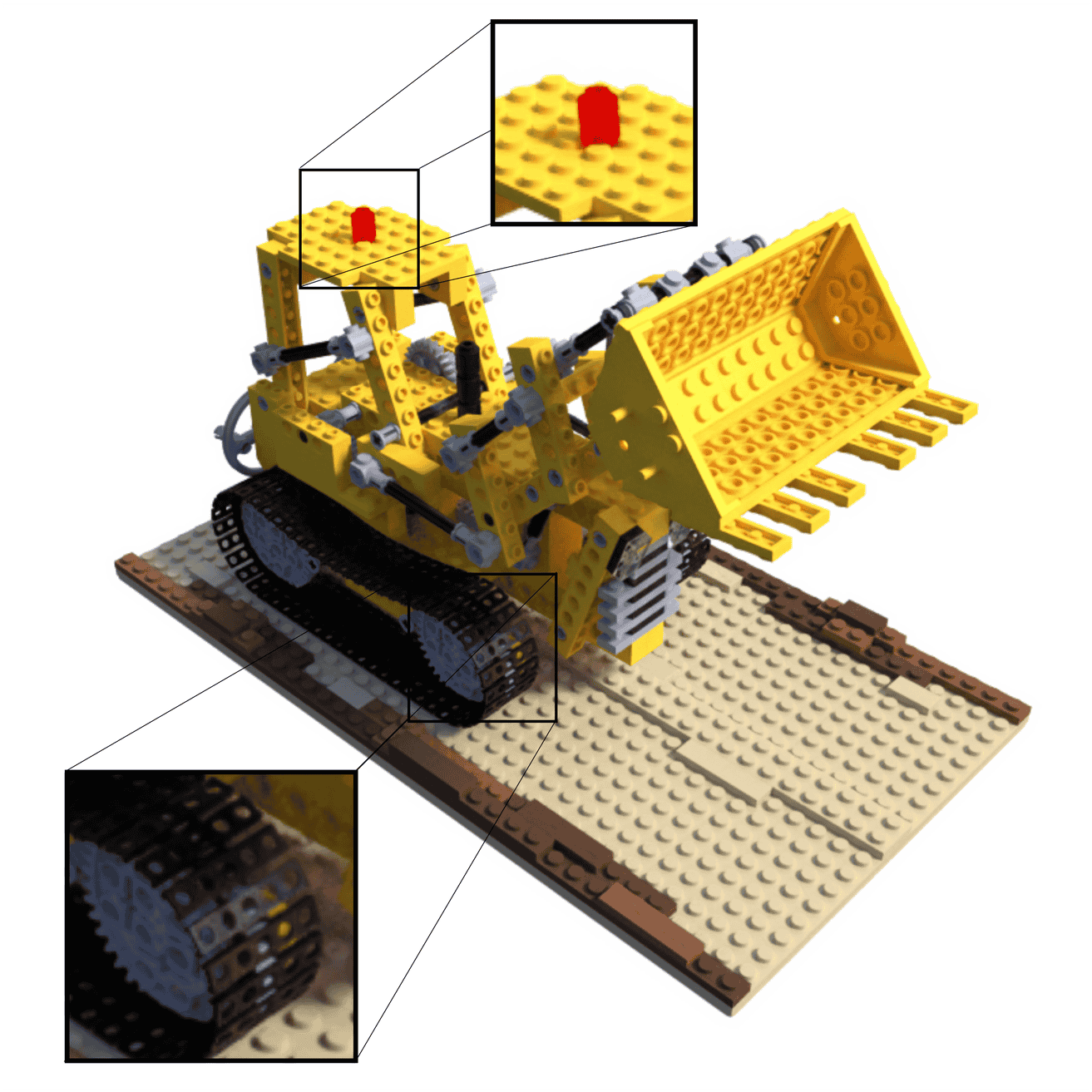} &
         \includegraphics[width=0.25\textwidth]{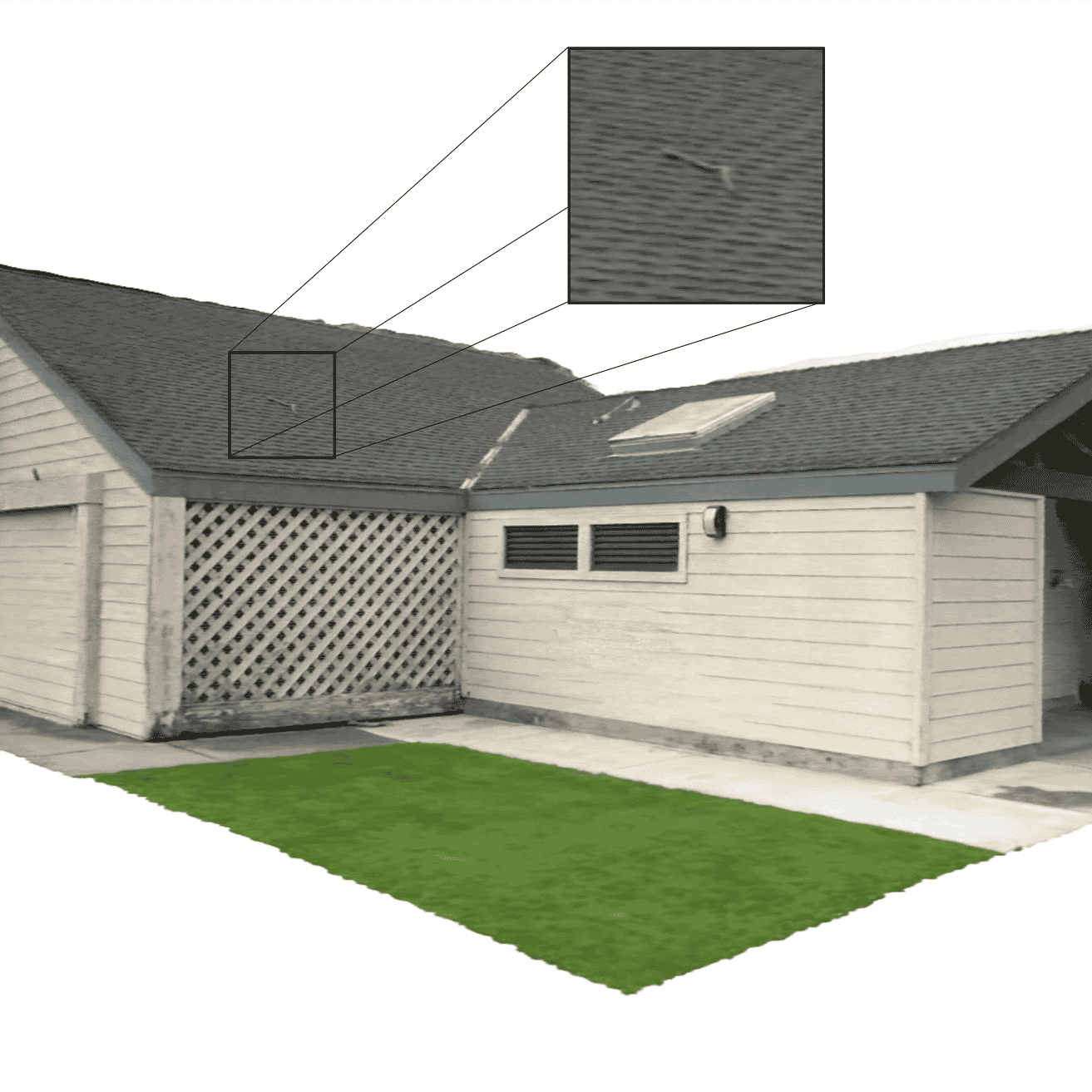} &
         \includegraphics[width=0.25\textwidth]{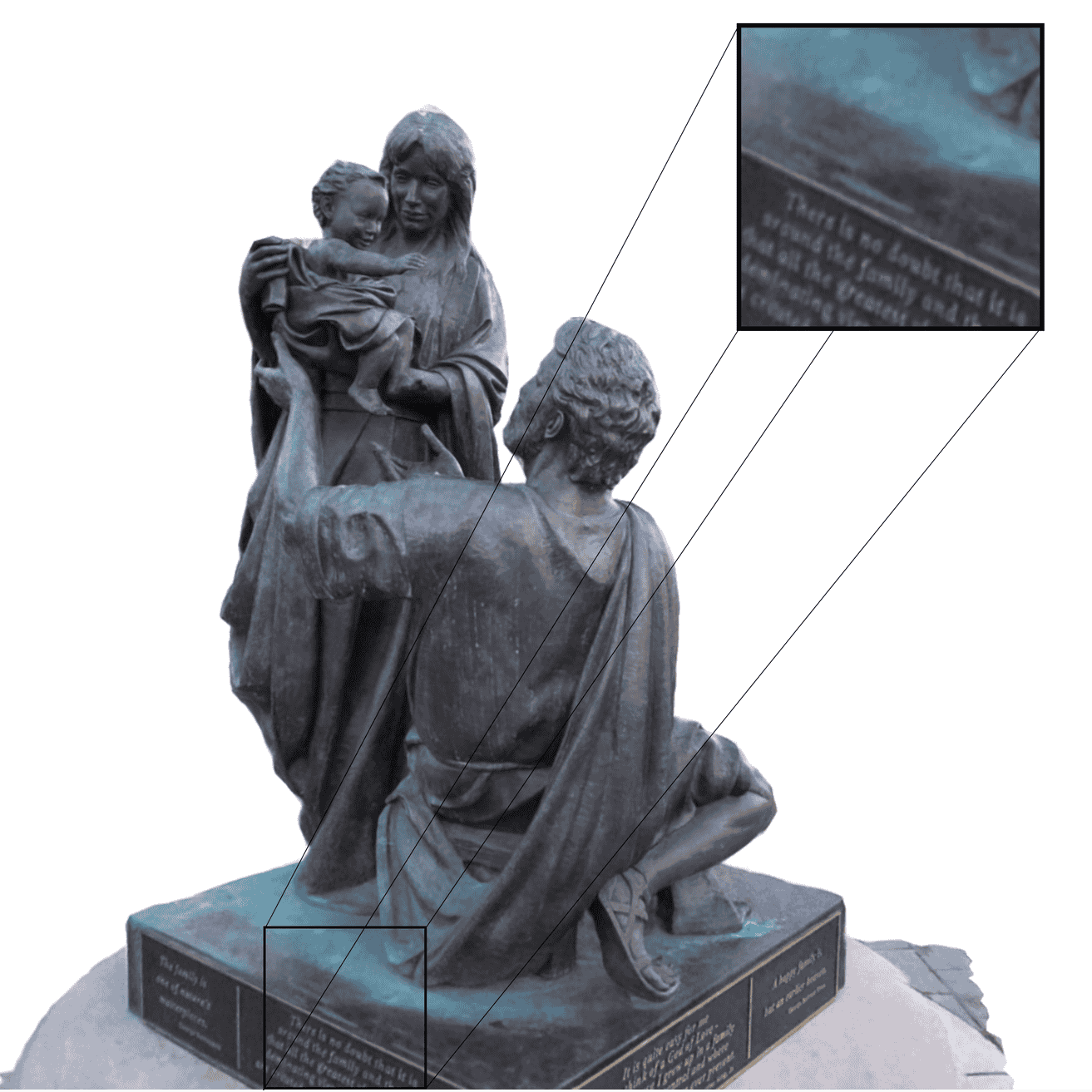} \\
        \multicolumn{4}{c}{\textbf{(a)} HashGrid-GNF-SH (Ours).} \\
        
        \includegraphics[width=0.25\textwidth]{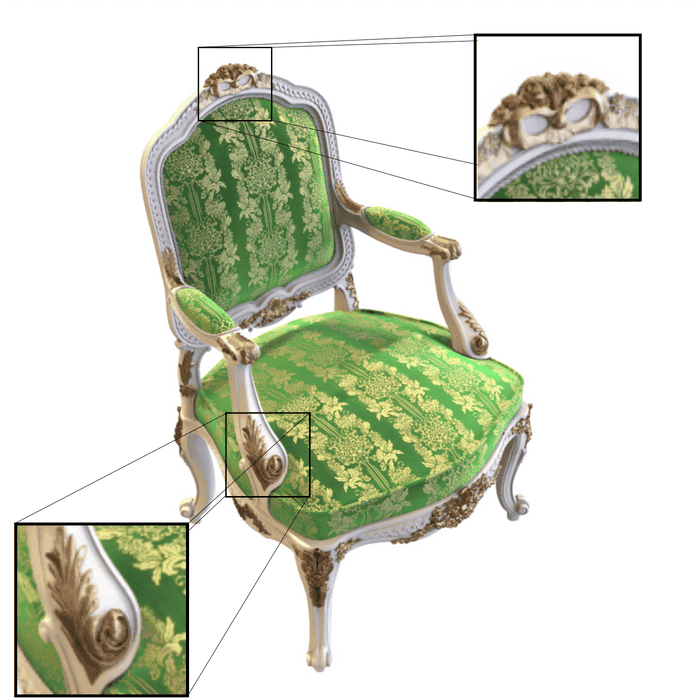} &
        \includegraphics[width=0.25\textwidth]{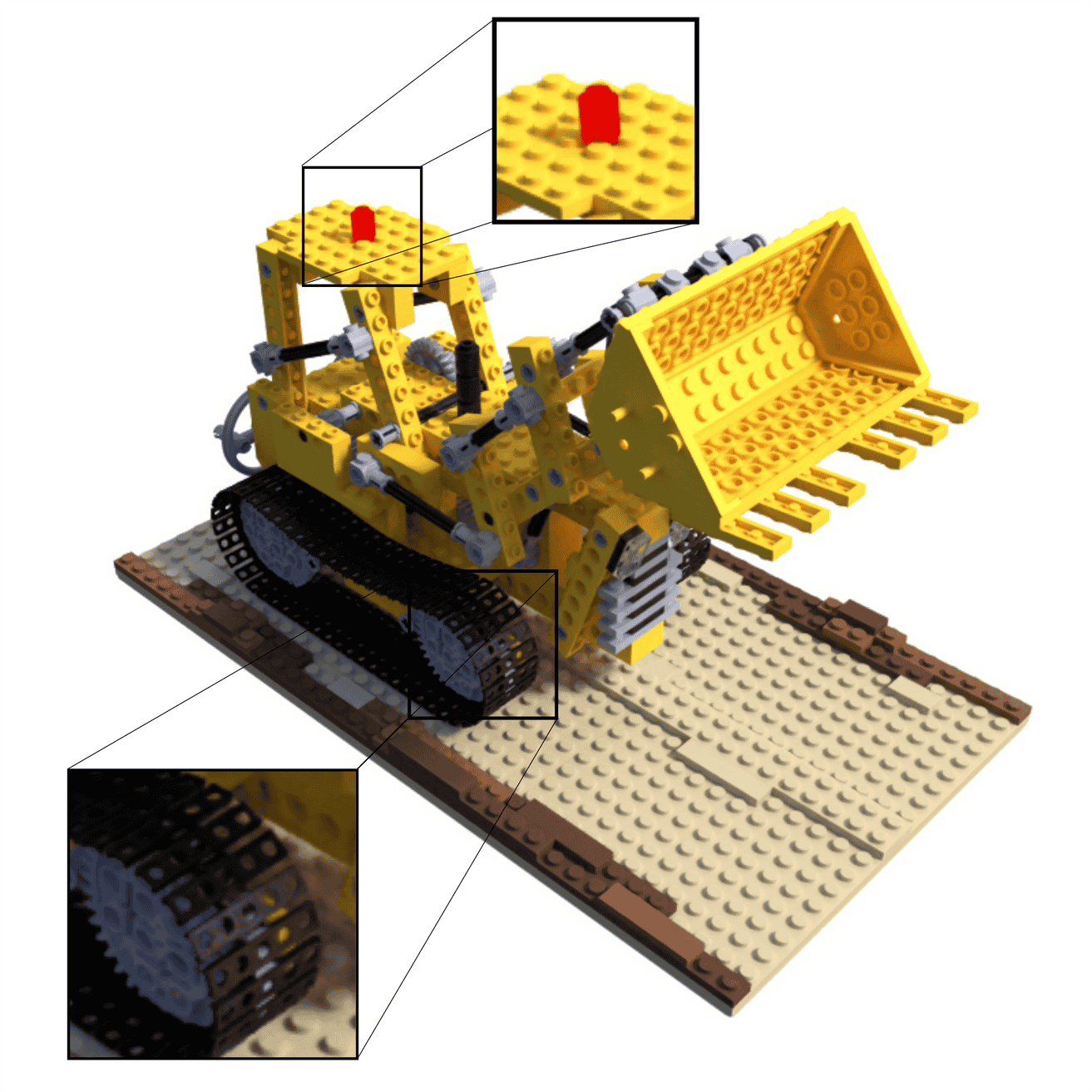} &
        \includegraphics[width=0.25\textwidth]{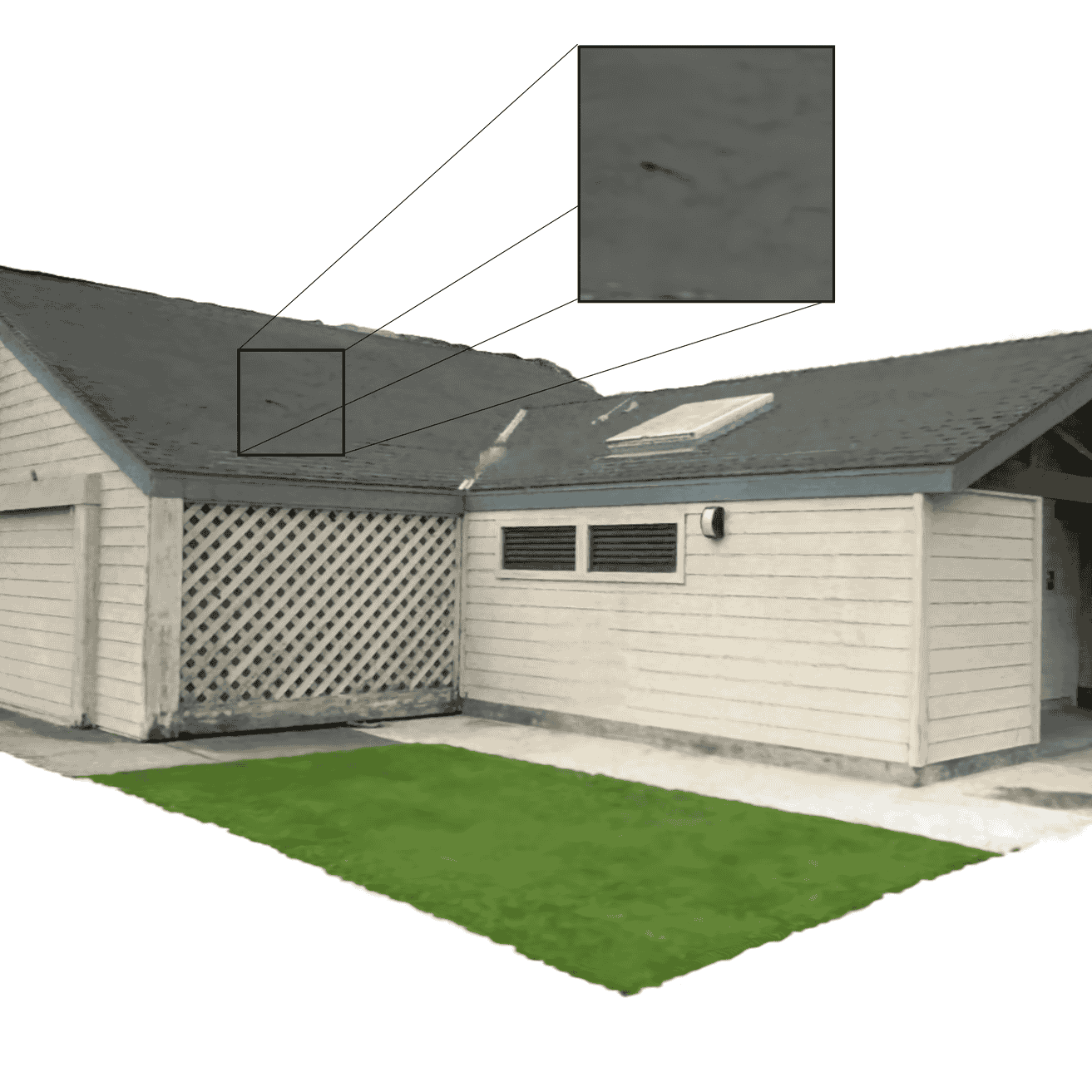} &
        \includegraphics[width=0.25\textwidth]{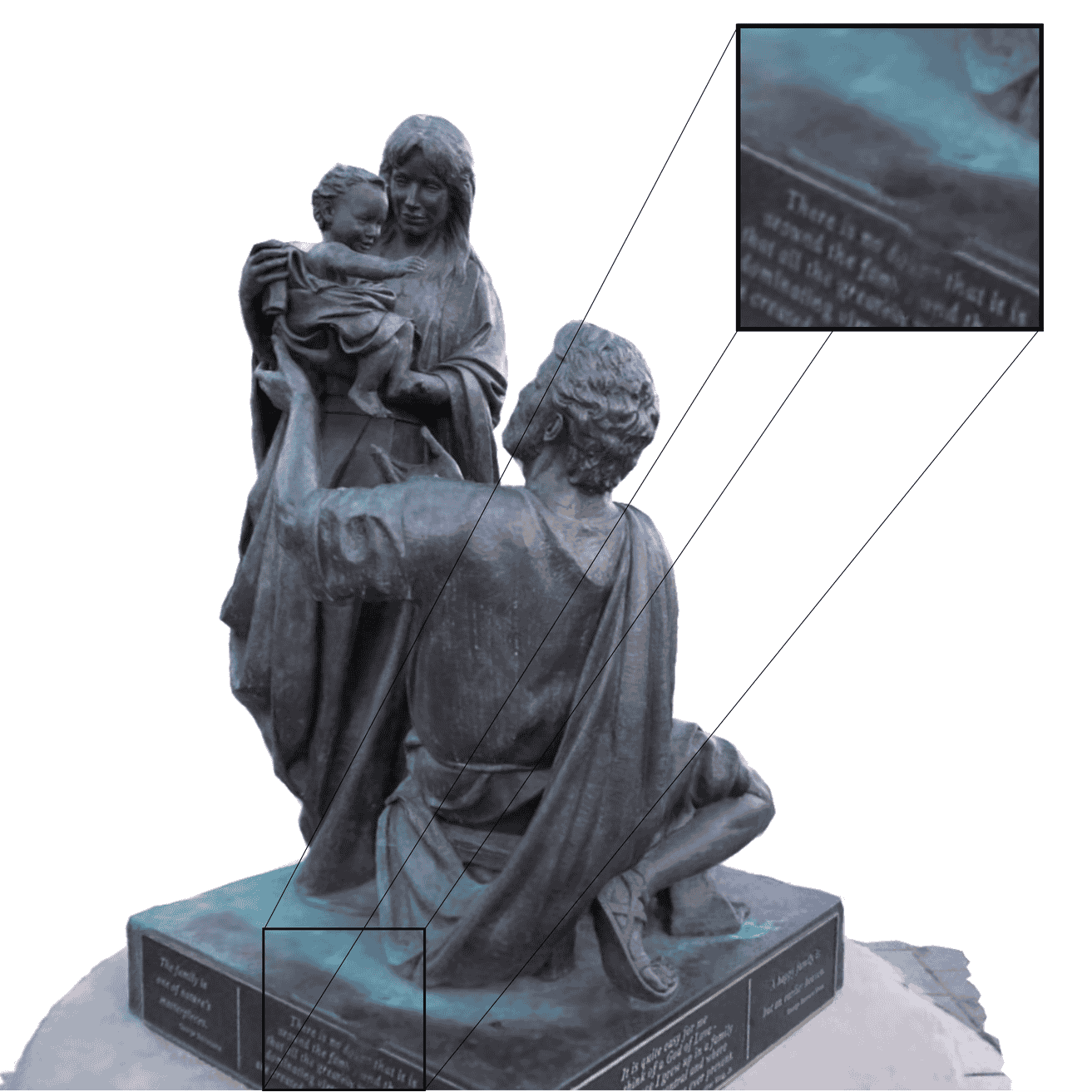} \\
        
        \multicolumn{4}{c}{\textbf{(b)} DiF-Grid (FactorFields)~\cite{chen2023dictionary}.} \\        
    
        \includegraphics[width=0.25\textwidth]{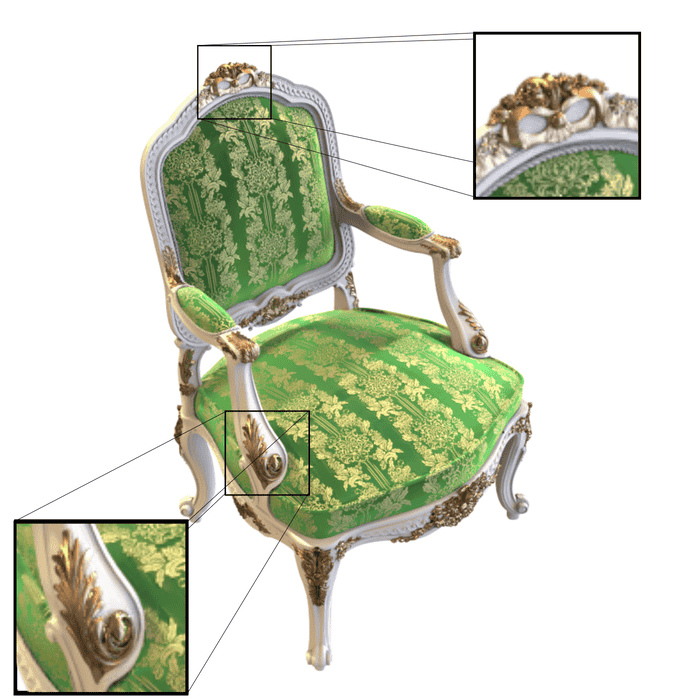}  &
        \includegraphics[width=0.25\textwidth]{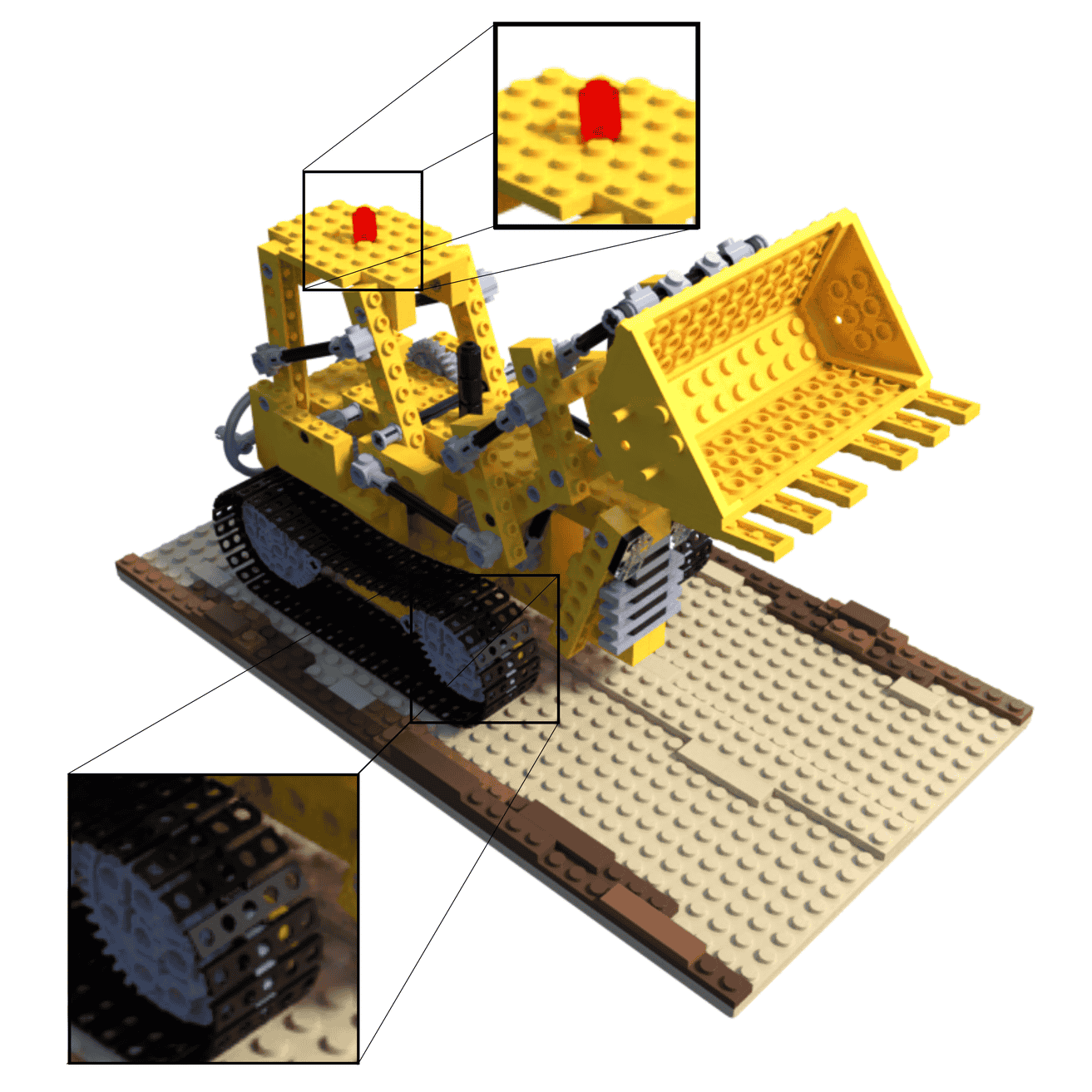} &
        \includegraphics[width=0.25\textwidth]{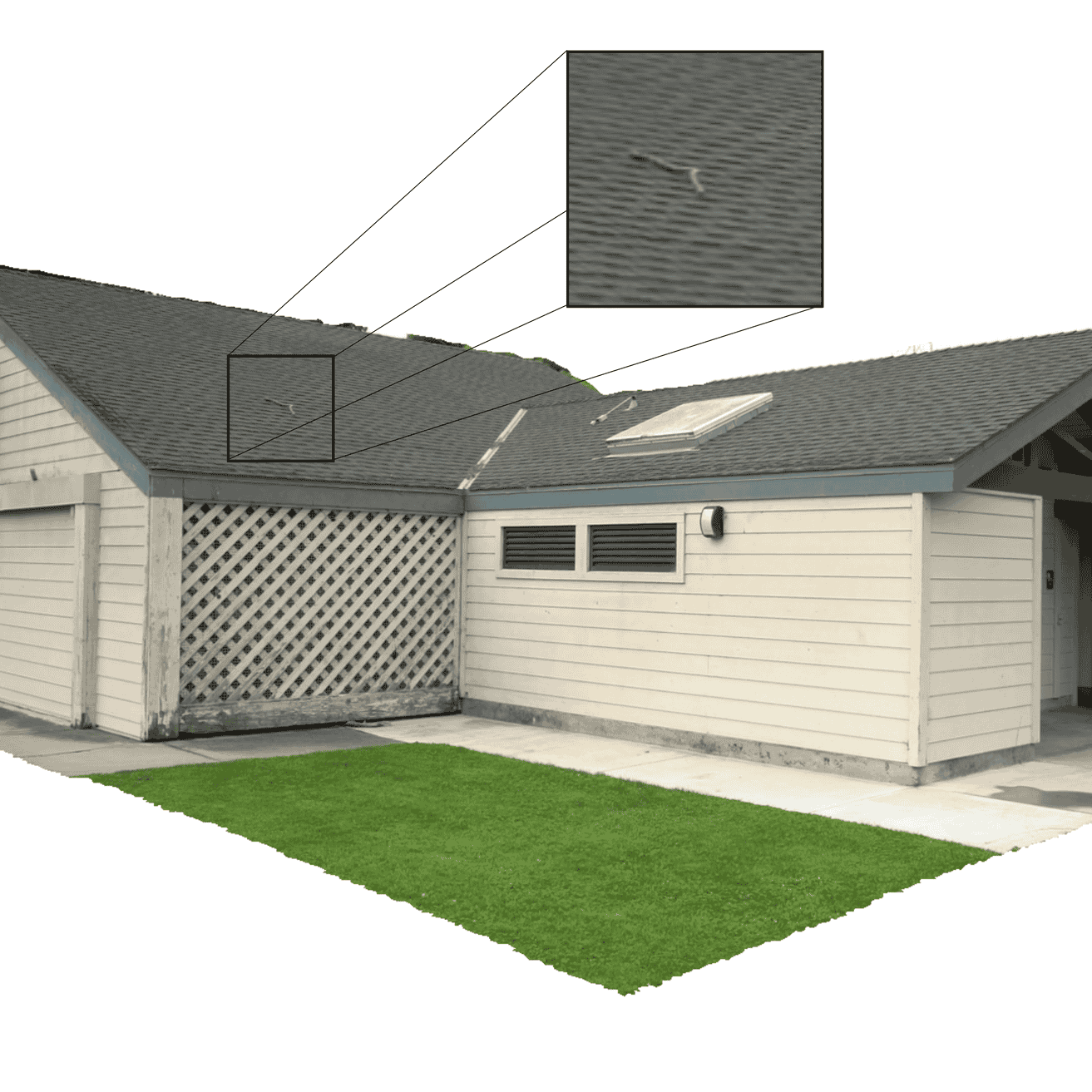} &
        \includegraphics[width=0.25\textwidth]{figures/Nerf/Family/ours_family.png}\\            
        \multicolumn{4}{c}{\textbf{(c)} Ground Truth.}
     \end{tabular} 

    \caption{Qualitative comparison of novel view generation using our model, DiF-Grid (FactorFields) model, and ground truth test images. Observe in the case of the Chair that our model is able to capture the green reflection of the chair color on the decorative pattern (top row, bottom left zoom), which is not the case with other methods. The Lego example (second column) shows that our model provides better color fidelity of the reflectance on the light (see the top-right zoom). In the third column, we can see that our model is able to recover detailed rooftop patterns. It is also able to capture the stick. The last column shows that our method is able to accurately capture the text and the shape. Overall, our model can capture small details that are missed by FactorFields.}
    \label{fig:novel_views_comparison_grid}
\end{figure*}

\subsection{Radiance fields} 
\label{sec:results_radiance_field}
Finally, we evaluate the performance of the proposed framework on novel view synthesis using the synthetic NeRF~\cite{mildenhall2020nerf} and Tanks and Temples~\cite{knapitsch2017tanks} datasets. Table~\ref{tab:nerf_1vs1} presents a quantitative comparison of our proposed model against the baseline NeRF~\cite{mildenhall2020nerf} and state-of-the-art methods, focusing on training time, model size, and the quality of the rendered novel views. While InstantNGP remains the fastest due to its custom CUDA-based implementation,  our \ModelName{} decoder,  built on the standard PyTorch framework,  achieves a competitive average training time of $10.2$ mins, which is $2$ mins and $5$ mins faster than DiF-Grids (FactorFields) and DVGO, respectively. It is significantly faster than NeuRBF, which takes $33.6$ mins, and the baseline NeRF, which takes more than $35$ hrs to train. The use of PyTorch not only ensures ease of implementation but also facilitates extensibility to other tasks.

Table~\ref{tab:side_side_comparison} presents a one-to-one comparison between our RBF-based decoder and standard 3-layer MLP decoders, evaluated across three different encoding strategies: Hash encoding~\cite{muller2022instant}, K-Planes~\cite{fridovich2023k}, and the DiF-Grid model from DiF-Grid (FactorFields)~\cite{chen2023dictionary}. This experiment highlights the benefits of replacing the conventional MLP decoder with our compact, single-layer RBF-based architecture. While the  model size (measured in terms of the number of parameters) is largely dominated by the encoder, our decoder offers three key advantages: 
\begin{itemize}
    \item It uses less than half the parameters of its MLP counterpart.
    \item It delivers a comparable or a slightly superior reconstruction quality, and 
    \item It achieves faster training times, offering a consistent speed-up across all tested encoders. 
\end{itemize}    
    
\noi For example, with the K-Planes encoder, our model trains nearly twice as fast as the original K-Planes method, which uses MLP-based decoding. Importantly, during inference, the decoding stage is the primary bottleneck in MLP-based methods due to their layered computation. Interpolation-based methods such as Plenoxels~\cite{fridovich2022plenoxels} are fast at inference time, since they do not use MLPs, but are very expensive in terms of memory requirements. For instance, Plenoxels' size is $194$M; see Table~\ref{tab:nerf_1vs1}. Our lightweight RBF-based decoder alleviates this bottleneck, enabling faster and more scalable inference without compromising quality. 

Figure~\ref{fig:novel_views_comparison_grid} presents a comparative analysis of novel views generated by our model and DiF-Grid (FactorFields)~\cite{chen2023dictionary}. In the Chair example ($1$st column), our model successfully captures the green reflection of the chair's color on the decorative pattern (top row, bottom-left zoom), a geometric detail that is missed by other methods. In the Lego example ($2$nd column), our model demonstrates superior color fidelity, accurately reflecting the light's color (see the top-right zoom).  The third column highlights that our model recovers the detailed rooftop patterns. It even captures the stick with high accuracy. Finally, in the last column, our model accurately captures both the text and the shape. Our model effectively captures fine details that FactorFields misses.


\section{Conclusion}
\label{sec:conclusion}
This paper demonstrates that a single-layer Radial Basis Function (RBF) network can effectively represent complex signals, including RGB images, 3D geometry, and radiance fields. Our method achieves comparable accuracy to state-of-the-art approaches while utilizing the same or fewer parameters. Notably, it offers significantly faster training and inference times. This efficiency positions our framework as a valuable step towards more compact and scalable neural field representations. While we chose Gaussian kernels for their desirable mathematical properties, \ie smoothness, continuity, and infinite differentiability, thus making them well-suited for gradient-based optimization, alternative kernels may perform better for specific signal types. Future work will explore spatially adaptive kernel selection methods that can dynamically choose the appropriate kernel based on the local properties of the signal.
Other promising avenues for future research include the extension of our framework to dynamic and unbounded scenes, as well as investigating the theoretical and practical benefits of stacking multiple RBF layers.

\vspace{3pt}
\noi\textbf{Acknowledgment.} This work is supported by the Australian Research Council (ARC) Discovery Project no. DP220102197 and the Region Hauts-de-France in France.

\bibliographystyle{IEEEtran}
\bibliography{references}

\end{document}

%% file: notation.tex
\newcommand{\estimated}[1]{\tilde{#1}}

\newcommand{\npoints}{n}	
\newcommand{\point}{p}
\newcommand{\viewdir}{\textbf{v}}
						
\newcommand{\thecolor}{c}

\newcommand{\width}{w}                
\newcommand{\batchsize}{B}            

\newcommand{\FLOPs}{\text{FLOPs}}     

\newcommand{\bigO}[1]{\mathcal{O}(#1)}
\newcommand{\domain}{\Omega}					
\newcommand{\inputdim}{d}
\newcommand{\outputdim}{q}
\newcommand{\featuredim}{m}
\newcommand{\featdim}{\featuredim}              
\newcommand{\ModelName}{GNF}
\newcommand{\neuralfunc}{f}

\newcommand{\enc}{\gamma}
\newcommand{\dec}{h}
\newcommand{\geometrydec}{\dec_\text{geo}}
\newcommand{\colordec}{\dec_\text{color}}

\newcommand{\density}{\sigma}

\newcommand{\params}{\Theta}

\newcommand{\featurevector}{\textbf{x}}   
\newcommand{\latentcode}{\featurevector}
\newcommand{\geometrylatentcode}{\latentcode_\text{geo}}

\newcommand{\meanfeaturevector}{\bar{\featurevector}}

\newcommand{\radialbasis}{\mathbf{b}}
\newcommand{\bases}{\mathbf{B}}
\newcommand{\weight}{w}
\newcommand{\Weight}{W}
\newcommand{\weights}{\mathbf{\Weight}}

\newcommand{\thecenter}{c}

\newcommand{\bandwidth}{\beta}
\newcommand{\polynomial}{\Phi}
\newcommand{\nbases}{N}          
\newcommand{\nrbf}{\nbases}                 

\newcommand{\loss}{\mathcal{L}}
\newcommand{\lone}{\mathcal{L}_1}
\newcommand{\ltwo}{\mathcal{L}_2}

\newcommand{\signal}{\neuralfunc}

\newcommand{\s}{\ensuremath{\mathbb{S}}}

\newcommand{\stwo}{\ensuremath{\mathbb{S}^2}}

\newcommand{\sn}[1]{\s^{n}}

\newcommand{\noi}{\noindent}

\newcommand{\eg}{\emph{e.g.,\  }}
\newcommand{\ie}{\emph{i.e.,\ }}

\newcommand{\etal}{\emph{et al.}}

\newcommand{\etalnospace}{\emph{et al.}}

\usepackage{framed, color}
\definecolor{shadecolor}{rgb}{1,0.5,0}

\newcommand{\real}{\mathbb{R}}

\newcommand{\rplus}{\real_{>0}}
\newcommand{\rnonneg}{\real_{\geq 0}}
\newcommand{\rtwo}{\real^{2}}

\newcommand{\rcubed}{\real^{3}}
\newcommand{\rthree}{\rcubed}

